\documentclass[11pt]{article}
\usepackage[pdftex]{graphicx}
\usepackage[small,bf,sf,textfont={small,sf}]{caption}
\usepackage{vmargin}
\usepackage{subfig}
\usepackage{amsmath}
\usepackage{amssymb}
\usepackage{fancyheadings}
\usepackage{color}
\usepackage[colorlinks=true,
            linkcolor=blue,
            urlcolor=blue,
            citecolor=blue]{hyperref}
\usepackage[round]{natbib}
\usepackage{doi}
\usepackage{booktabs}
\usepackage{float}
\usepackage{lineno}
\usepackage{epstopdf}
\usepackage{bm}
\usepackage{dsfont}
\usepackage{multirow}
\usepackage{pdflscape}

\floatstyle{ruled}
\newfloat{floatbox}{thp}{lob}
\floatname{floatbox}{Algorithm}

\newcommand{\trp}{^{\scriptsize \textrm{T}}}

\newcommand{\inv}{^{\scriptsize -1}}

\newcommand{\x}{\mathbf{x}}

\newcommand{\z}{\mathbf{z}}

\newcommand{\e}{\mathbf{e}}

\newcommand{\w}{\mathbf{w}}

\newcommand{\dsim}{\mathbf{d}}
\newcommand{\dobs}{\mathbf{d}_{\textrm{obs}}}

\newcommand{\C}{\mathbf{C}}

\newcommand{\Ce}{\mathbf{C}_{\mathbf{e}}}

\setpapersize{USletter}
\setmarginsrb{1.5in}{1in}{1in}{0.5in}{12pt}{11mm}{12pt}{30pt}

\begin{document}


\title{Towards a Robust Parameterization for Conditioning Facies Models Using Deep Variational Autoencoders and Ensemble Smoother}

\author{Smith W. A. Canchumuni$^\star$, Alexandre A. Emerick$^\dag$ and Marco Aur{\'e}lio C. Pacheco$^\star$}
\maketitle

\begin{center}
\small $^\star$PUC-RIO \\
\small $^\dag$Petrobras
\end{center}

\section*{Abstract}
\label{Sec:Abstract}

History matching is a jargon used to refer to the data assimilation problem in oil and gas reservoirs. The literature about history matching is vast and despite the impressive number of methods proposed and the significant progresses reported in the last decade, conditioning reservoir models to dynamic data is still a challenging task. Ensemble-based methods are among the most successful and efficient techniques currently available for history matching. These methods are usually able to achieve reasonable data matches, especially if an iterative formulation is employed. However, they sometimes fail to preserve the geological realism of the model, which is particularly evident in reservoir with complex facies distributions. This occurs mainly because of the Gaussian assumptions inherent in these methods. This fact has encouraged an intense research activity to develop parameterizations for facies history matching. Despite the large number of publications, the development of robust parameterizations for facies remains an open problem.

Deep learning techniques have been delivering impressive results in a number of different areas and the first applications in data assimilation in geoscience have started to appear in literature. The present paper reports the current results of our investigations on the use of deep neural networks towards the construction of a continuous parameterization of facies which can be used for data assimilation with ensemble methods. Specifically, we use a convolutional variational autoencoder and the ensemble smoother with multiple data assimilation. We tested the parameterization in three synthetic history-matching problems with channelized facies. We focus on this type of facies because they are among the most challenging to preserve after the assimilation of data. The parameterization showed promising results outperforming previous methods and generating well-defined channelized facies. However, more research is still required before deploying these methods for operational use.

\section{Introduction}
\label{Sec:Intro}

Ensemble-based methods have been applied with remarkable success for data assimilation in geosciences. However, these methods employ Gaussian assumptions in their formulation, which make them better suited for covariance-based (two-point statistics) models \citep{guardiano:93}. This fact lead several researches to propose a variate of parameterizations to adapt these methods for models with non-Gaussian priors, such as models generated with object-based \citep{deutsch:98bk} and multiple-point geostatistics \citep{mariethoz:14bk}. Among these parameterizations, we can cite, for example, truncated plurigaussian simulation \citep{liu:05a,agbalaka:08,sebacher:13a,zhao:08}; level-set functions \citep{moreno:08a,chang:10,moreno:11a,lorentzen:12a,ping:14a}; discrete cosine transform \citep{jafarpour:08a,zhao:17a,jung:17a}; Wavelet transforms \citep{jafarpour:10a}; K-singular value decomposition \citep{sana:16a,kim:18a}; kernel principal component analysis (KPCA) \citep{sarma:08,sarma:09}; PCA with thresholds defined to honor the prior cumulative density function \citep{chen:14b,chen:15a,gao:15a,honorio:15a} and optimization-based PCA (OPCA) \citep{vo:14a,emerick:17a}. There are also works based on updating probability maps followed by re-sampling steps with geostatistical algorithms \citep{tavakoli:14a,chang:15a,jafarpour:11c,duc:15b,sebacher:15a}. However, despite the significant number of works, the development of robust parameterizations for facies data assimilation remains an open problem. One clear indication that facies parameterization is an unsolved issue is the fact that the large majority of the publications consider only small 2D problems.

Deep learning became the most popular research topic in machine learning with revolutionary results in areas such as computer vision, natural language processing, voice recognition and image captioning, just to mention a few. The success of deep learning in different areas has inspired applications in inverse modeling for geosciences. Despite the fact that the first investigations in this direction are very recent, the number of publications grew very fast in the last two years. For example, \citet{mosser:17a} used a generative adversarial network (GAN) \citep{goodfellow:14b} to generate three-dimensional images of porous media. \citet{laloy:17b} used a variational autoencoder (VAE) \citep{kingma:13a} to construct a low-dimensional parameterization of binary facies models for data assimilation with Markov chain Monte Carlo. Later in \citep{laloy:18a}, the same authors extended the original work using spatial GANs. \citet{canchumuni:17a} used an autoencoder to parameterize binary facies values in terms of continuous variables for history matching with an ensemble smoother. Later, \citet{canchumuni:18a} extended the same parameterization using deep belief networks (DBN) \citep{hinton:06a,hinton:06b}. \citet{chan:17a} used a Wasserstein GAN \citep{arjovsky:17a} for generating binary channelized facies realizations. In \citep{chan:18a}, the same authors coupled an inference network to a previously trained GAN to generate facies realizations conditioned to facies observations (hard data). \citet{dupont:18a} also addressed the problem of conditioning facies to hard data. They used a semantic inpainting with GAN \citep{yeh:16a}. \citet{liu:18a} used the fast neural style transfer algorithm \citep{johnson:16a} as a generalization of OPCA to generate conditional facies realizations using randomized maximum likelihood \citep{oliver:96e}.

The present work is a continuation of the investigation reported in \citep{canchumuni:17a,canchumuni:18a} in the sense that it is also based on using an autoencoder-type of network to construct a continuous parameterization for facies. However, the present work addresses the fact that our previous results were limited to small problems due to difficulties to train the neural networks and the fact that the resulting facies realizations did not preserve the desired geological realism. Here, we investigate the use of convolutional VAE (CVAE) to construct the parameterization. Note that \citet{laloy:17b} also used a CVAE to parameterize facies. Unlike \citet{laloy:17b}, we consider the use of this parameterization in conjunct with an ensemble smoother for assimilation of hard data and dynamic (production) data. A similar approach was recently applied to parameterize seismic data for history matching with an ensemble smoother by \citep{liu:18b}.

The rest of the paper is organized as follows. In the next section, we briefly review generative models. In this section, we describe autoencoders, VAE and convolutional layers. After that, we describe the proposed parameterization for data assimilation applied to petroleum reservoirs using the method ensemble smoother with multiple data assimilation (ES-MDA) \citep{emerick:13b}. Then, we present three test problems with increasing level of complexity followed by comments on potential issues in the parameterization. The last section of the paper summarizes the conclusions. All data and codes used in this paper are available for download at \href{https://github.com/smith31t/GeoFacies_DL/}{\texttt{https://github.com/smith31t/GeoFacies\_DL}}.

\section{Generative Models}
\label{Sec:GM}

Generative models are machine learning methods designed to generate samples from complex (and often with unknown closed form) probability distributions in high-dimensional spaces. These methods use unsupervised and semi-supervised techniques to learn the structure of the input data so it can be used to generate new instances.

Let $\x \in \mathds{X}$ denote a vector in the space $\mathds{X}$ containing the facies values of a reservoir model and assume that each realizations of $\x$ are distributed according to some probability density function (PDF) $p(\x)$. Our goal is to construct a generative model that can create new random realizations of facies that are (hopefully) indistinguishable from samples of $p(\x)$. For concreteness, consider a deterministic function, $\bm{f}(\z; \w): \mathds{F} \rightarrow \mathds{X}$ which receives as argument a random vector $\z \in \mathds{F}$ with known and easy to sample PDF $p(\z)$. Here, we refer to $\z$ as latent vector which belongs to a feature space $\mathds{F}$. Moreover, let $\bm{f}(\z; \w)$ be parameterized by deterministic vector $\w \in \mathds{W}$. Even though $\bm{f}$ is a function with deterministic parameters, $f(\z; \w)$ is a random vector in $\mathds{X}$ because $\z$ is random. We want to replace $\bm{f}(\z; \w)$ by a deep neural network which can be trained (determine the parameters $\w$) such that we can sample $\z \sim p(\z)$ and generate samples $\widehat{\x} \sim p(\x | \z; \w)$ which are likely to resemble samples from $p(\x)$. There are several generative models described in the machine learning literature such as restricted Boltzmann machines, DBNs, GANs, VAEs among others. Here, we focus our attention to a specific model based on convolutional neural network \citep{lecun:89} and VAE \citep{kingma:13a}. An overview about generative models in the context of deep learning methods is presented in \citep[Chap.~20]{goodfellow:16bk}. Before we introduce the proposed method, we briefly review the concepts of autoencoders, VAE and convolutional layers.

\subsection{Autoencoders}

\begin{figure}
\centering
	\includegraphics[width=0.85\linewidth]{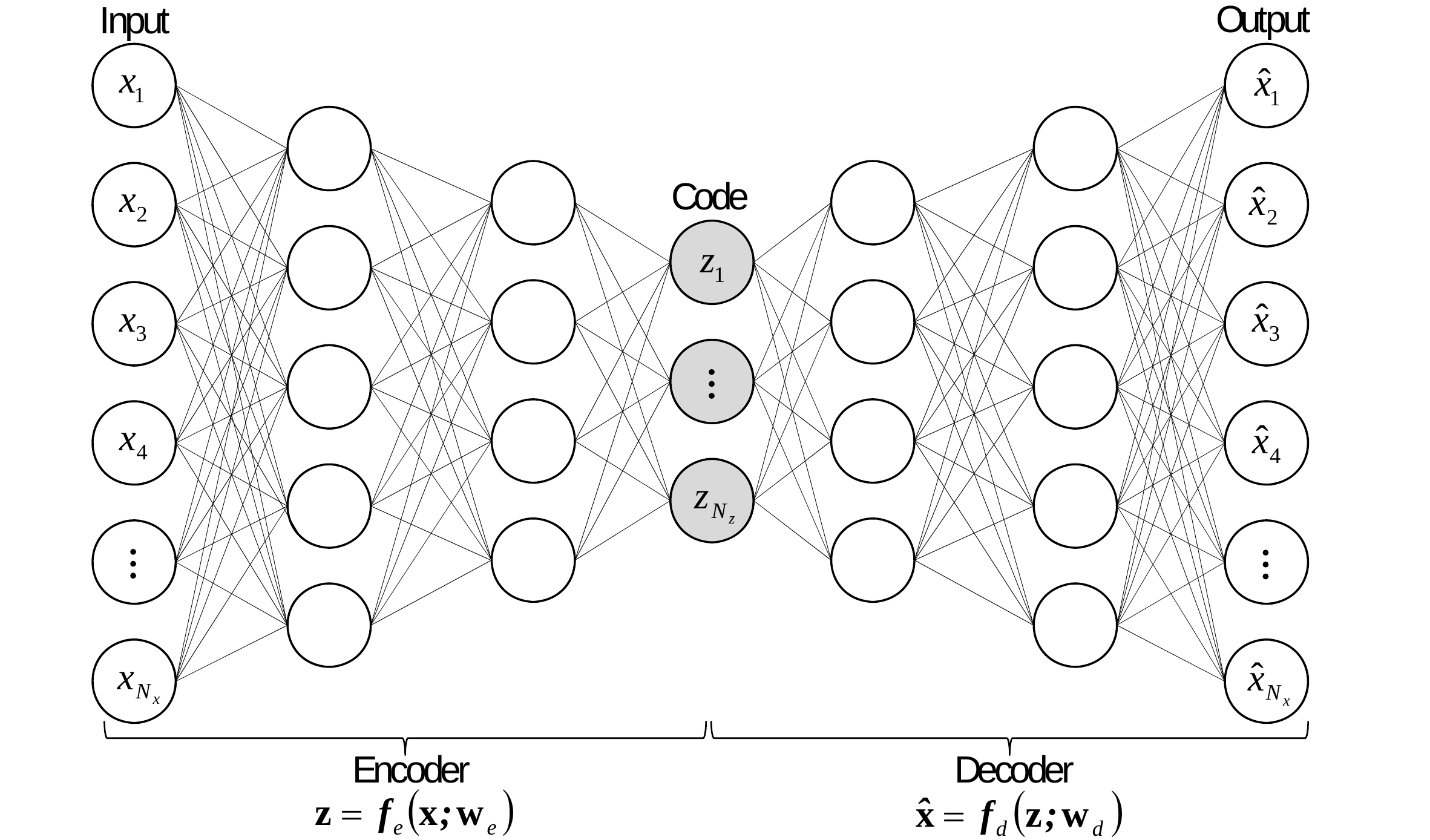}
\caption{Schematic architecture of an standard deep autoencoder with six fully connected layers.}
\label{Fig:AE}
\end{figure}

Autoencoder is an unsupervised neural network trained to learn complex data representations. The typical applications of autoencoders include data compression and noise removal. However, especially in the last decade, autoencoders become widely used as building blocks of deep generative models \citep{goodfellow:16bk}. Figure~\ref{Fig:AE} illustrates a standard deep autoencoder network composed by six fully-connected layers. The first three layers (encoder) are responsible for mapping the input space to a feature space, $\bm{f}_e(\x;\w_e): \mathds{X} \rightarrow \mathds{F}$. The last three layers (decoder) correspond to the inverse mapping $\bm{f}_d(\z; \w_d): \mathds{F} \rightarrow \mathds{X}$. The central layer is called code. The training process consists of minimizing a loss function that measures the dissimilarity between $\x$ and $\widehat{\x}=\bm{f}_d(\bm{f}_e(\x; \w_e); \w_d)$, for example, the mean square error. After training, the autoencoder is able to represent (encode) the most important features of $\x$ in $\z$. When the decoder function is linear and the loss function is the mean square error, the autoencoder learns to span the same subspace of PCA \citep{goodfellow:16bk}. Hence, autoecoders with nonlinear encoder and decoder functions may be interpreted as nonlinear generalizations of PCA \citep{deng:17a}.

\subsection{Variational Autoencoders}

\begin{figure}
\centering
	\includegraphics[width=0.85\linewidth]{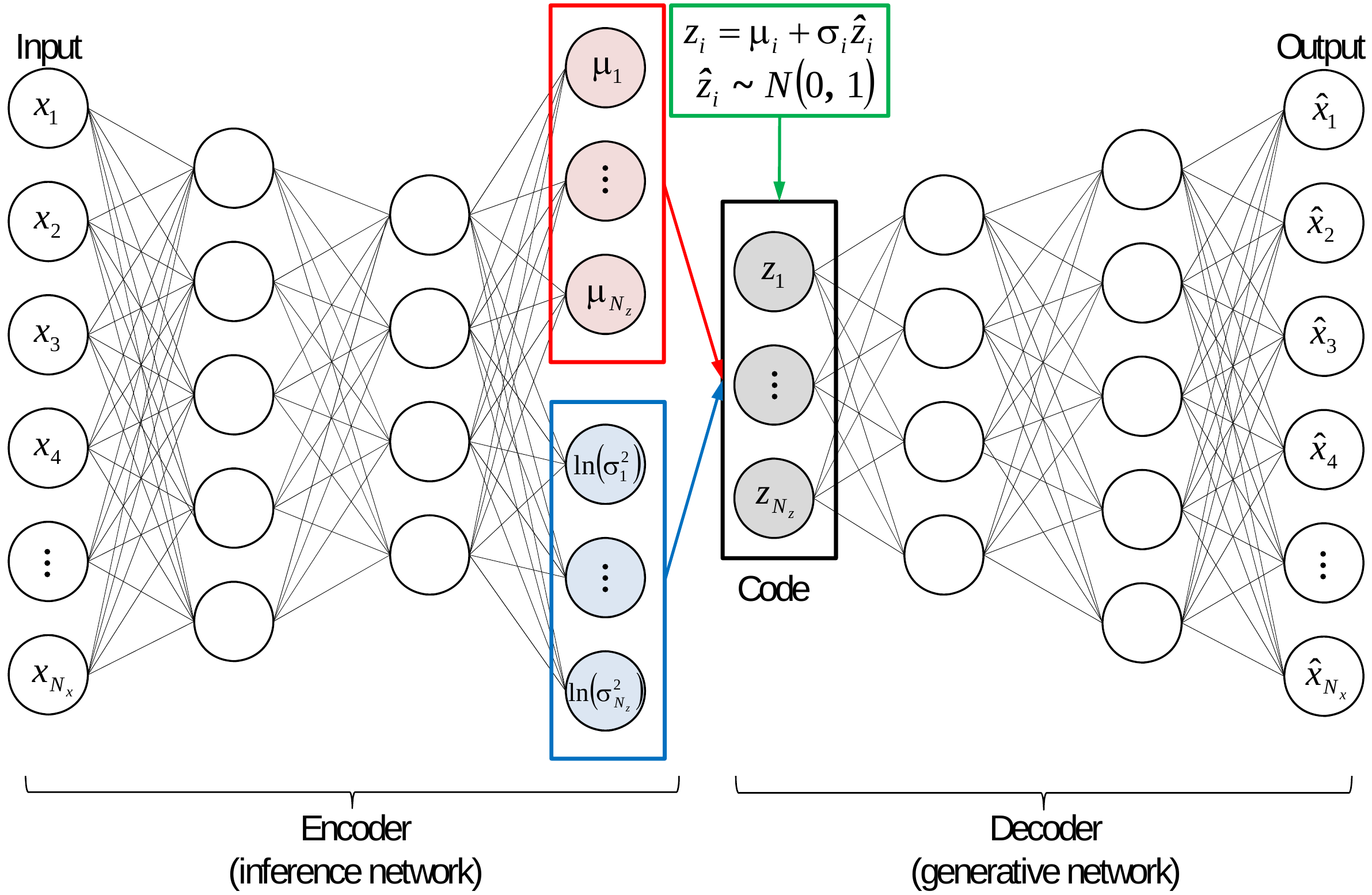}
\caption{Schematic architecture of a VAE.}
\label{Fig:VAE}
\end{figure}

A VAE is similar to a standard autoencoder in the sense that it is composed by an encoder and a decoder network. However, unlike standard autoencoders, a VAE has an extra layer responsible for sampling the latent vector $\z$ and an extra term in the loss function that forces to generate the latent vector with approximately a specified distribution, $p(\z)$, usually assumed a standard Gaussian, $\mathcal{N}(\mathbf{0},\mathbf{I})$. This extra term corresponds to the Kullback-Liebler divergence which measures how closely the distribution of the encoded latent vectors $p(\z|\x)$ is from the desired distribution $p(\z)$, i.e.,

\begin{equation}\label{Eq:VAE}
  \mathcal{L}(\x) =  \mathcal{L}_{\textrm{RE}}(\x) + \lambda \mathcal{D}_{\textrm{KL}} \left( p(\z|\x) \| p(\z) \right),
\end{equation}
where $\mathcal{L}(\x)$ is the total loss function. $\mathcal{L}_{\textrm{RE}}(\x)$ is the reconstruction error. Here, we use the binary cross-entropy function given by

\begin{equation}\label{Eq:BCE}
\mathcal{L}_{\textrm{RE}}(\x)= - \frac{1}{N_x}\sum_{i=1}^{N_x}\left[x_i \ln(\widehat{x}_i) + (1 - x_i) \ln(1-\widehat{x}_i)\right],
\end{equation}
where $x_i$ assumes values 0 or 1 and $\widehat{x}_i$ assumes continuous values in $(0,1)$. The term $\lambda$ in Eq. \ref{Eq:VAE} is a weight factor (for the test cases presented in this paper we use $\lambda = 1$). $\mathcal{D}_{\textrm{KL}}\left( p(\z|\x) \| p(\z) \right)$ is the Kullback–Leibler divergence from $p(\z|\x)$ to $p(\z)$. This term can be interpreted as a regularization imposed in the feature space. However, the term $ \mathcal{D}_{\textrm{KL}} \left( \cdot \right)$ in Eq.~\ref{Eq:VAE} has a more theoretical basis and it is derived from a variational Bayesian framework \citep{kingma:13a}. For the case where  $ p(\z|\x) = \mathcal{N}([\mu_1, \ldots, \mu_{N_z}]\trp, \textrm{diag}[\sigma_1^2, \ldots, \sigma^2_{N_z} ]\trp)$ and $p(\z) = \mathcal{N}(\mathbf{0},\mathbf{I})$ the Kullback-Leibler divergence becomes

\begin{equation}
  \mathcal{D}_{\textrm{KL}} \left( p(\z|\x) \| p(\z) \right) = \frac{1}{2} \sum_{i=1}^{N_z} \left( \mu_i^2 + \sigma_i^2 - \ln \left(\sigma_i^2\right) - 1 \right),
\end{equation}
where $\mu_i$ and $\sigma_i$ are the $i$th components of the mean and standard deviation vectors. During training, instead of generating the latent vector $\z$, the encoder generates vectors of means, $\bm{\mu}$, and log-variance, $\bm{\ln(\sigma^2)}$. Then, the vector $\widehat{\z}$ is drawn from $\mathcal{N}(\mathbf{0},\mathbf{I})$ and rescaled to generate the latent vector $\z = \bm{\mu} + \bm{\sigma} \circ \widehat{\z}$, which goes in the decoder to generate a reconstructed vector $\widehat{\x}$. Note that the minimization of the loss function imposes $\widehat{\x}$ to be as close as possible to the input vector $\x$ while the term $\mathcal{D}_{\textrm{KL}} \left( p(\z|\x) \| p(\z) \right)$ pushes $\bm{\mu}$ and $\bm{\sigma}$ towards the zero and the unity vectors, respectively. After training, the decoder can be used to generate new realizations $\widehat{\x}$ by sampling $\z \sim \mathcal{N}(\mathbf{0},\mathbf{I})$. Conceptually, we are generating samples $\widehat{\x}$ from a distribution $p(\x|\z) = \mathcal{N}(\bm{f}_d(\z; \w_d), \gamma^2 \mathbf{I})$, which is a Gaussian with mean given by a trained decoder with parameters $\w_d$ and covariance equals to the identity multiplied by a scaling parameter $\gamma^2$  \citep{doersch:16a}. Figure~\ref{Fig:VAE} shows a VAE illustrating the main components. The encoder corresponds to an inference network and the decoder corresponds to a generative model. A detailed discussion about the principles behind VAE is presented in \citep{doersch:16a}.

\subsection{Convolutional Layers}

The neural networks illustrated in Figs.~\ref{Fig:AE} and \ref{Fig:VAE} are based on fully-connected layers, i.e., each neuron is connected to all neurons in the previous layer. Unfortunately, fully-connected networks do not scale well, i.e., the number of training parameters (weights and bias terms) increase dramatically when the size of the input space is large, which is the case of facies realizations where the number of gridblocks can be easily on the order of hundreds of thousands. This is one of the main limitations observed in our previous work with DBN \citep{canchumuni:18a}. For this reason, in this work we resort to convolutional neural networks \citep{lecun:89} to construct the encoder and decoder of our VAE network. These networks gained significant attention in the deep learning area after the very successful application in the ImageNet image classification challenge \citep{krizhevsky:12a}. Convolution neural networks are specialized in data with grid structure such as images and time series \citep{goodfellow:16bk}. Usually each layer of a convolutional network consists of a sequence of convolutional operations, followed by the application of activation functions (detection stage) and pooling operations, which modify the size of the outputs for the next layer and reduce the number parameters and processing time in the next layers. The convolutional operations consist of a series of trainable filters (kernels) which are convolved with the input image to generate activation maps. These convolutions are essentially dot products between the entries of the kernel and the input at any position. Because the size of the kernels is much smaller than the dimension of the input data, the use of the convolutional layers reduces vastly the number of training parameters allowing deeper architectures. The activation functions are applied over the activation maps generated by the convolutional operations. The most common is the rectified linear units (ReLU) function. The pooling operation replaces the output by some statistic of the nearby outputs, typically the maximum output within a rectangular neighborhood (max-pooling). There are also hyperparameters which include the size and the number of kernels and the level of overlapping in the kernel (stride). For a detailed discussion about convolution networks we recommend \citep[Chap.~9]{goodfellow:16bk} and \citep{dumoulin:18a}.

\section{ES-MDA-CVAE}

\begin{figure}
\centering
	\includegraphics[width=\linewidth]{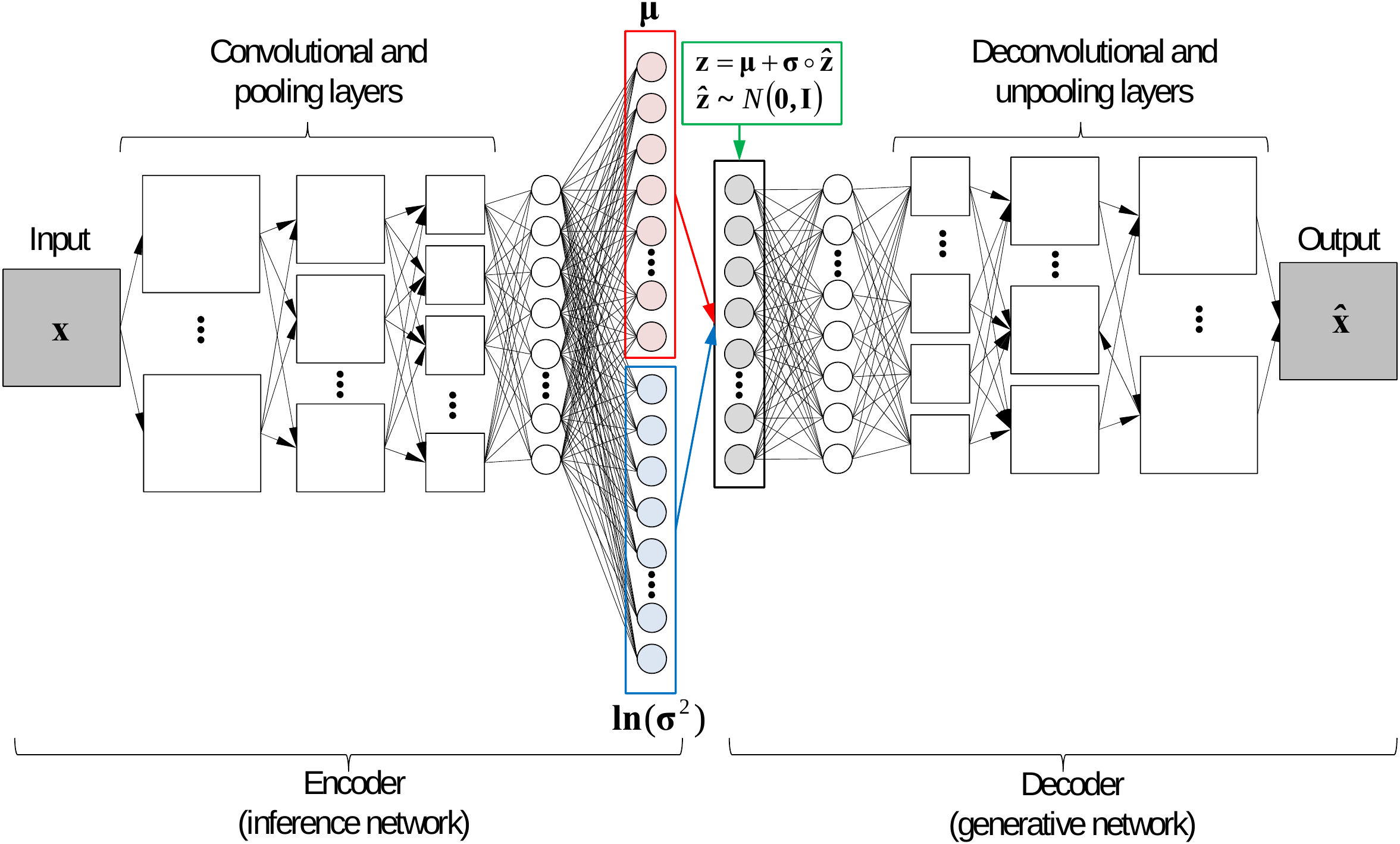}
\caption{Schematic architecture of a CVAE.}
\label{Fig:CVAE}
\end{figure}

Figure~\ref{Fig:CVAE} illustrates the final CVAE architecture with convolutional and fully-connected layers. We implemented the CVAE using \verb"Keras" \citep{keras:15} with \verb"TensorFlow" \citep{tensorflow:15} as backend engine. This network is trained using a large number of prior facies realizations, on the order of $\mathcal{O}(10^4)$ realizations. Note that no reservoir simulations are required in this process. After training, the CVAE is conceptually equipped to generate new realizations by simply sampling the random vector $\z \sim \mathcal{N}(\mathbf{0},\mathbf{I})$ and passing it to the decoder. At this point, the decoder works as a substitute model for the geostatistical algorithm used to construct initial realizations.

\begin{figure}
\centering
	\includegraphics[width=0.7\linewidth]{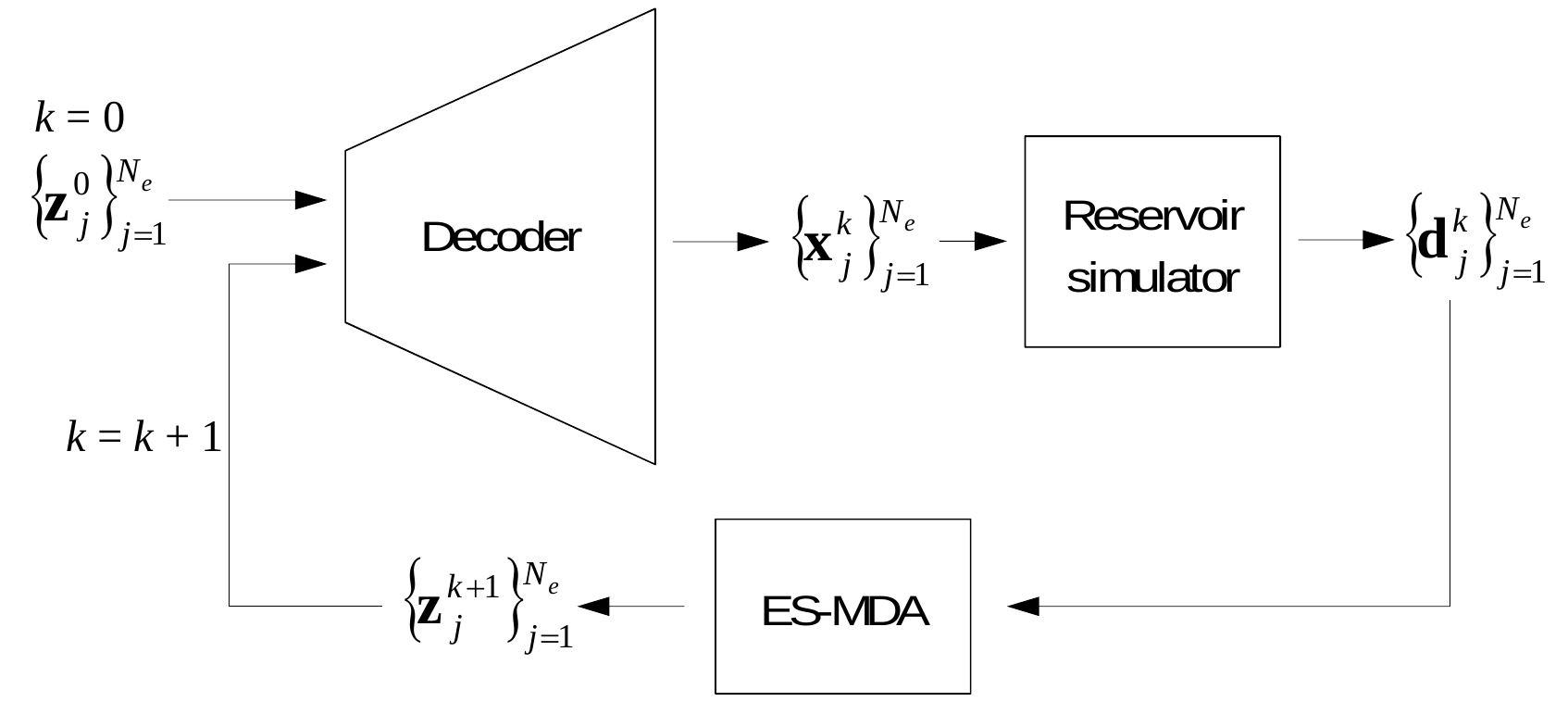}
\caption{Workflow of ES-MDA-CVAE}
\label{Fig:ES-MDA-CVAE}
\end{figure}

The data assimilation is done combining the trained decoder with the method ES-MDA. Essentially, we use ES-MDA to update an ensemble of realizations of the latent vector $\z$ to account for reservoir data and use the decoder to reconstruct the corresponding facies models. Here, we refer to this procedure as ES-MDA-CVAE. Figure~\ref{Fig:ES-MDA-CVAE} illustrates this workflow. The data assimilation stars with a set of prior realizations of the latent vector, denoted as $\{\z^0_j\}_{j=1}^{N_e}$ in Fig.~\ref{Fig:ES-MDA-CVAE}, where $N_e$ is the number of ensemble members. These prior latent vectors can be generated by sampling $\mathcal{N}(\mathbf{0},\mathbf{I})$ or being the result of the encoder for a set of $N_e$ prior facies realizations generated with geostatistics, which is the option adopted in the cases presented in this paper. The ensemble of latent vectors is used in the decoder to generate an ensemble of facies $\{\x^k_j\}_{j=1}^{N_e}$ which goes in the reservoir simulator to compute an ensemble of predicted data $\{\dsim^k_j\}_{j=1}^{N_e}$. The ES-MDA updating equation is used to update $\{\z^k_j\}_{j=1}^{N_e}$ and the process continue until the number of data assimilation iterations is achieved. Because process requires $N_e$ reservoir simulations to computed the vectors of predicted data, which can be very time consuming depending on the size of the model, we limite $N_e$ on the order of $\mathcal{O}(10^2)$ realizations.

The resulting ES-MDA updating equation can be written as

\begin{equation}\label{Eq:ES-MDA}
  \z^{k+1}_j = \z^k_j + \C^k_{\z\dsim} \left(\C^k_{\dsim\dsim} + \alpha_k \Ce \right)\inv \left(\dobs + \e^k_j - \dsim^k_j\right),~~\text{for}~j=1,\ldots,N_e,
\end{equation}
where $\C^k_{\z\dsim}$ and $\C^k_{\dsim\dsim}$ are matrices containing the cross-covariances between $\z$ and predicted data $\dsim$ and auto-covariances of $\dsim$, respectively. Both matrices are estimated using the current ensemble. $\Ce$ is the data-error covariance matrix. $\dobs$ is the vector containing the observations and $\e^k_j$ is a random vector sampled from $\mathcal{N}(\mathbf{0}, \alpha_k\Ce)$, where $\alpha_k$ is the data-inflation factor. In a standard implementation of ES-MDA, Eq.~\ref{Eq:ES-MDA} is applied a pre-defined number of times, $N_a$ and the values of $\alpha_k$ should be selected such that $\sum_{k=1}^{N_a} \alpha_k\inv = 1$ \citep{emerick:13b}. Here, we wrote Eq.~\ref{Eq:ES-MDA} in terms of only the latent vector for simplicity. However, we can easily introduce more uncertainty parameters of the reservoir in the data assimilation by updating an augmented vector.

\section{Test Cases}

\subsection{Test Case 1}

\begin{figure}
  \centering
  \subfloat{
        \includegraphics[width=0.32\linewidth]{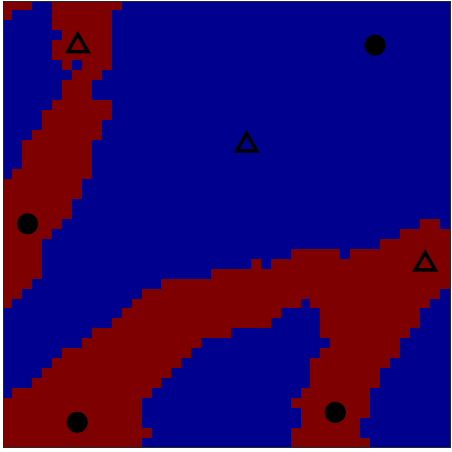}
  }
  \subfloat{
        \includegraphics[width=0.085\linewidth]{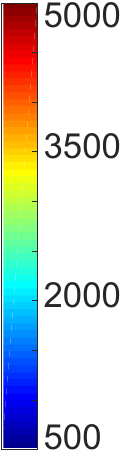}
  }
\caption{Reference permeability field (mD). Test case 1. Circles are oil producing wells and triangles are water injection wells.}
\label{Fig:Case1-True}
\end{figure}

The first test case corresponds to the same case used in \citep{canchumuni:18a}. This is a channelized facies model generated using the algorithm \verb"snesim" \citep{strebelle:02}. Figure~\ref{Fig:Case1-True} shows the reference (true) permeability field. The model has two facies: channels with constant permeability of 5,000~mD and background with permeability of 500~mD. The size of the model is $45 \times 45$ gridblocks, all gridblocks with $100~\text{ft} \times 100~\text{ft}$ and constant thickness of 50~ft.

\subsubsection{CVAE architecture and training}

The training set consists of 24,000 facies realizations generated using \verb"snesim" with the same training image of the reference model. We also use 6,000 additional realizations for validation. The architecture of the network is described in Table~\ref{Tab:CVAE-Case1} in the Appendix. The source code is available at \href{https://github.com/smith31t/GeoFacies_DL/}{\texttt{https://github.com/smith31t/GeoFacies\_DL}}. The input data of the CVAE are pre-processed facies images where each facies type corresponds to an color channel with the value one at the corresponding facies. This process is analogous to the pre-processing applied to color pictures where the image is divided in three color channels (red, green and blue). Essentially the encoder is composed of three convolutional layers followed by three fully-connected layers and one dropout layer \citep{srivastava:14a} to avoid overfitting. In the initial steps of the research, we tested different setups of the network, especially the dimension of the feature space. Because the encoder uses fully-connected layers to compute the latent vector, it is desirable to keep the size of this vector, $N_z$, as small as possible to reduce the computational requirements for training. Unfortunately, fully-connected layers are not efficiently parallelizable even using GPU. Our limited set of tests indicated that for the problems presented in this paper, we did not observe significant improvements for $N_z \geq 100$. Hence, we selected $N_z = 100$. The decoder has a mirrored architecture of the encoder with transposed-convolutional layers (often referred to as deconvolutional layers \citep{dumoulin:18a}). Before the last layer of the decoder, we introduced an up-sampling layer with bilinear interpolation to resize the output for the same size of the final model. Note that only the last layer has sigmoid activation function, which is used for classification of the facies type in each gridblock of the model.

\begin{figure}
  \centering
        \includegraphics[width=0.18\linewidth]{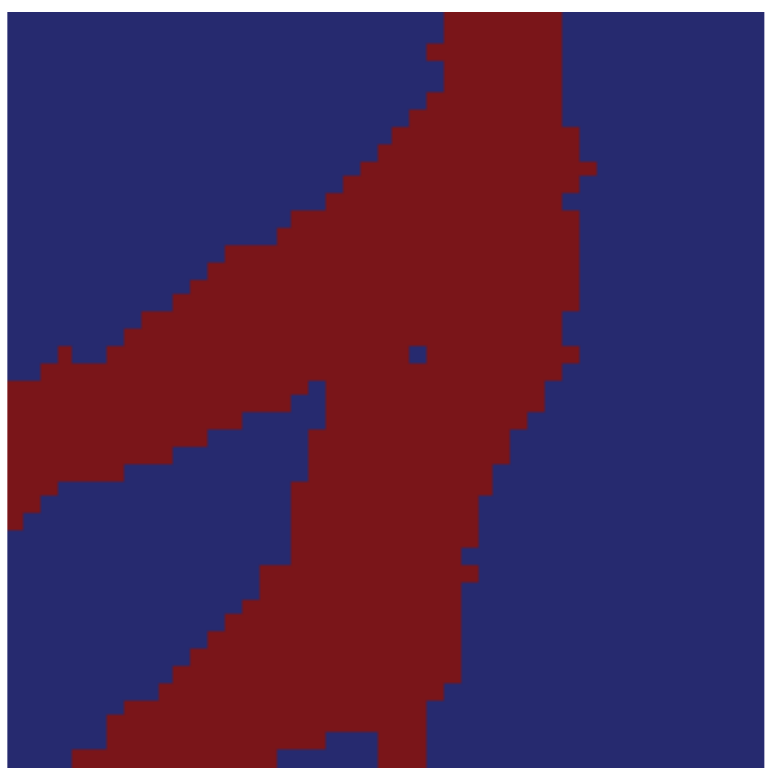}
        \includegraphics[width=0.18\linewidth]{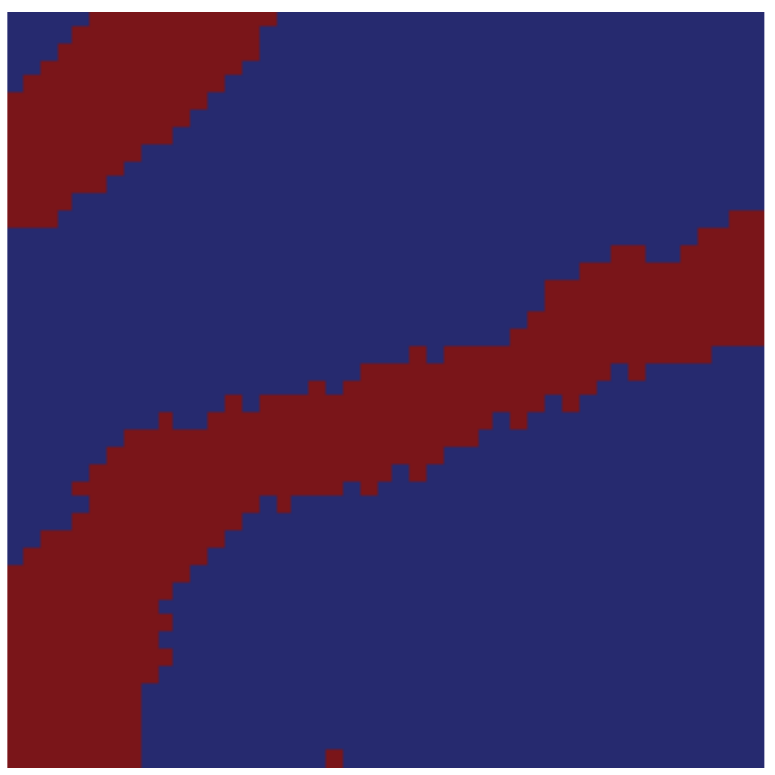}
  \subfloat[Input]{
        \includegraphics[width=0.18\linewidth]{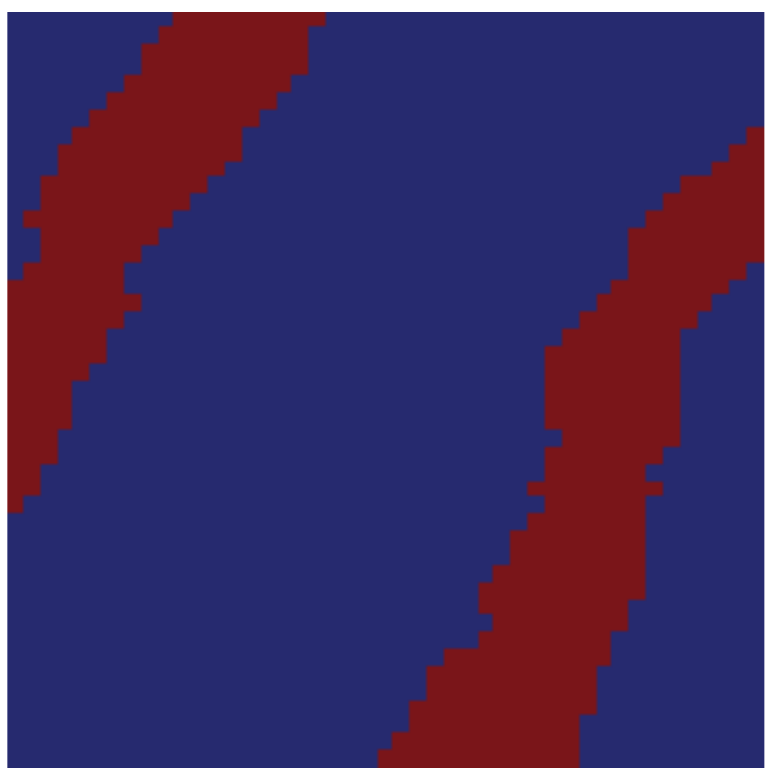}
  }
        \includegraphics[width=0.18\linewidth]{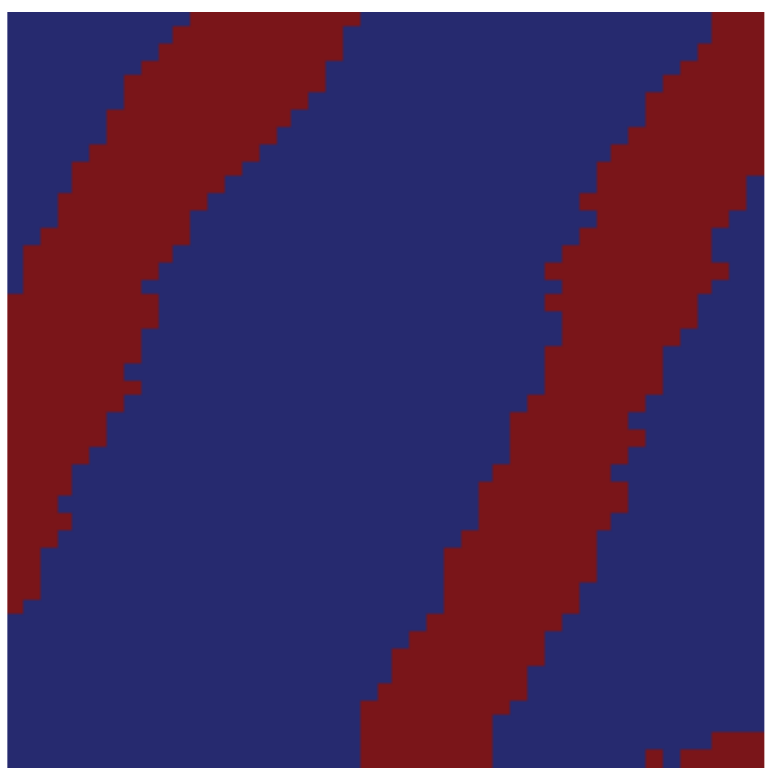}
        \includegraphics[width=0.18\linewidth]{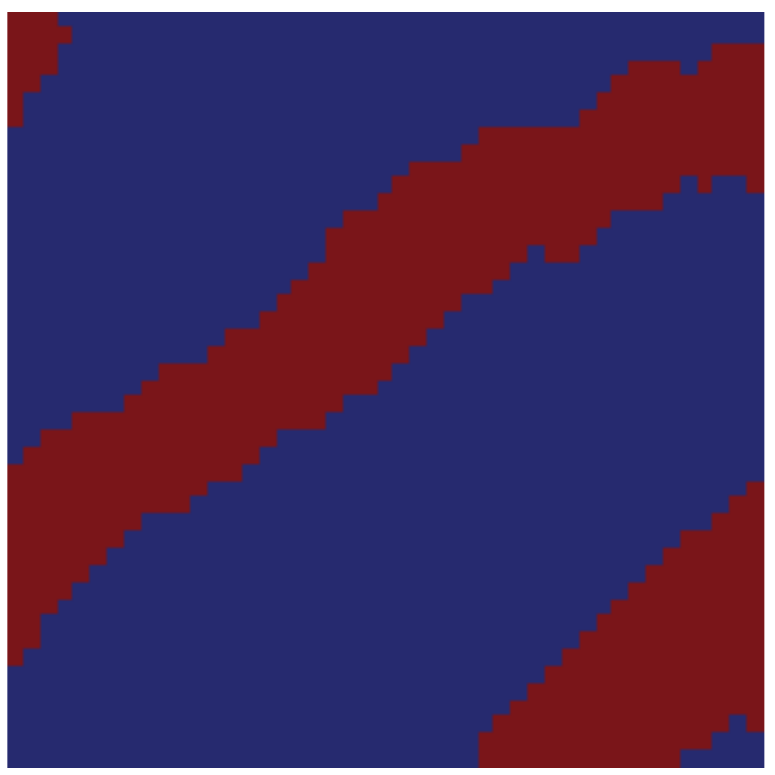}
\linebreak
        \includegraphics[width=0.18\linewidth]{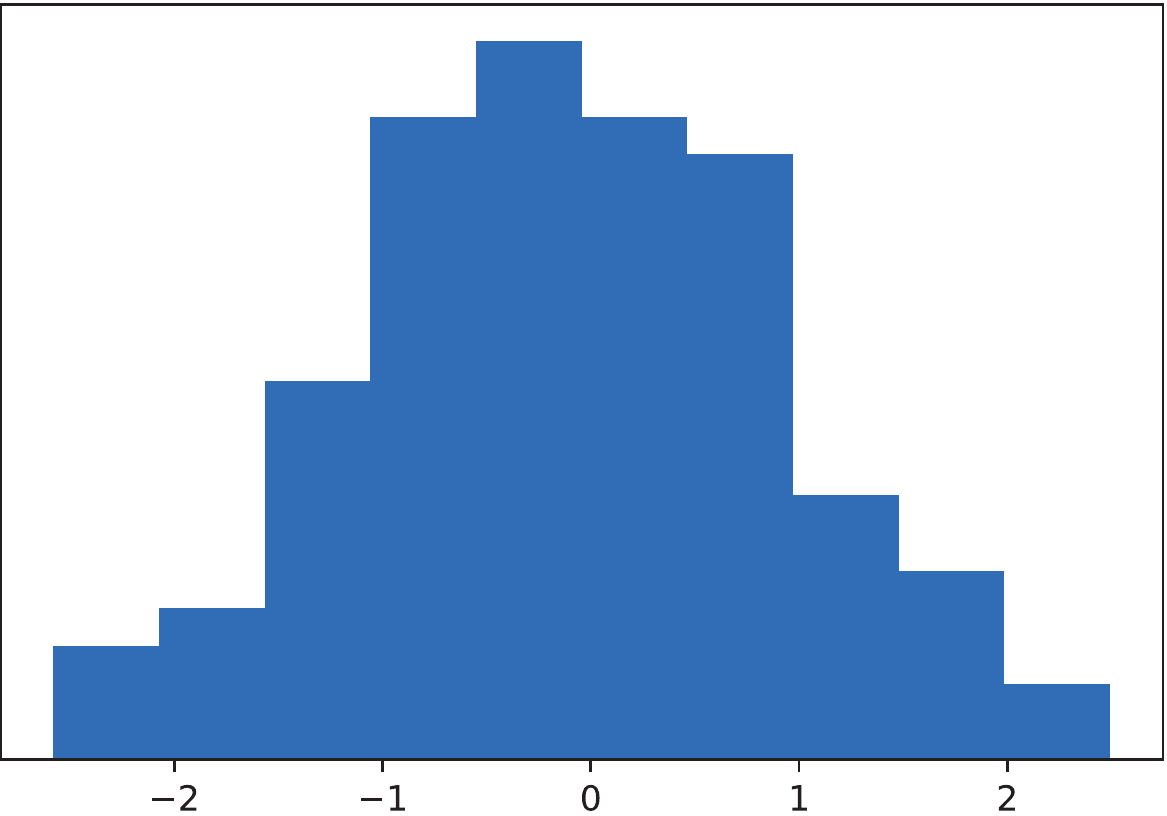}
        \includegraphics[width=0.18\linewidth]{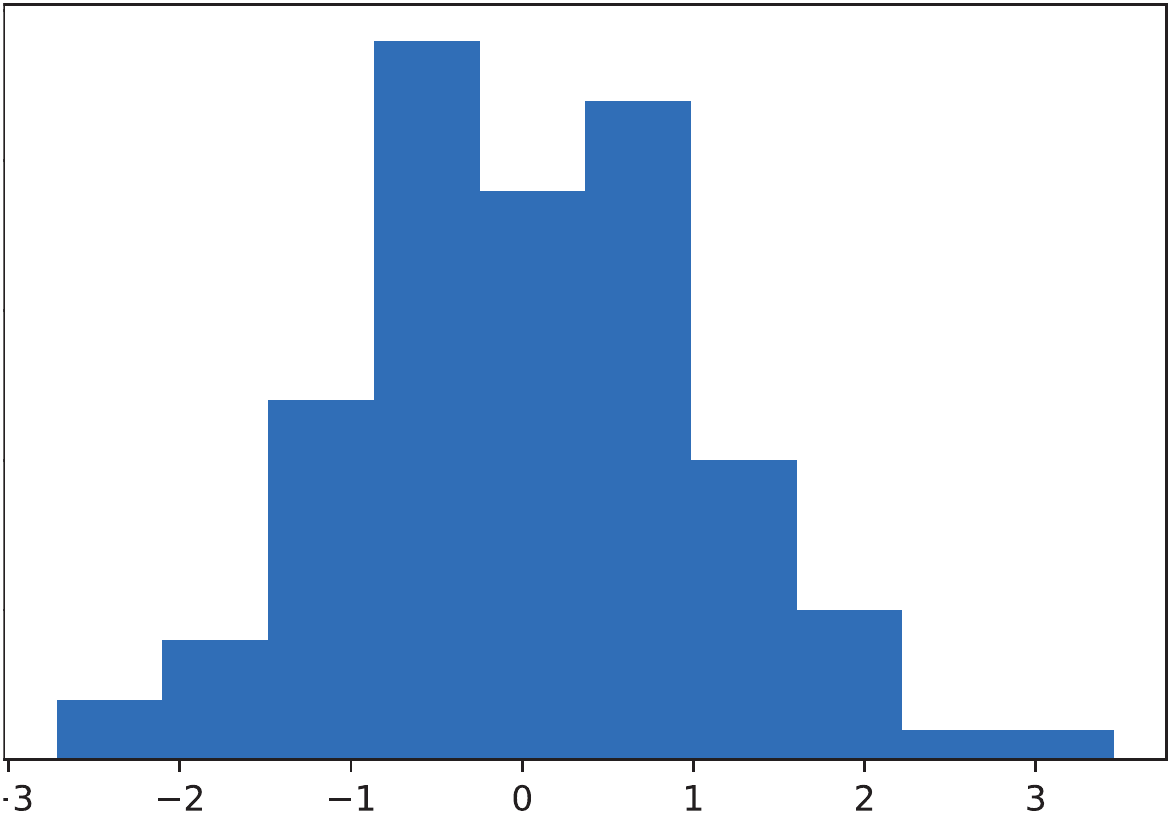}
  \subfloat[Latent]{
        \includegraphics[width=0.18\linewidth]{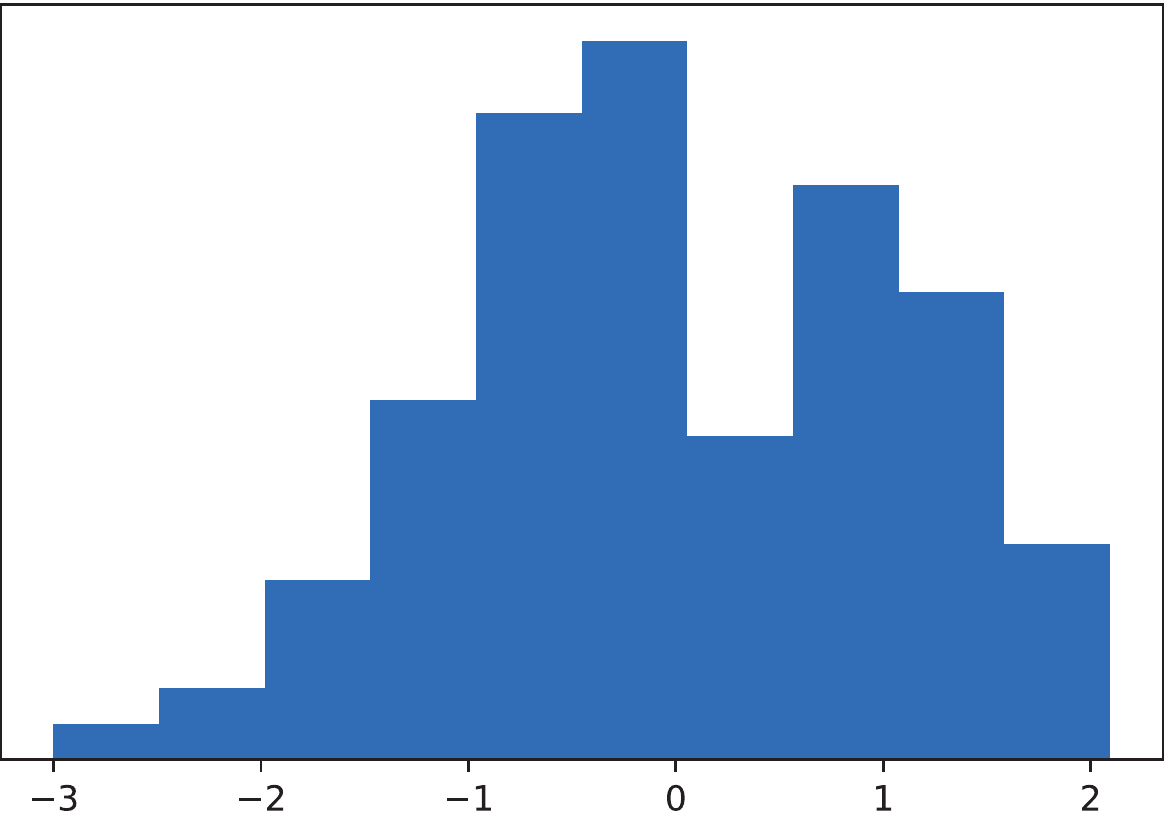}
  }
        \includegraphics[width=0.18\linewidth]{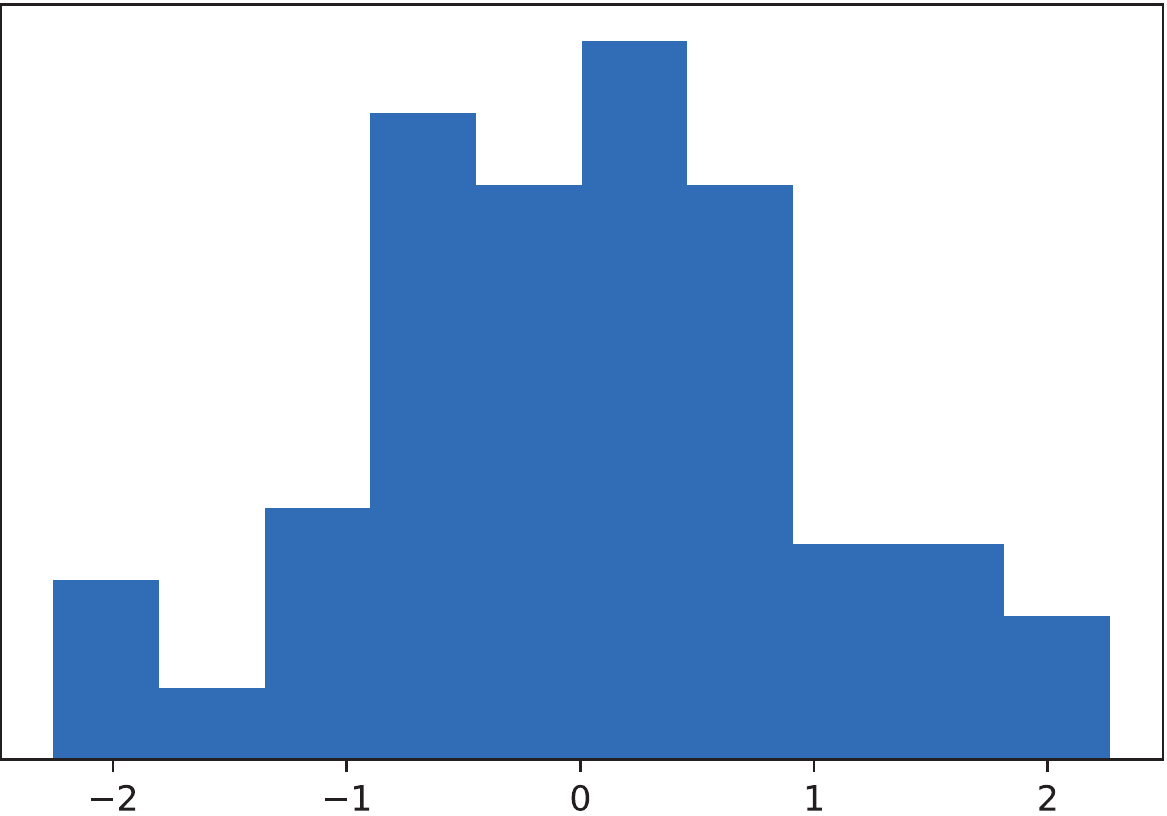}
        \includegraphics[width=0.18\linewidth]{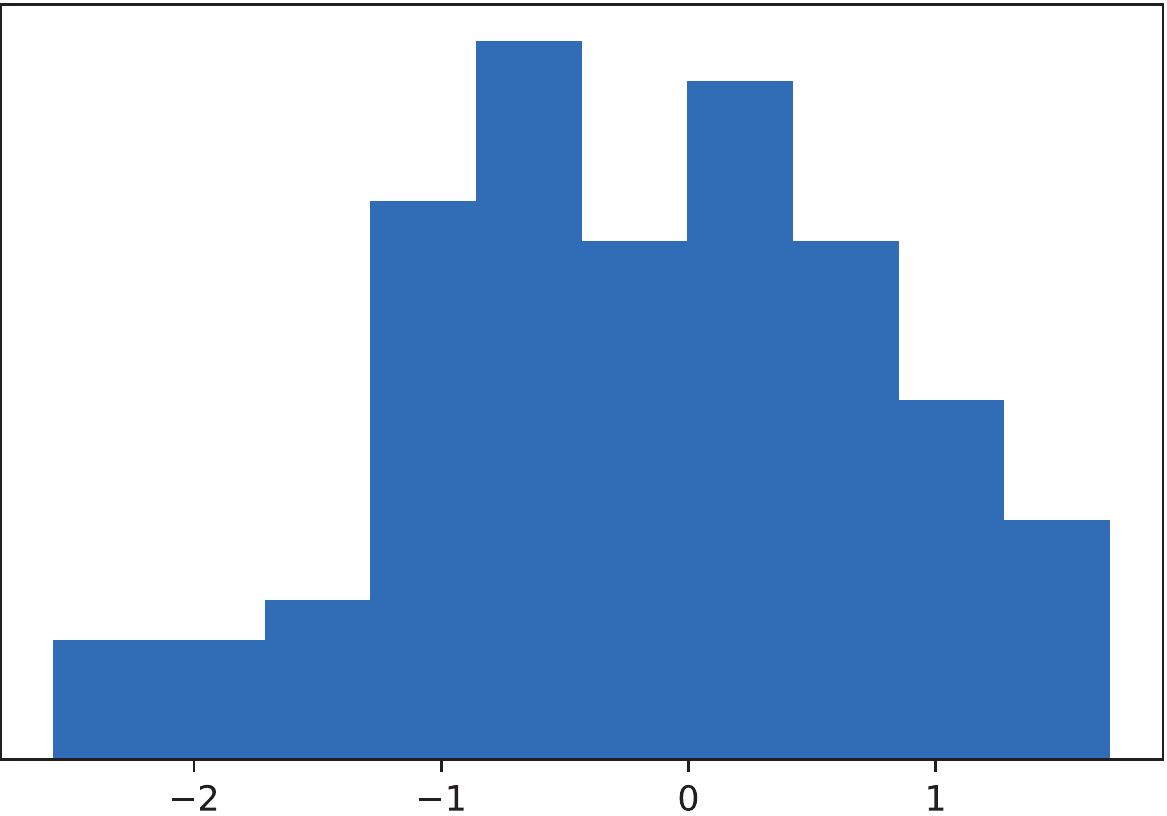}
\linebreak
        \includegraphics[width=0.18\linewidth]{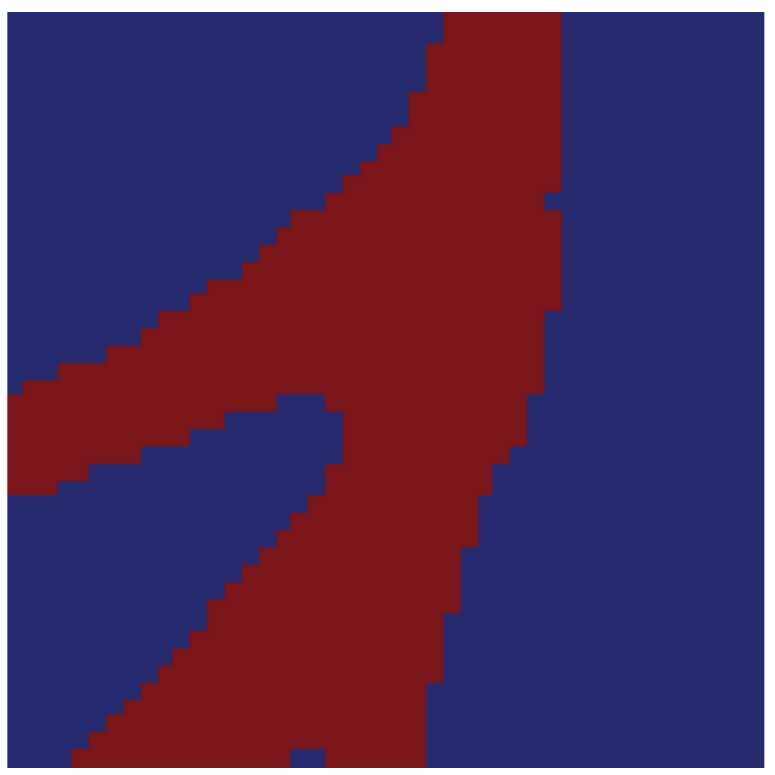}
        \includegraphics[width=0.18\linewidth]{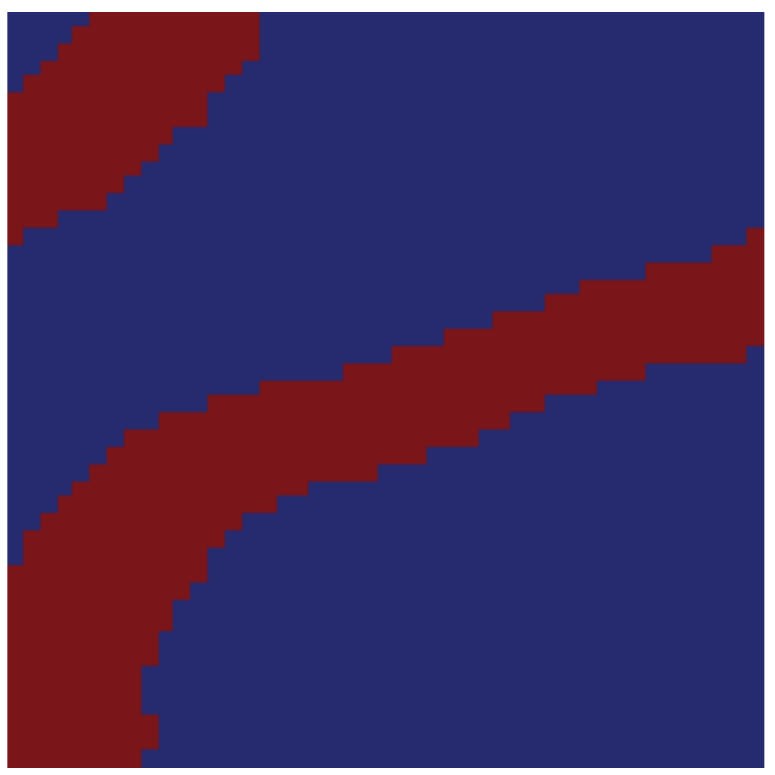}
  \subfloat[Reconstructed]{
        \includegraphics[width=0.18\linewidth]{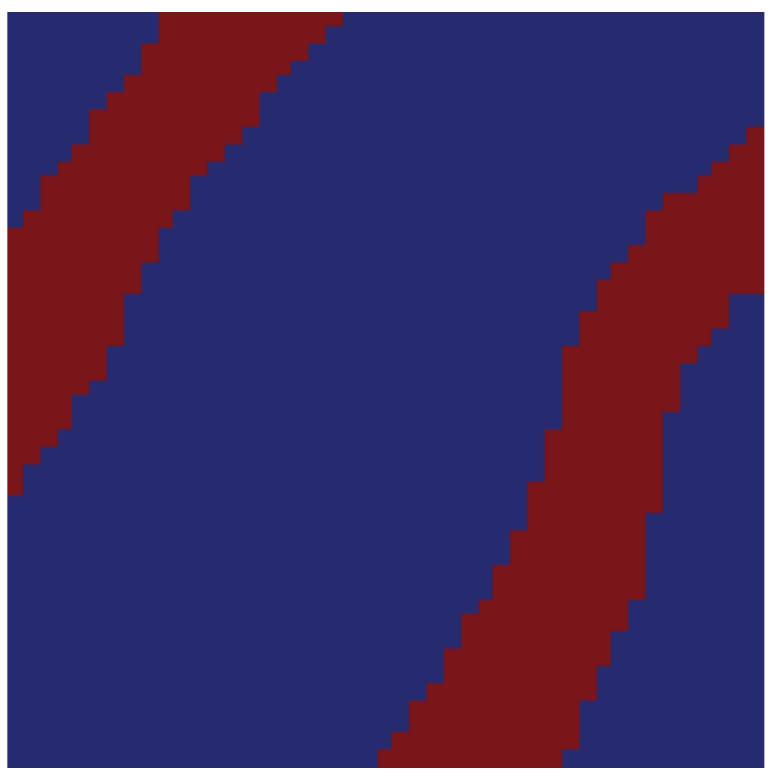}
  }
        \includegraphics[width=0.18\linewidth]{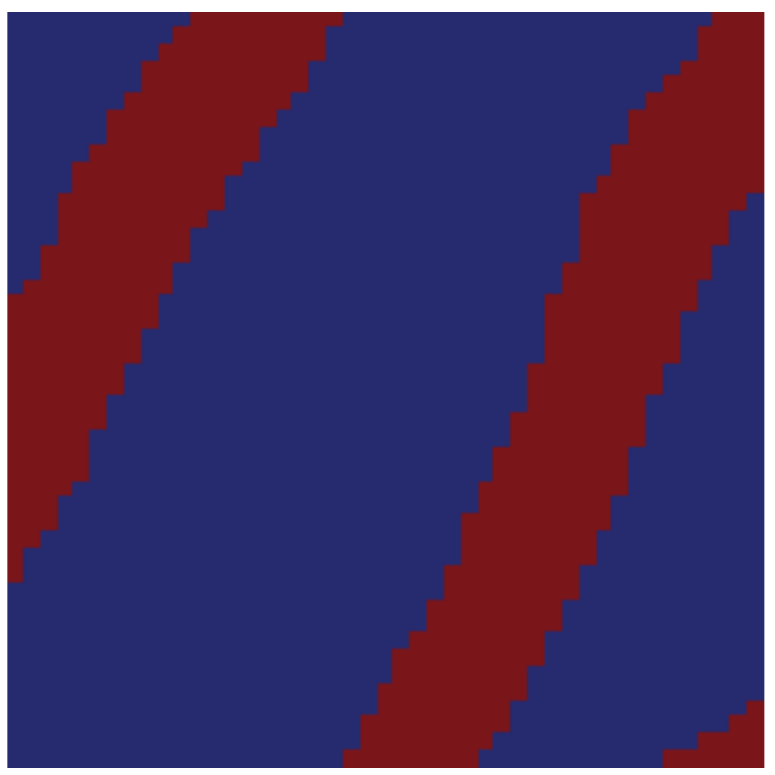}
        \includegraphics[width=0.18\linewidth]{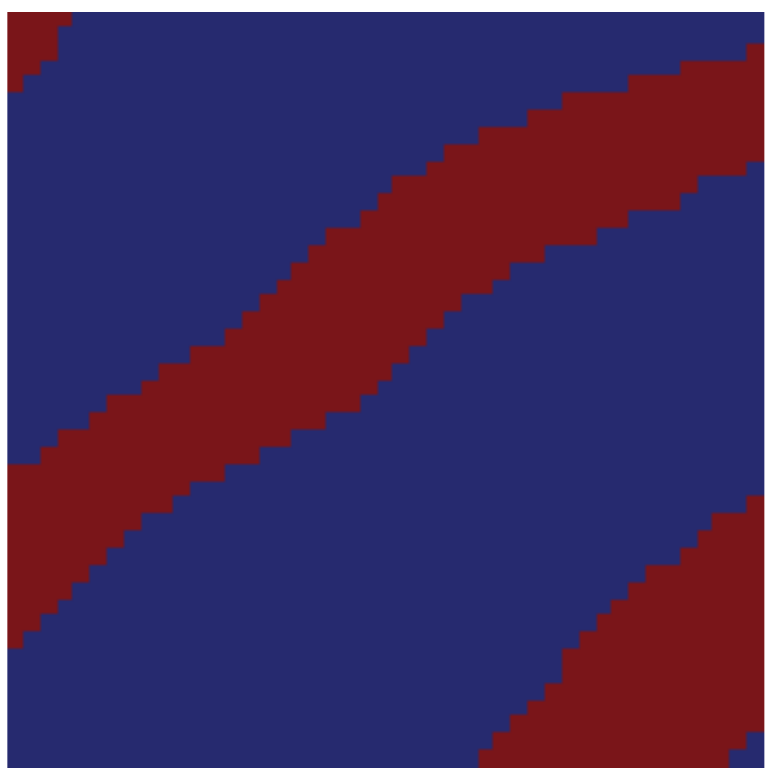}
\caption{Training process showing the first five realizations of the validation set and the corresponding histograms of the latent vector. Test case 1.}
\label{Fig:Case1-Training}
\end{figure}

The training required approximately 13 minutes in a cluster with four GPUs (NVIDIA TESLA P100) with 3584 cuda cores each. The final reconstruction accuracy for the validation set was 96.7\%. Figure~\ref{Fig:Case1-Training} shows the first five realizations of the validation set before and after reconstruction. The results in this figure show that the designed CVAE was able to successfully reconstruct the facies. This figure also shows the corresponding histograms of the latent vectors showing nearly Gaussian marginal distributions.

\subsubsection{Conditioning to facies data}

\begin{figure}
  \centering
  \includegraphics[width=1\linewidth]{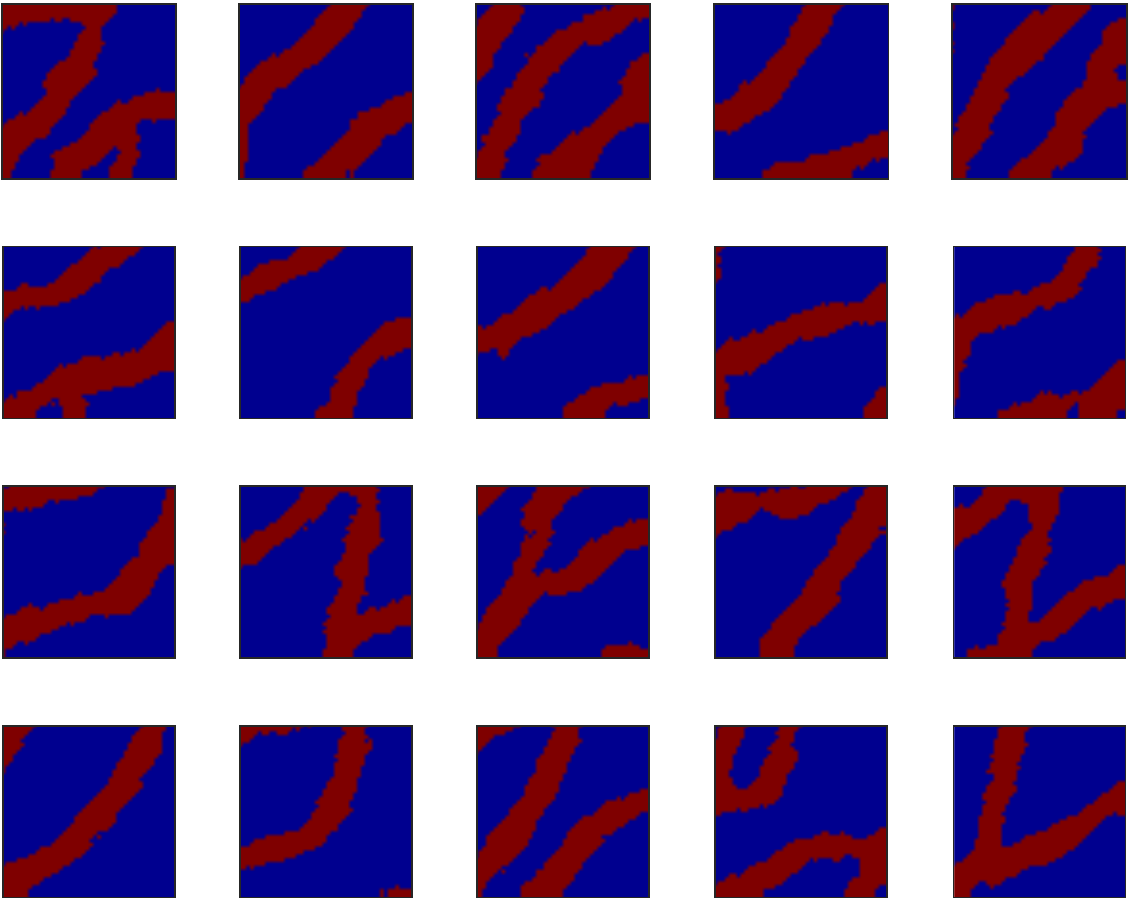}
\caption{First 20 prior realizations before assimilation of facies data. Test case 1.}
\label{Fig:Case1-Prior7HardData}
\end{figure}

\begin{figure}
	\centering
	\includegraphics[width=0.25\linewidth]{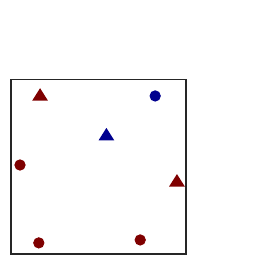}
	\caption{Seven hard data points using the generate conditional facies data. Test case 1.}
	\label{Fig:Case1-7HardData}
\end{figure}

\begin{figure}
  \centering
  \includegraphics[width=1\linewidth]{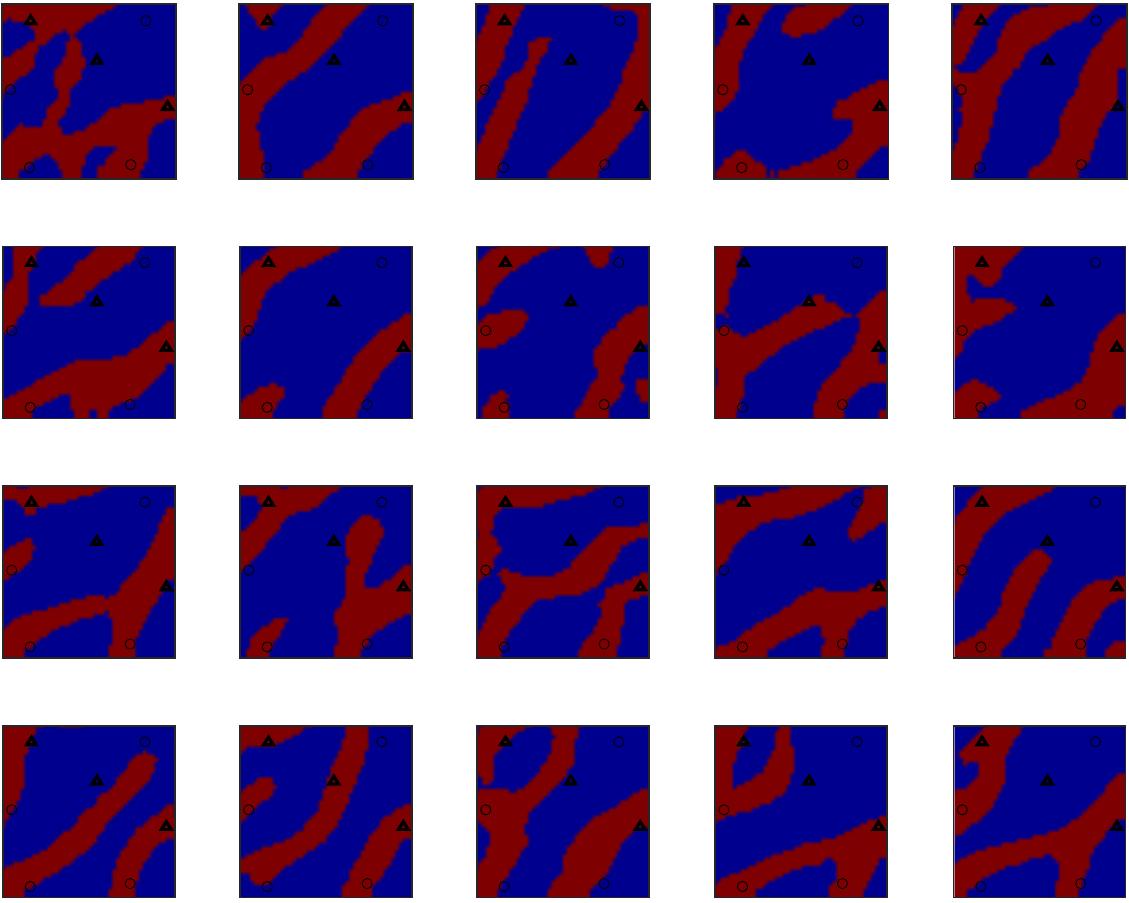}
\caption{First 20 posterior realizations after assimilation of facies data at seven well locations. Test case 1. The circles and triangles represent well locations with facies data.}
\label{Fig:Case1-Post7HardData}
\end{figure}

\begin{figure}
	\centering
	\includegraphics[width=0.25\linewidth]{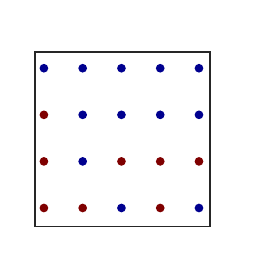}
	\caption{Twenty hard data points using the generate conditional facies data. Test case 1.}
	\label{Fig:Case1-20HardData}
\end{figure}

\begin{figure}
  \centering
  \includegraphics[width=1\linewidth]{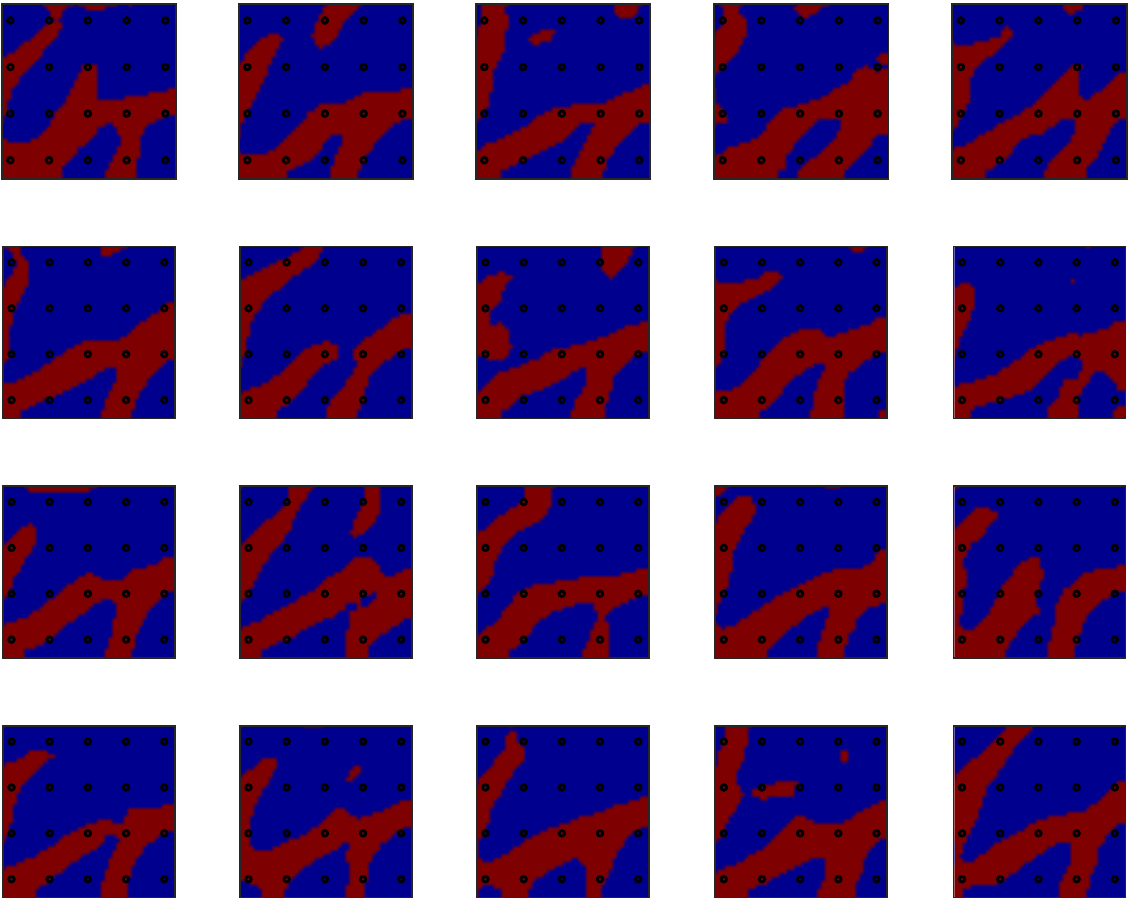}
\caption{First 20 posterior realizations after assimilation of facies data at 20 well locations. Test case 1. The circles represent the facies data locations.}
\label{Fig:Case1-Post20HardData}
\end{figure}

The facies realizations of the training and validation sets were generated without any hard data (facies type at well locations). However, in real-life applications, geological models are always constructed constrained to hard data. Our tests indicate that if we train the network with realizations conditioned to hard data, most of the reconstructed facies honor these data, but there is no guarantee. In fact, \citet{laloy:17b} reported that in one of their tests only 68\% of the realizations honor all nine hard data points imposed in the training set. For this reason, here we investigate the ability of the proposed ES-MDA-CVAE to condition the prior realizations to facies data. For this test, we used an ensemble of $N_e = 200$ prior realizations and $N_a = 4$ MDA iterations. We assumed a small value for the data-error variance of $\sigma_e^2 = 0.01$. Figure~\ref{Fig:Case1-Prior7HardData} shows the first 20 prior realizations. Figures~\ref{Fig:Case1-Post7HardData} and \ref{Fig:Case1-Post20HardData} show the corresponding realizations conditioned to seven (Fig.~\ref{Fig:Case1-7HardData}) and 20 (Fig.~\ref{Fig:Case1-20HardData}) hard data points, respectively. The results in these figures show that ES-MDA-CVAE was able to honor the facies type for all data points. The posterior realizations show well-defined channels, although we observe some ``broken'' channels. Nevertheless, the final realizations preserve reasonably well the main geological characteristics of the prior ones.


\clearpage

\subsubsection{Conditioning to production data}

\begin{figure}
  \centering
        \includegraphics[width=0.18\linewidth]{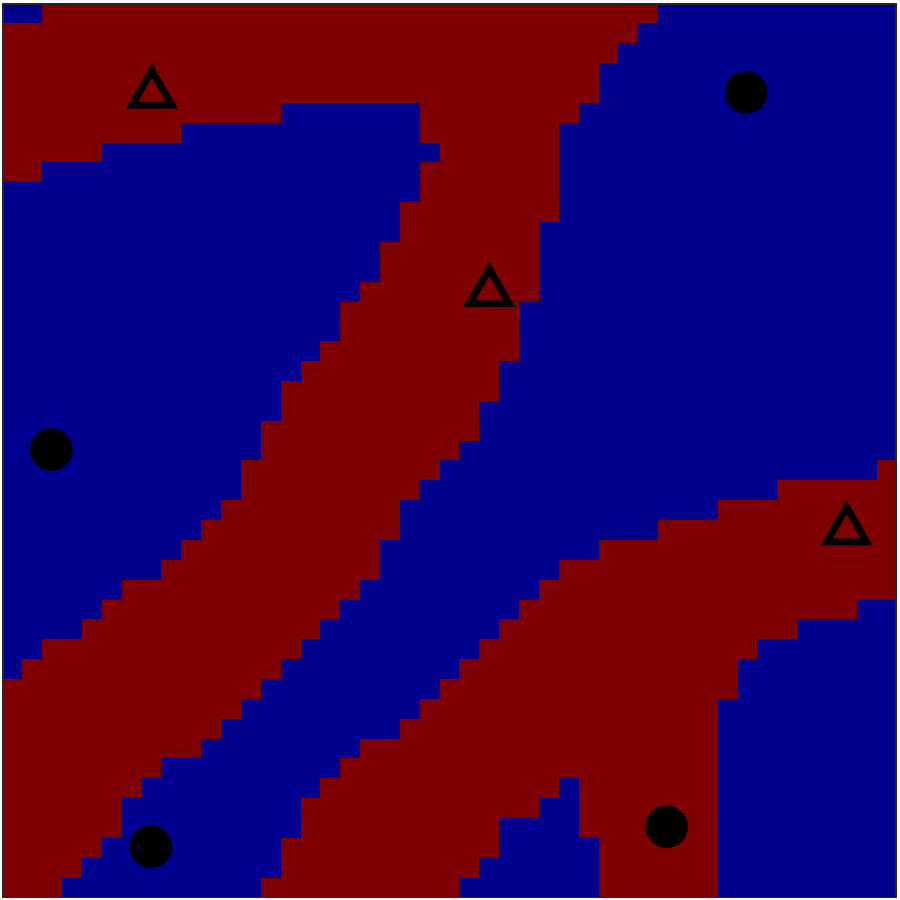}
        \includegraphics[width=0.18\linewidth]{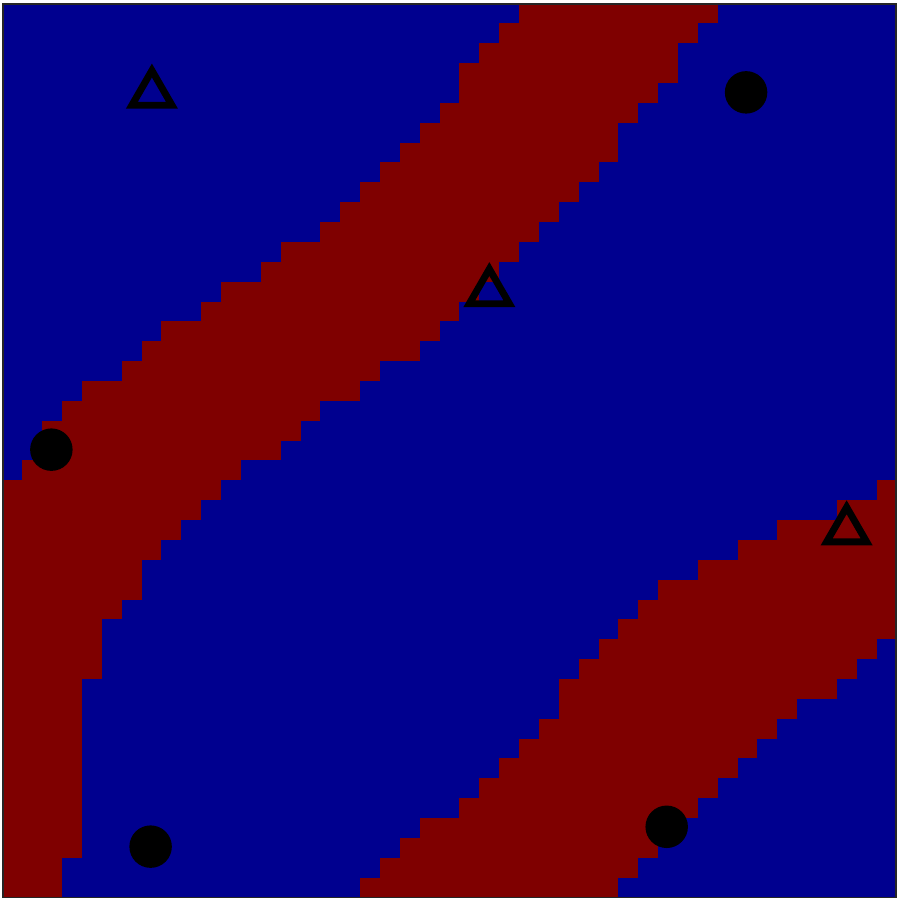}
  \subfloat[Prior]{
        \includegraphics[width=0.18\linewidth]{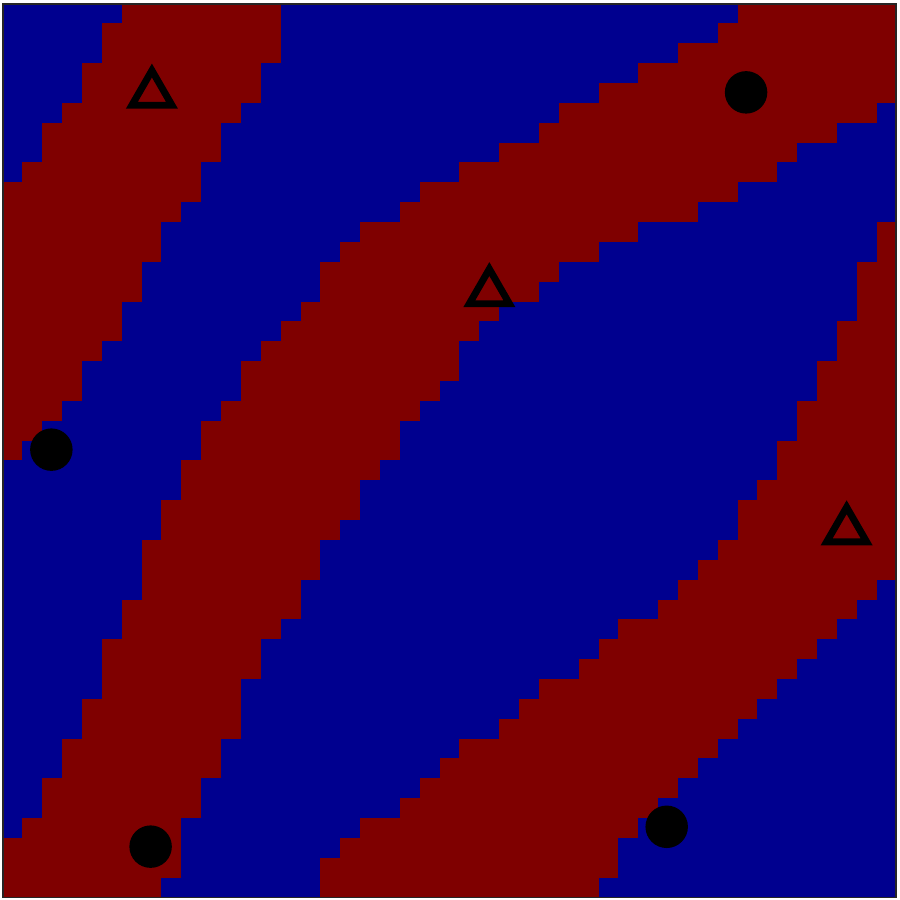}
  }
        \includegraphics[width=0.18\linewidth]{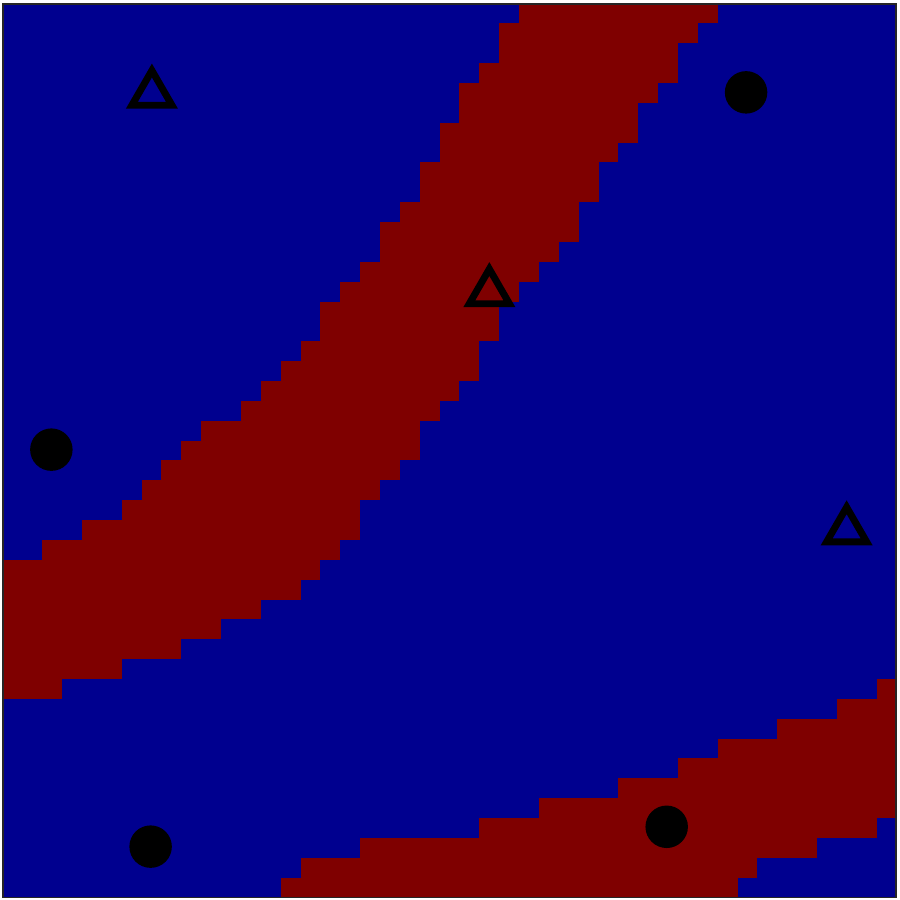}
        \includegraphics[width=0.18\linewidth]{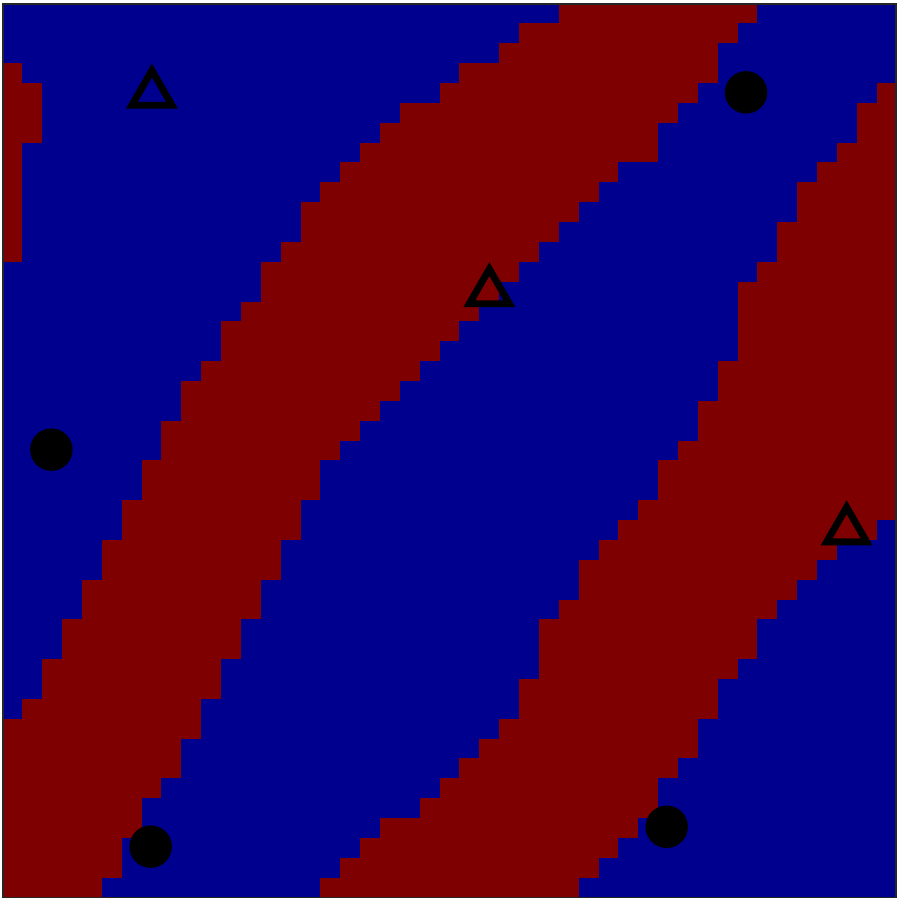}
\linebreak
        \includegraphics[width=0.18\linewidth]{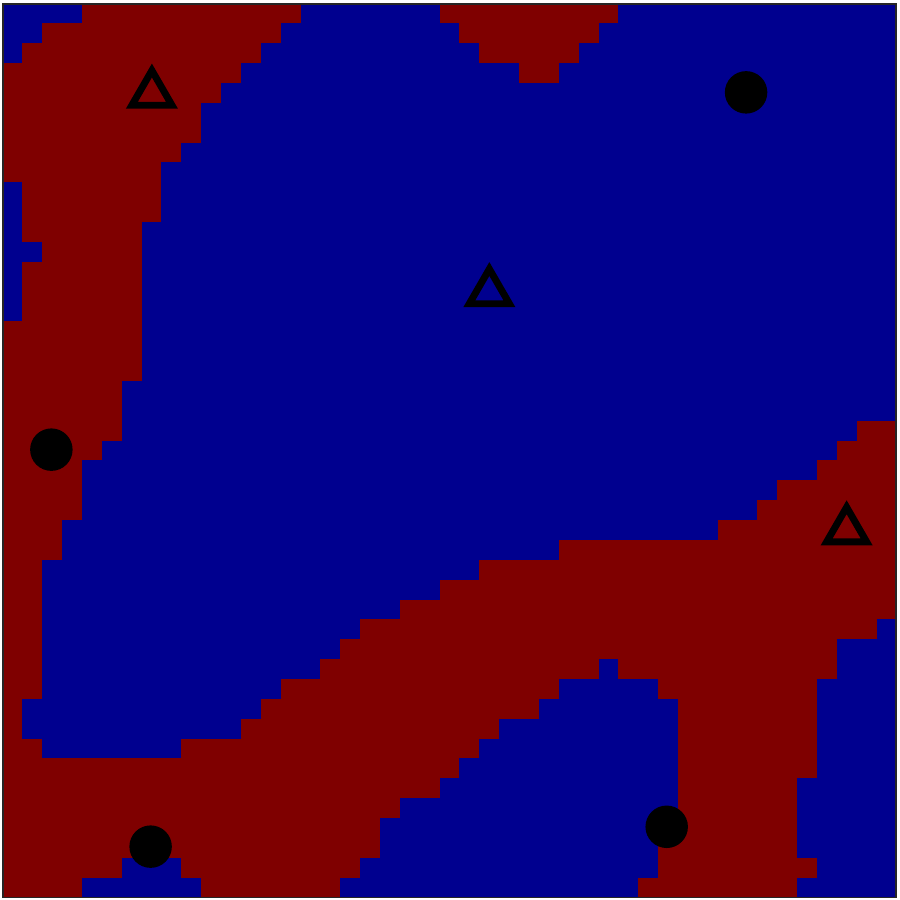}
        \includegraphics[width=0.18\linewidth]{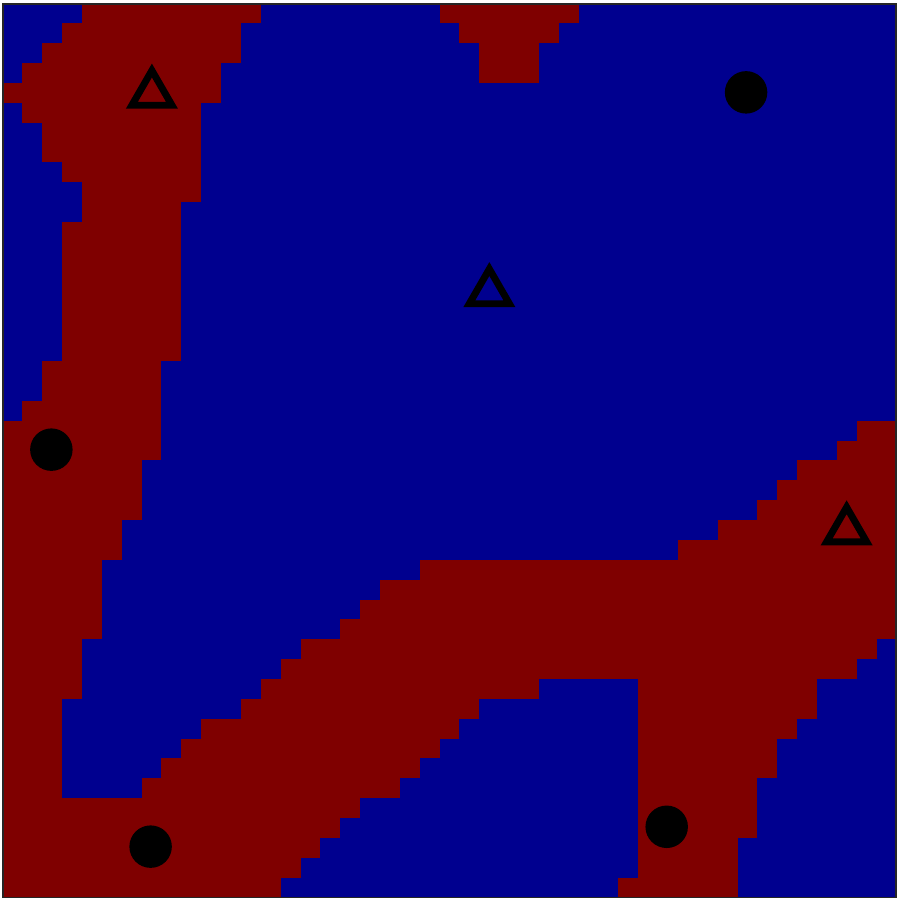}
  \subfloat[Post]{
        \includegraphics[width=0.18\linewidth]{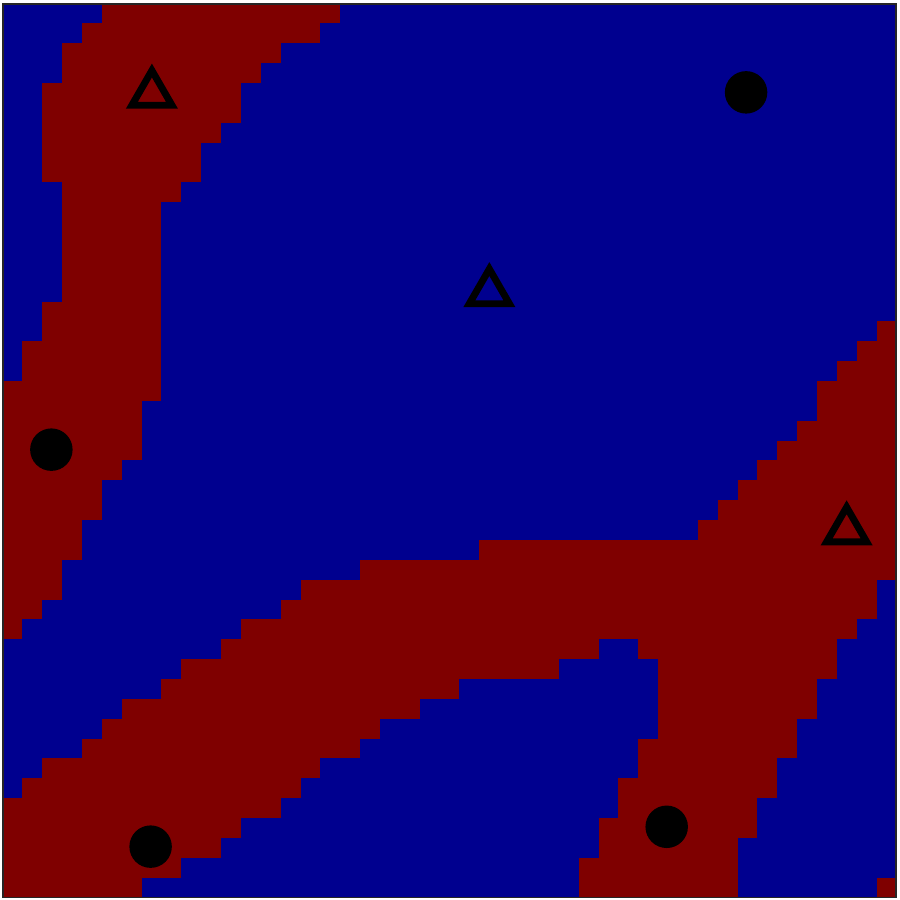}
  }
        \includegraphics[width=0.18\linewidth]{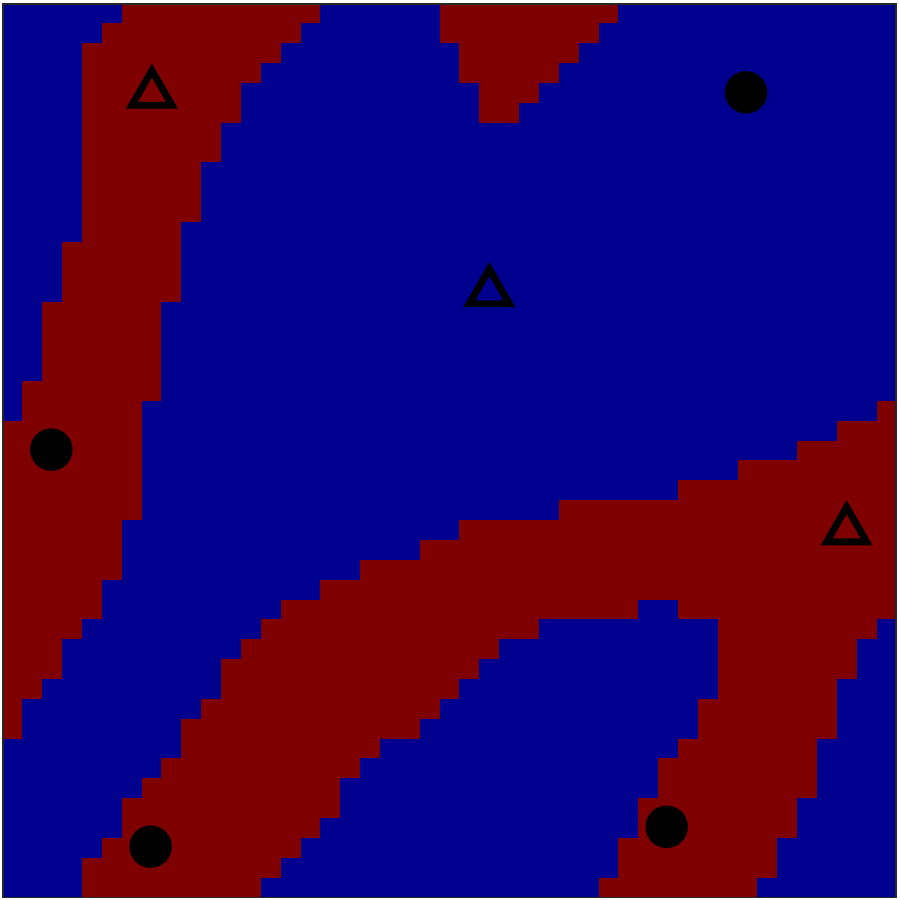}
        \includegraphics[width=0.18\linewidth]{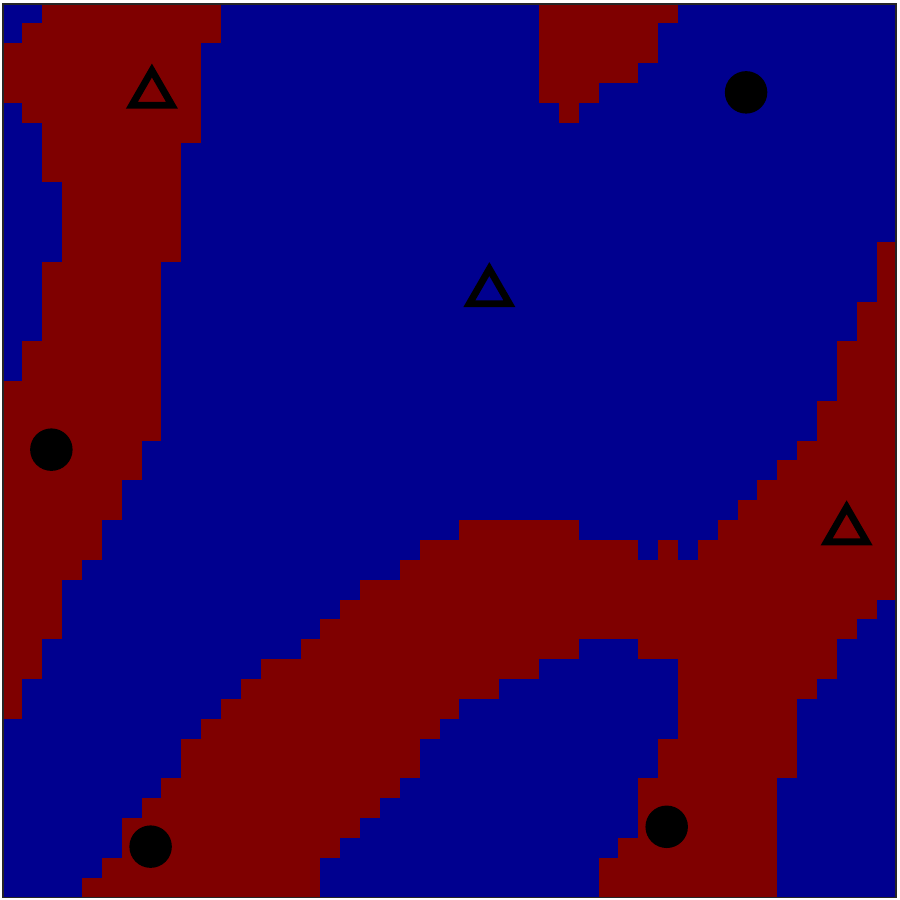}
\caption{First five prior and posterior realizations of permeability after assimilation of production data. Test case 1.}
\label{Fig:Case1-Realization}
\end{figure}

\begin{figure}
  \centering
  \subfloat[Well P1]{
        \includegraphics[width=0.4\linewidth]{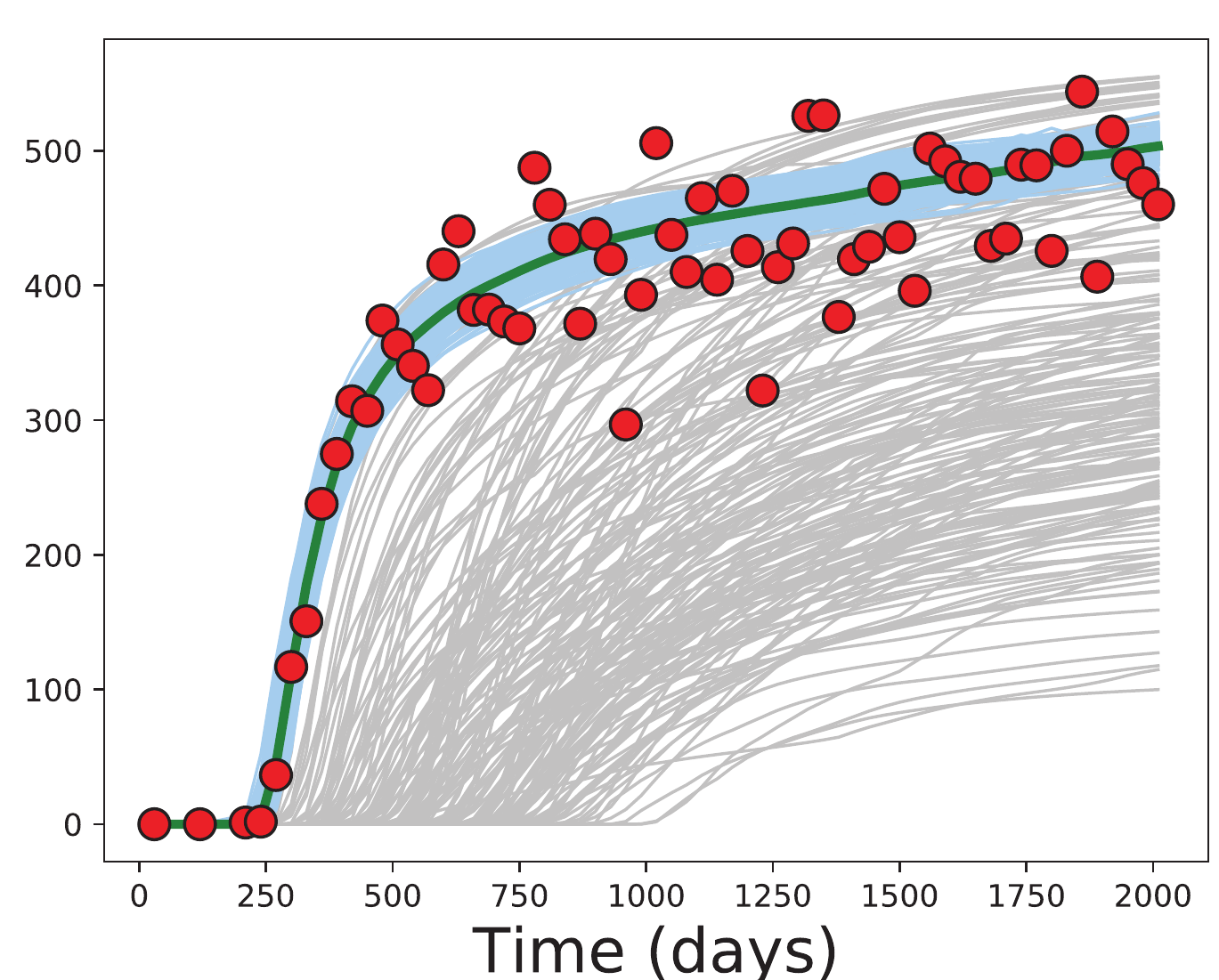}
  }
  \subfloat[Well P2]{
        \includegraphics[width=0.4\linewidth]{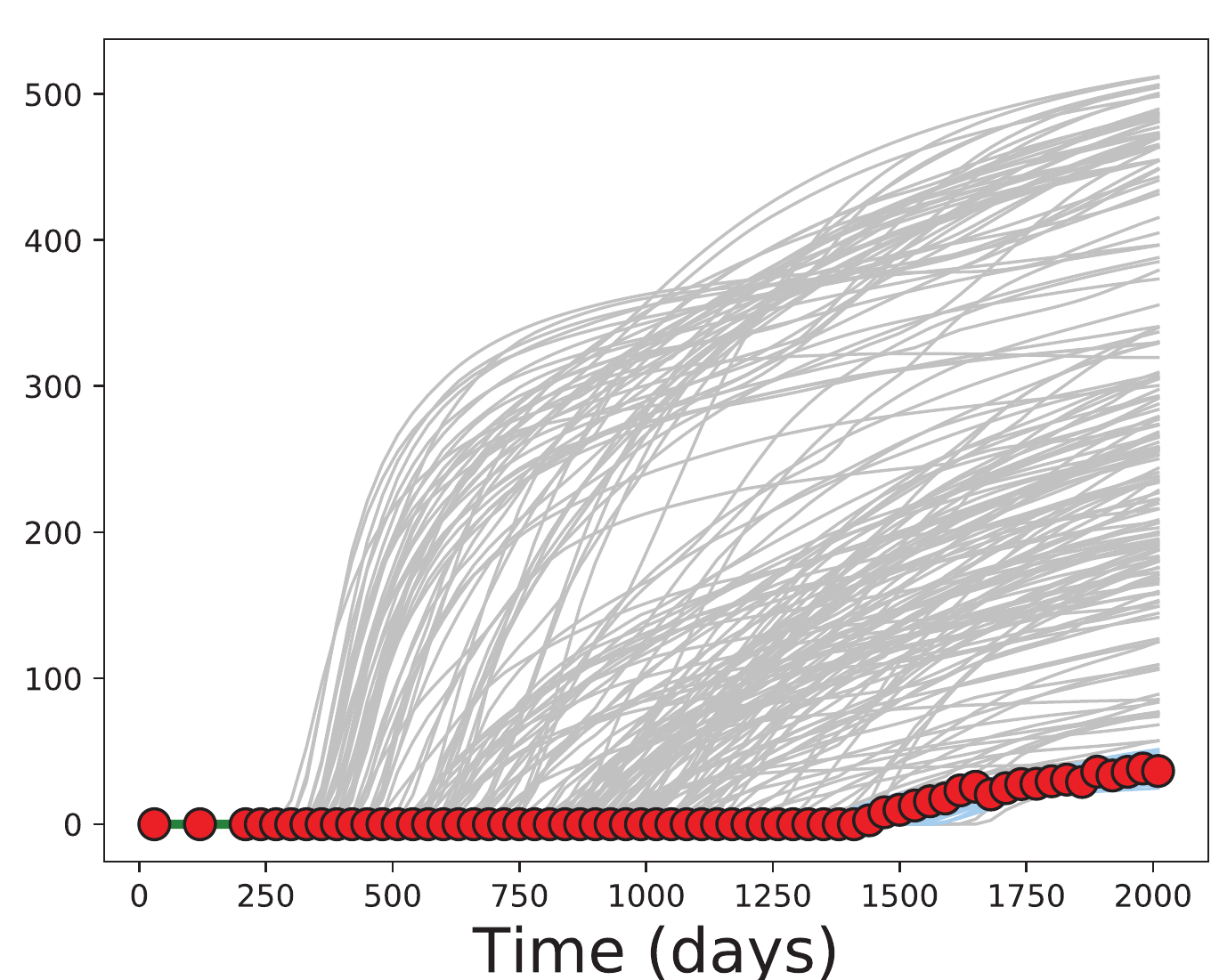}
  }
  \linebreak
  \subfloat[Well P3]{
        \includegraphics[width=0.4\linewidth]{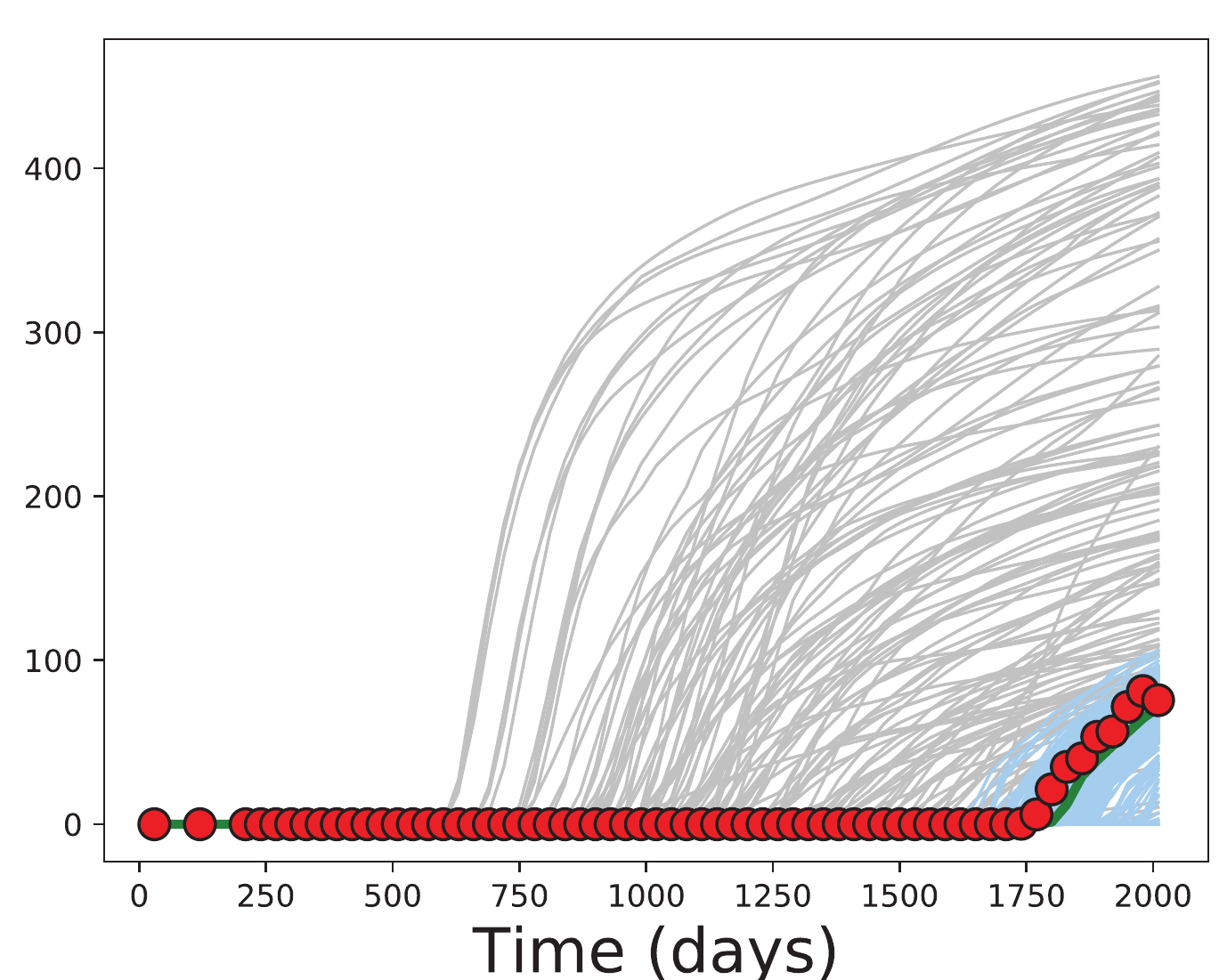}
  }
  \subfloat[Well P4]{
        \includegraphics[width=0.4\linewidth]{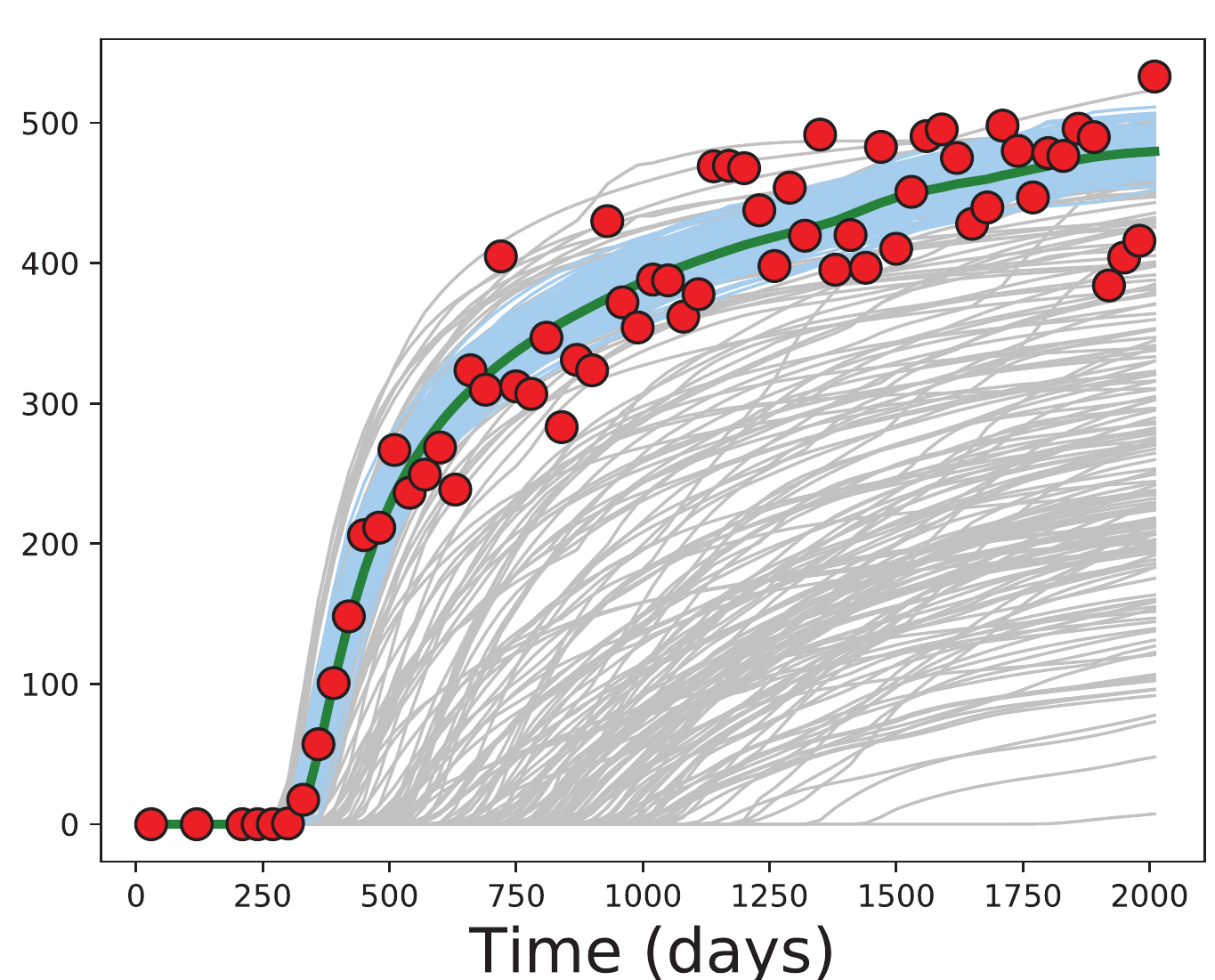}
  }
\caption{Water production rate in bbl/day. Test case 1. Red dots are the observed data points, gray and blue curves are the predicted data from the prior and posterior ensembles, respectively. The green curve is the mean of the posterior ensemble.}
\label{Fig:Case1-WPR}
\end{figure}

\begin{figure}
  \centering
  \subfloat[Reference]{
        \includegraphics[width=0.18\linewidth]{case1_reference.pdf}
  }
  \subfloat[ES-MDA]{
        \includegraphics[width=0.18\linewidth]{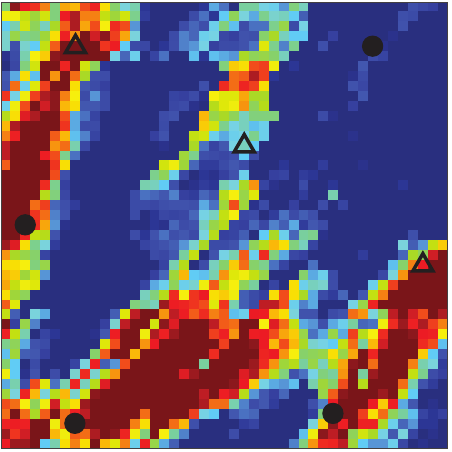}
  }
  \subfloat[ES-MDA-OPCA]{
        \includegraphics[width=0.18\linewidth]{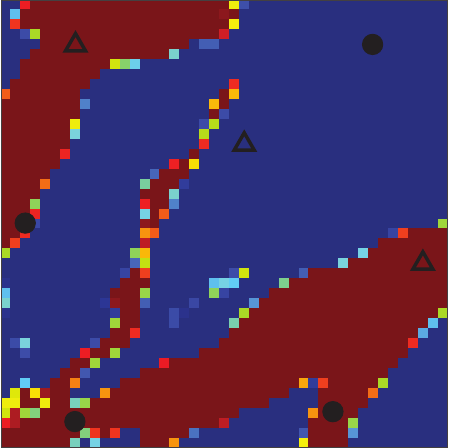}
  }
  \subfloat[ES-MDA-DBN]{
        \includegraphics[width=0.18\linewidth]{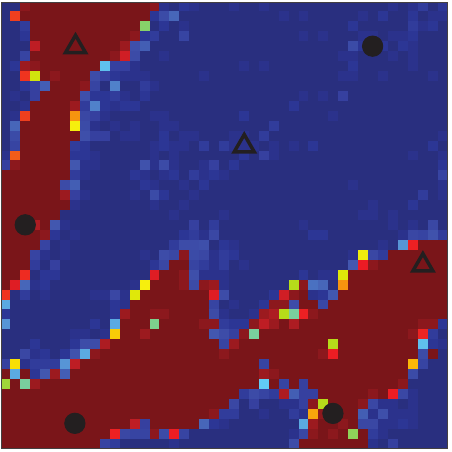}
  }
  \subfloat[ES-MDA-CVAE]{
        \includegraphics[width=0.18\linewidth]{case1_post_001.pdf}
  }
  \subfloat{
        \includegraphics[width=0.048\linewidth]{case1_colorbar.pdf}
  }
\caption{Comparison of the first realization of permeability obtained with standard ES-MDA, ES-MDA-OPCA, ES-MDA-DBN e ES-MDA-CVAE. Test case 1.}
\label{Fig:Case1-Comparison}
\end{figure}

We tested the proposed ES-MDA-CVAE to assimilate production data. We considered four oil producing and three water injection wells as shown in Fig.~\ref{Fig:Case1-True}. All producing wells operate at constant bottom-hole pressure of 3,000~psi. The water injection wells operate at 4,000~psi. The synthetic measurements correspond to oil and water rate data corrupted with Gaussian random noise with standard deviation of 5\% of the data predicted by the reference model. We use a prior ensemble with $N_e = 200$ realizations and $N_a = 4$ iterations. We did not include any facies data (hard data) in order to make the problem more challenging for assimilation of production data. Figure~\ref{Fig:Case1-Realization} shows the first five prior and posterior realizations obtained with ES-MDA-CVAE. Clearly all posterior realizations are able to reproduce the main features of the reference model (Fig.~\ref{Fig:Case1-True}). Figure~\ref{Fig:Case1-WPR} shows the observed and predicted water rate for four wells showing a good data match. In \citep{canchumuni:18a}, we used the same problem to test the standard ES-MDA and parameterizations with OPCA and DBN. Figure~\ref{Fig:Case1-Comparison} shows the first realization obtained with each method. The results in this figure clearly show the superior performance of ES-MDA-CVAE.

\subsection{Test Case 2}

\begin{figure}[ht!]
  \centering
  \subfloat{
    \includegraphics[width=0.25\linewidth]{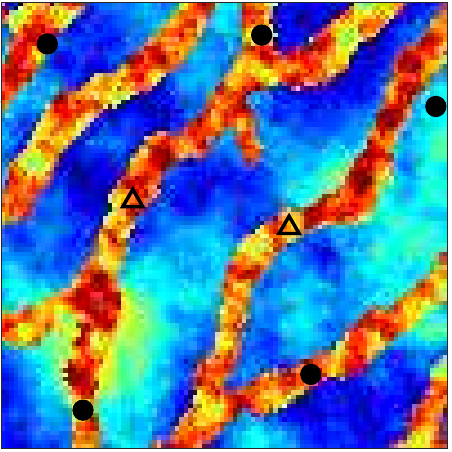}
    }
  \subfloat{
    \includegraphics[width=0.035\linewidth]{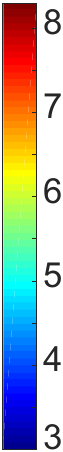}
    }
  \subfloat{
    \includegraphics[width=0.3\linewidth]{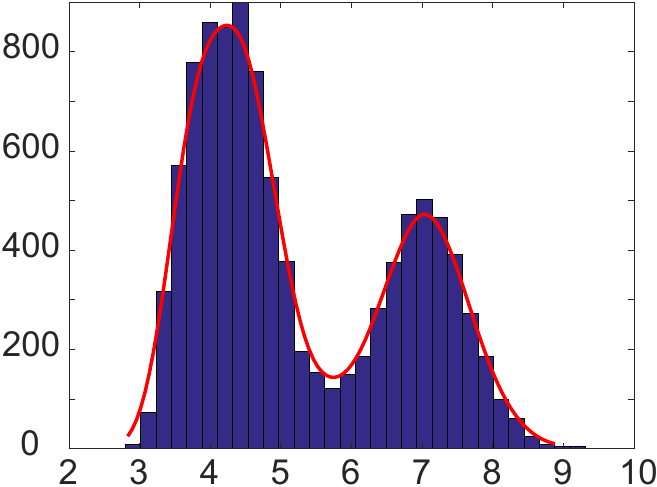}
    }
\caption{Reference log-permeability field (ln-mD). Test case 2. Circles are oil producing wells and triangles are water injection wells.}
\label{Fig:Case2-True}
\end{figure}

The second test case is the same used in \citep{emerick:17a}. Figure~\ref{Fig:Case2-True} shows the reference permeability field and the corresponding histogram. The model has $100 \times 100$ gridblocks with uniform size of 75~meters and constant thickness of 20~meters. Similarly to the first test case, this case has two facies (channel and background sand) generated with the \verb"snesim" algorithm. However, in this case we update the facies type and the permeability within each facies simultaneously. The permeability values within each facies were obtained with sequential Gaussian simulation. More details about the construction of this problem can be found in \citep{emerick:17a}.

\begin{figure}
  \centering
        \includegraphics[width=0.18\linewidth]{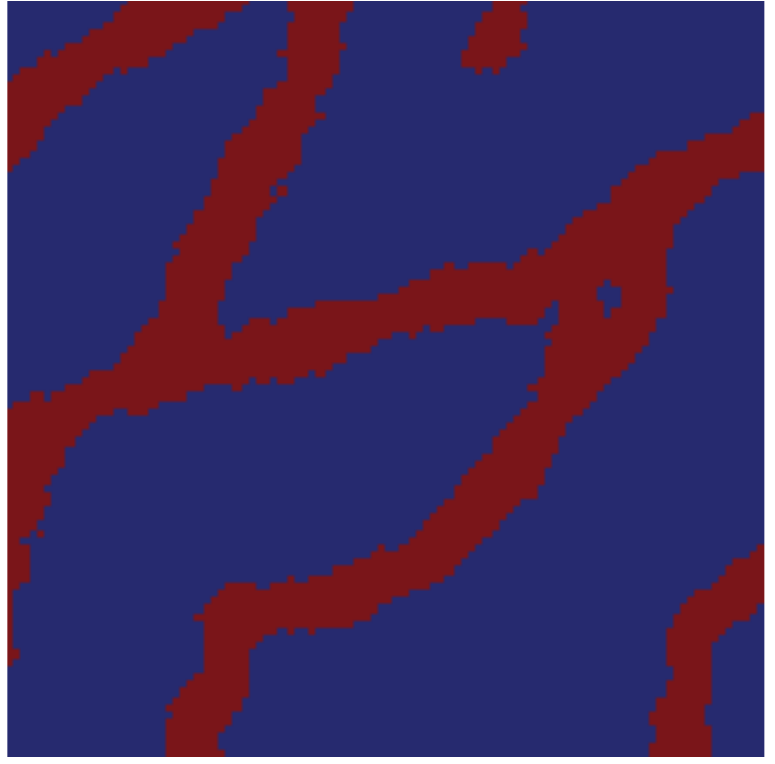}
        \includegraphics[width=0.18\linewidth]{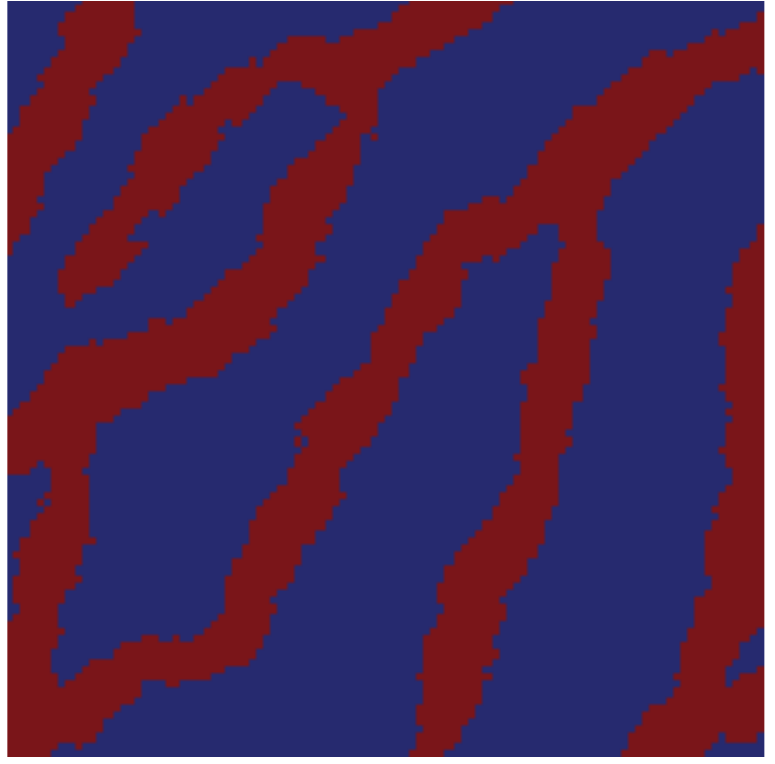}
  \subfloat[Input]{
        \includegraphics[width=0.18\linewidth]{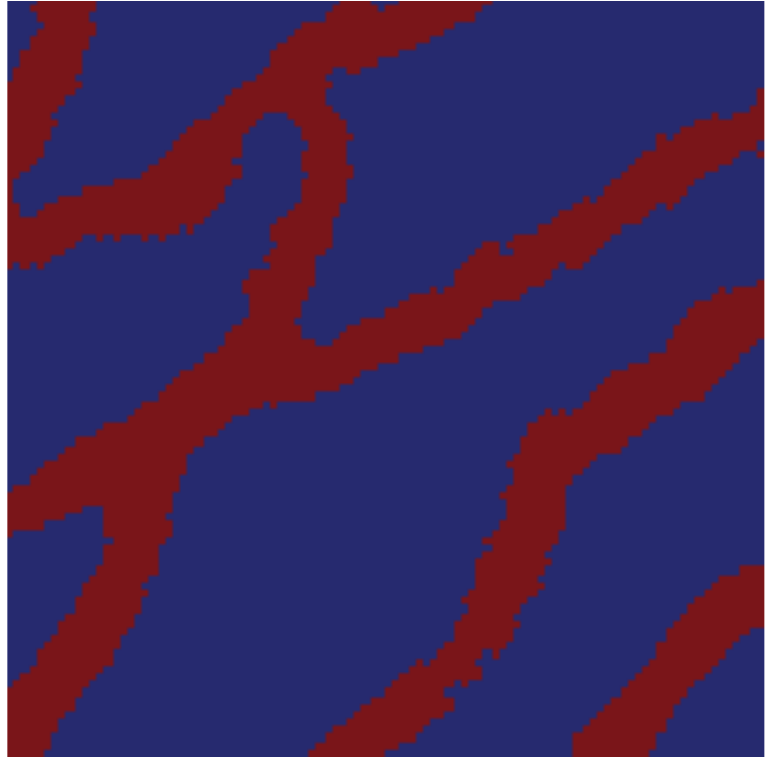}
  }
        \includegraphics[width=0.18\linewidth]{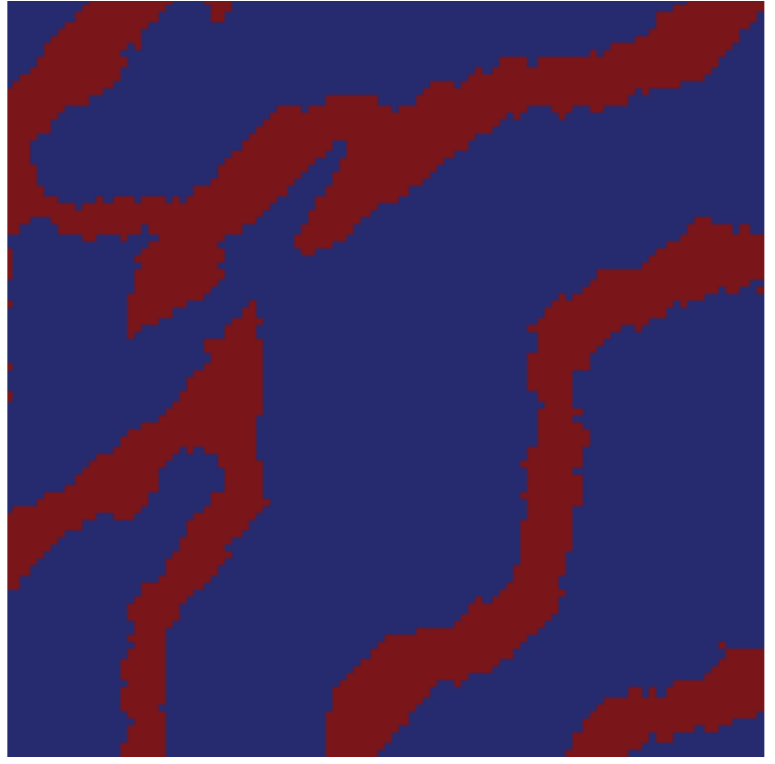}
        \includegraphics[width=0.18\linewidth]{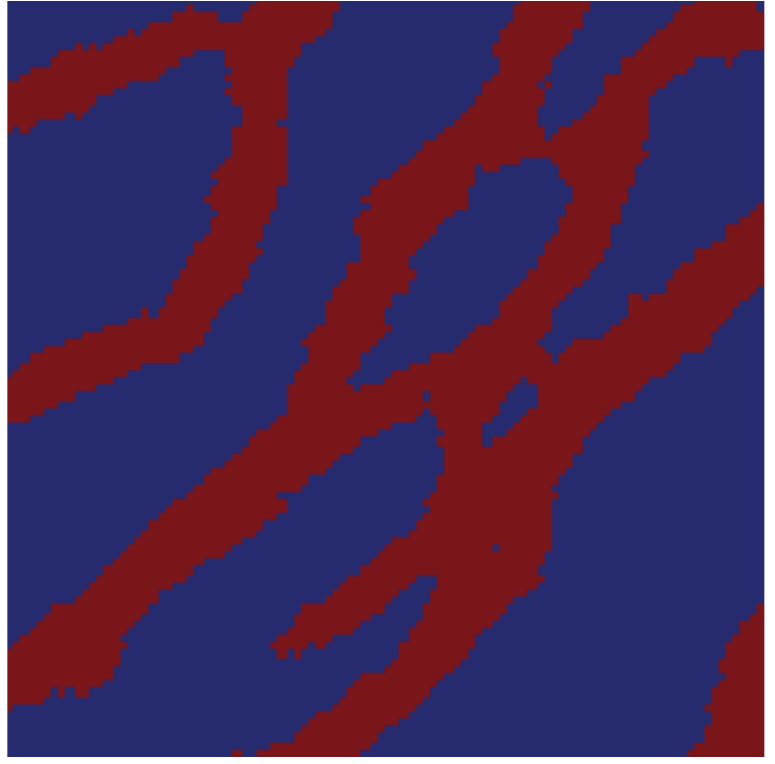}
\linebreak
        \includegraphics[width=0.18\linewidth]{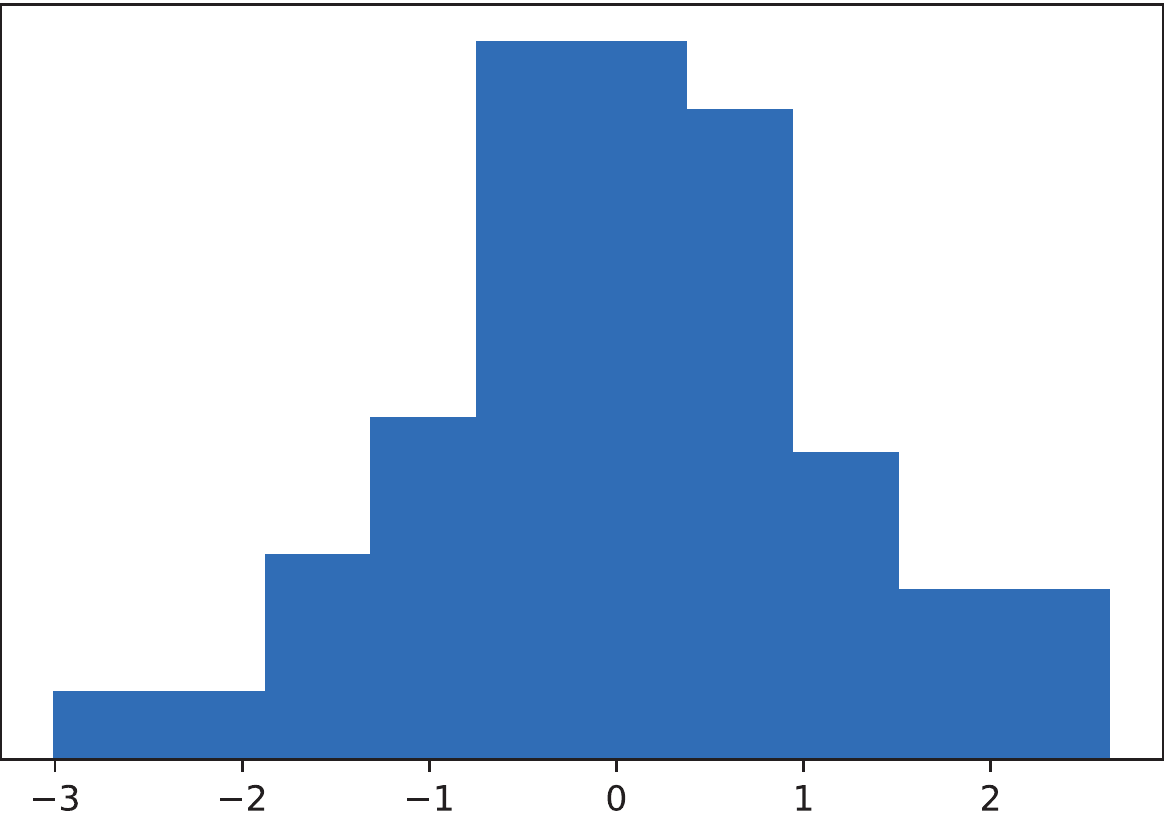}
        \includegraphics[width=0.18\linewidth]{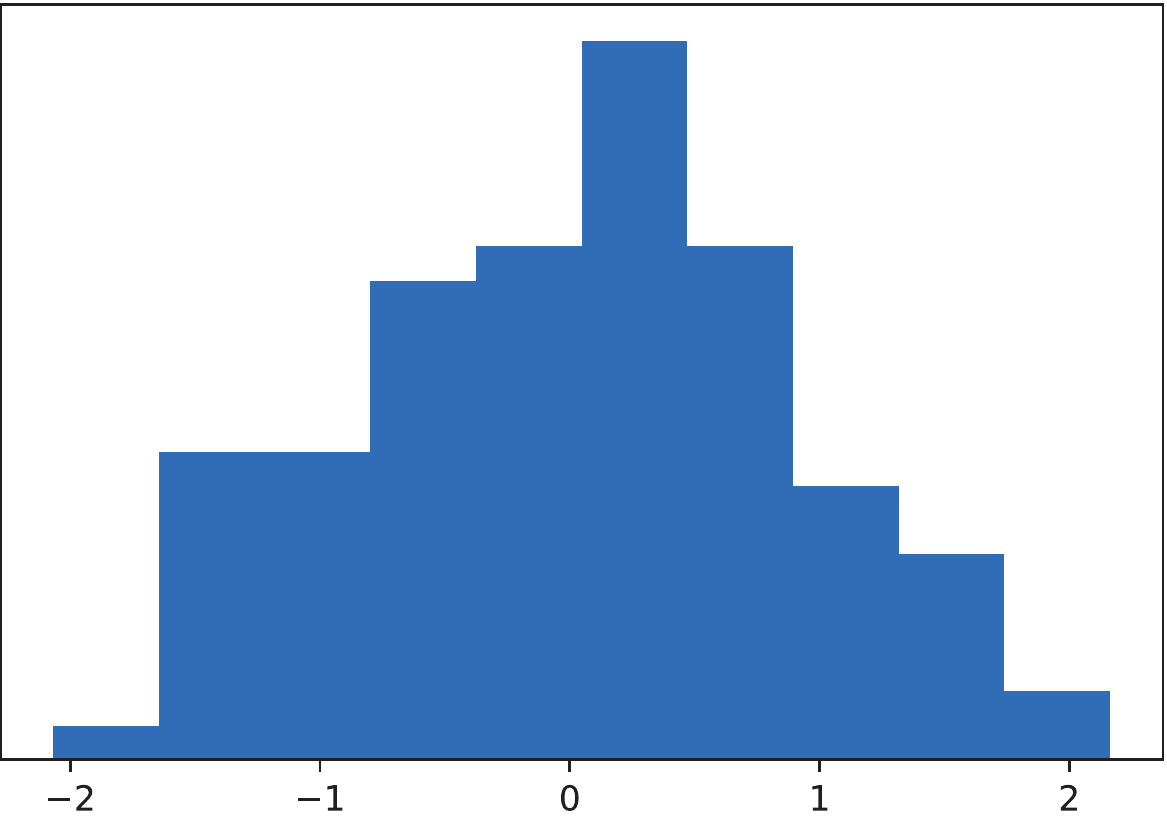}
  \subfloat[Latent]{
        \includegraphics[width=0.18\linewidth]{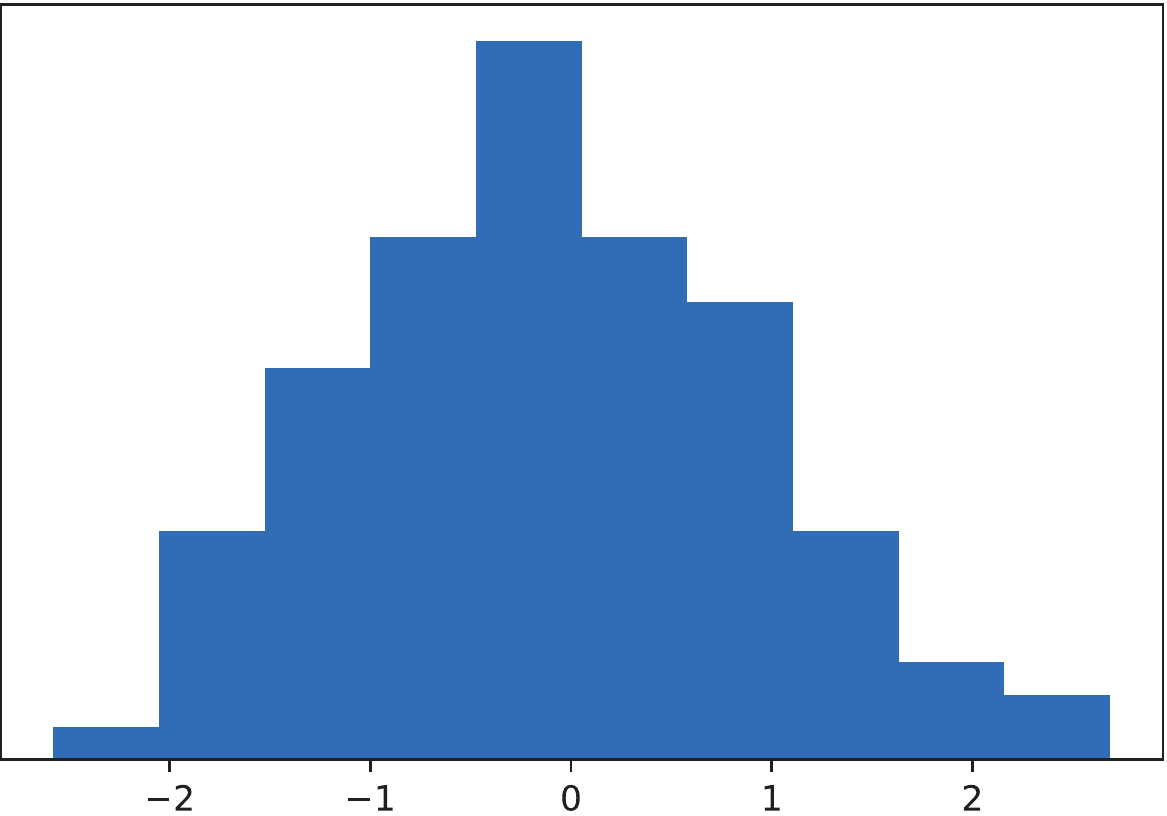}
  }
        \includegraphics[width=0.18\linewidth]{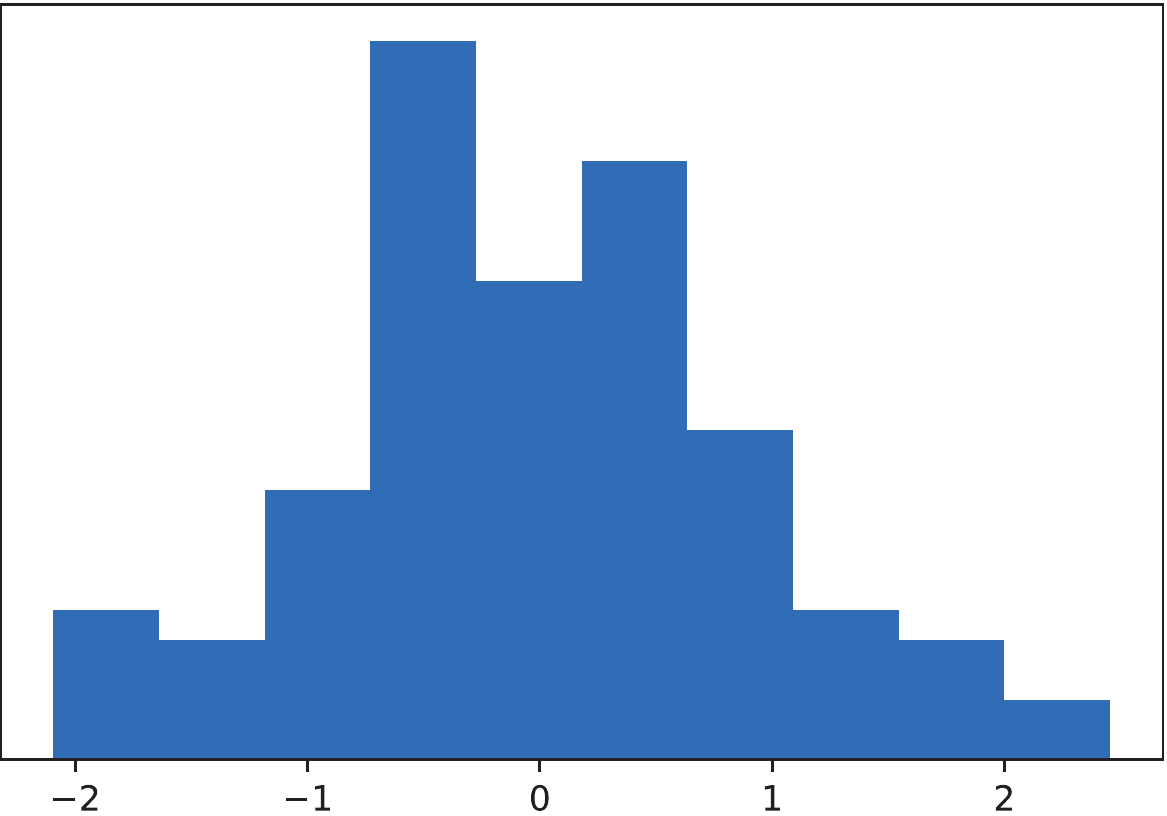}
        \includegraphics[width=0.18\linewidth]{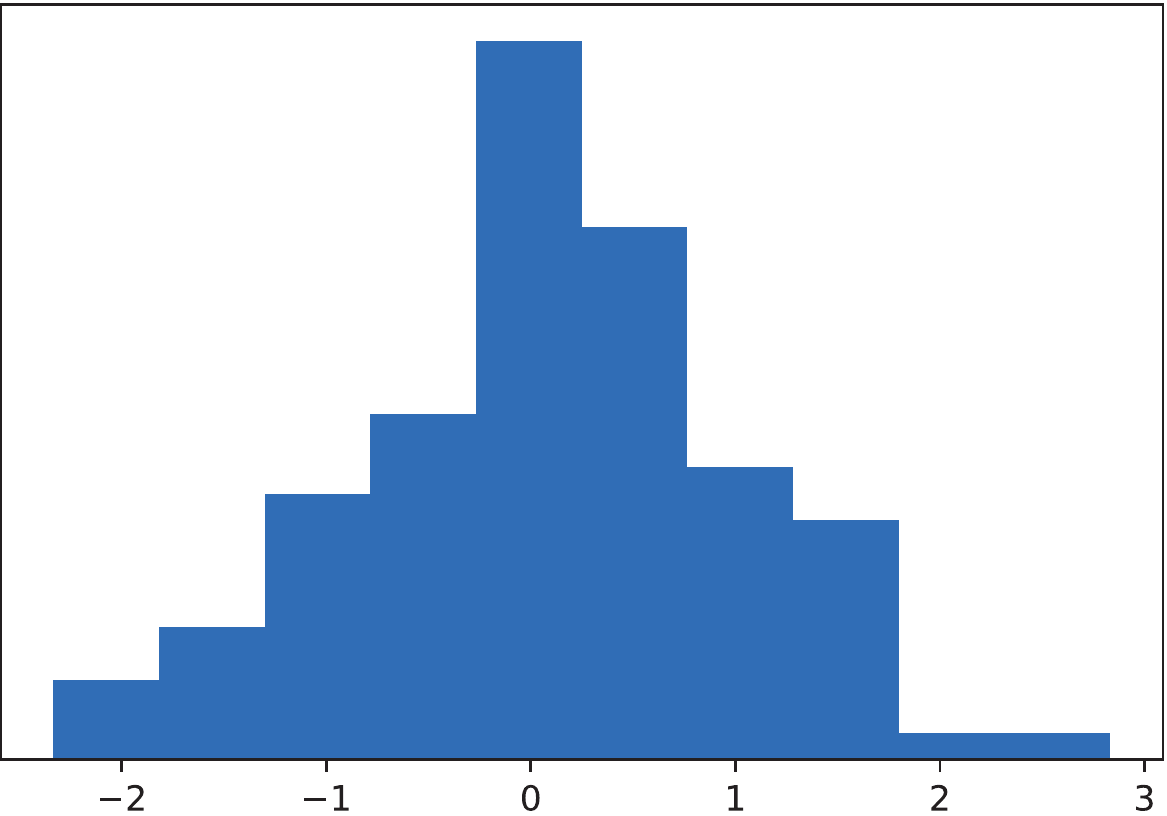}
\linebreak
        \includegraphics[width=0.18\linewidth]{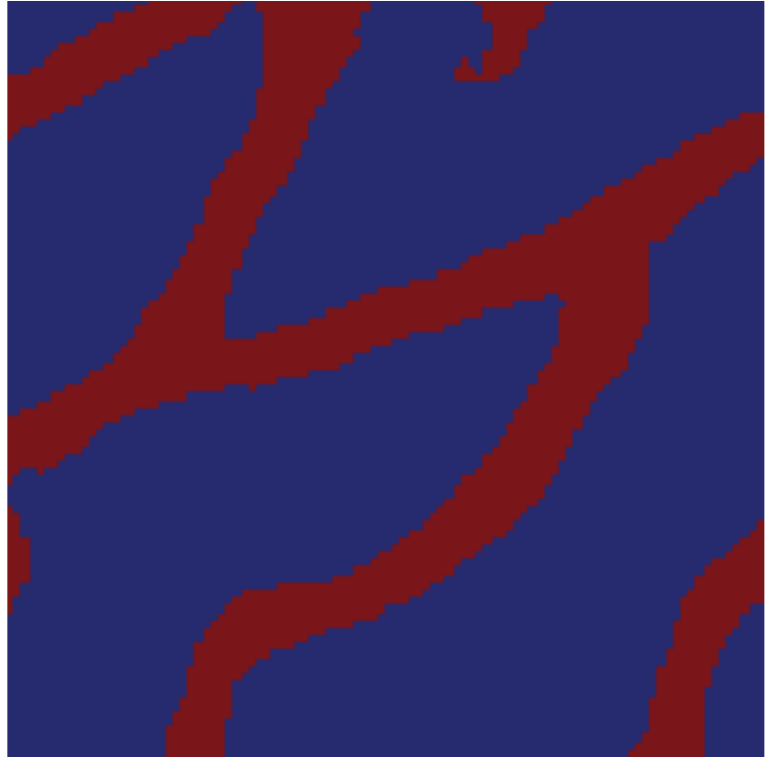}
        \includegraphics[width=0.18\linewidth]{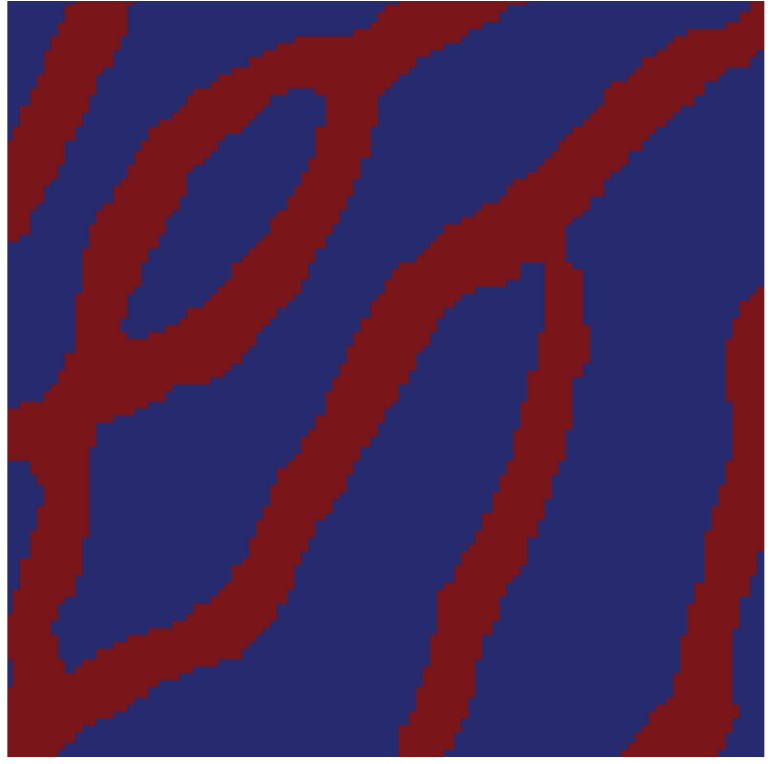}
  \subfloat[Reconstructed]{
        \includegraphics[width=0.18\linewidth]{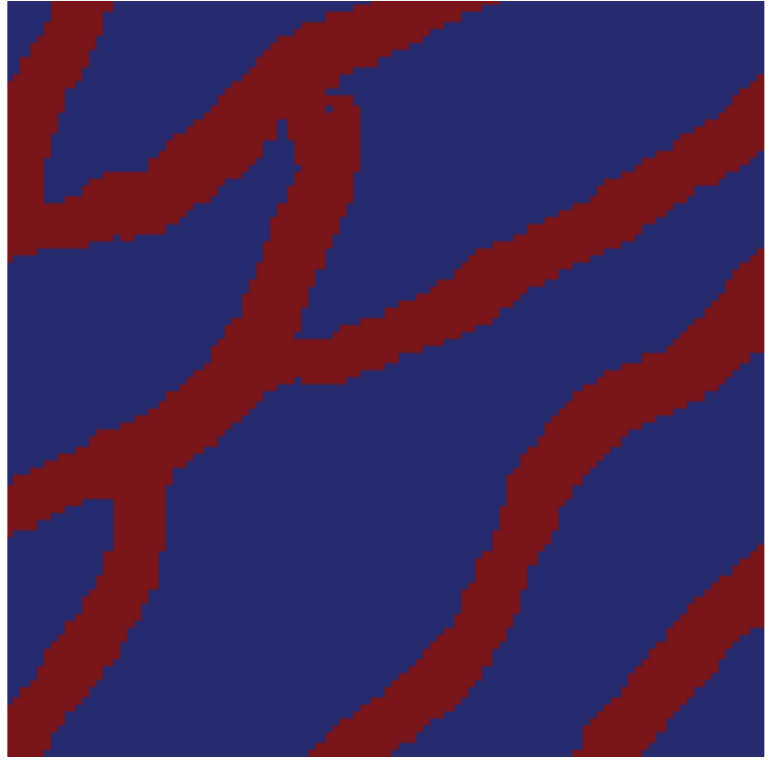}
  }
        \includegraphics[width=0.18\linewidth]{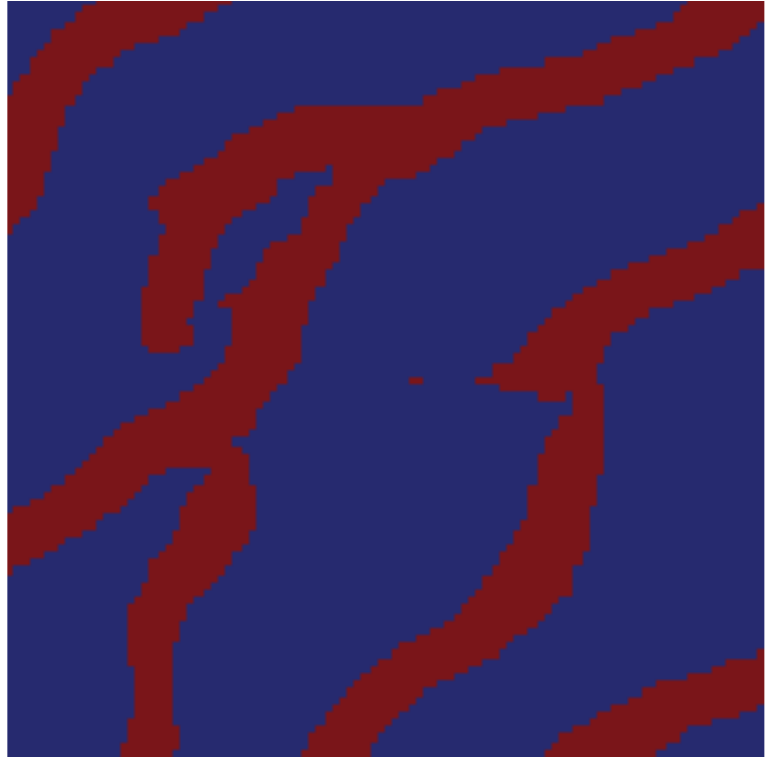}
        \includegraphics[width=0.18\linewidth]{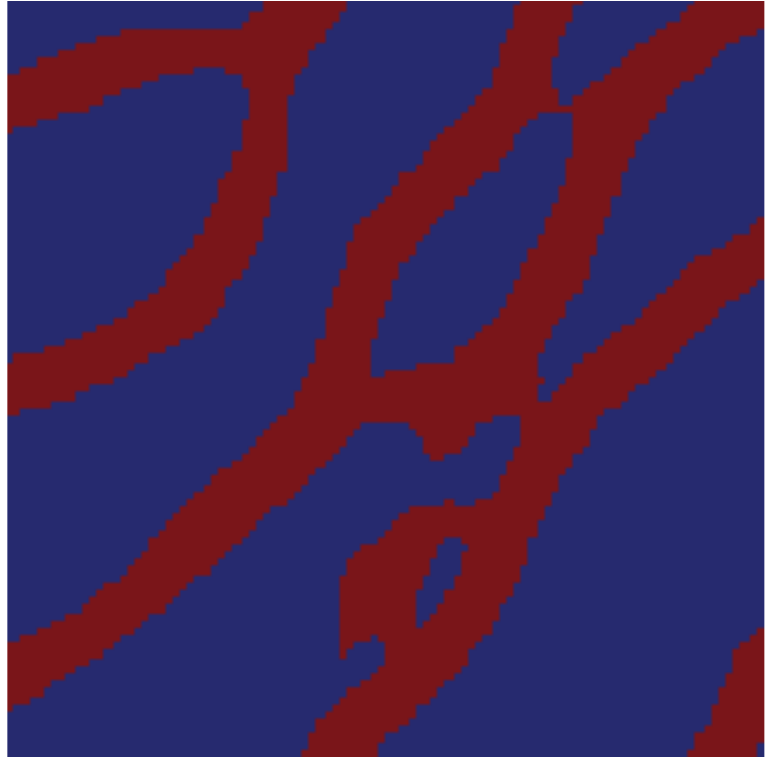}
\caption{Training process showing the first five realizations of the validation set and the corresponding histograms of the latent vector. Test case 2.}
\label{Fig:Case2-Training}
\end{figure}

For this test case, we used a CVAE network with architecture similar to the previous case, with few changes only to accommodate the fact that the size of the models are different. We used a training set with 32,000 realizations and 8,000 for validation. The training required 42 minutes in a cluster with four GPUs (NVIDIA TESLA P100). The final reconstruction accuracy for the validation set was 93.3\%. Figure~\ref{Fig:Case2-Training} shows the first five realizations of the validation set before and after reconstruction and the corresponding histograms of the latent vectors. Again the CVAE was able to achieve a reasonable reconstruction of the channels.

\begin{figure}
  \centering
        \includegraphics[width=0.18\linewidth]{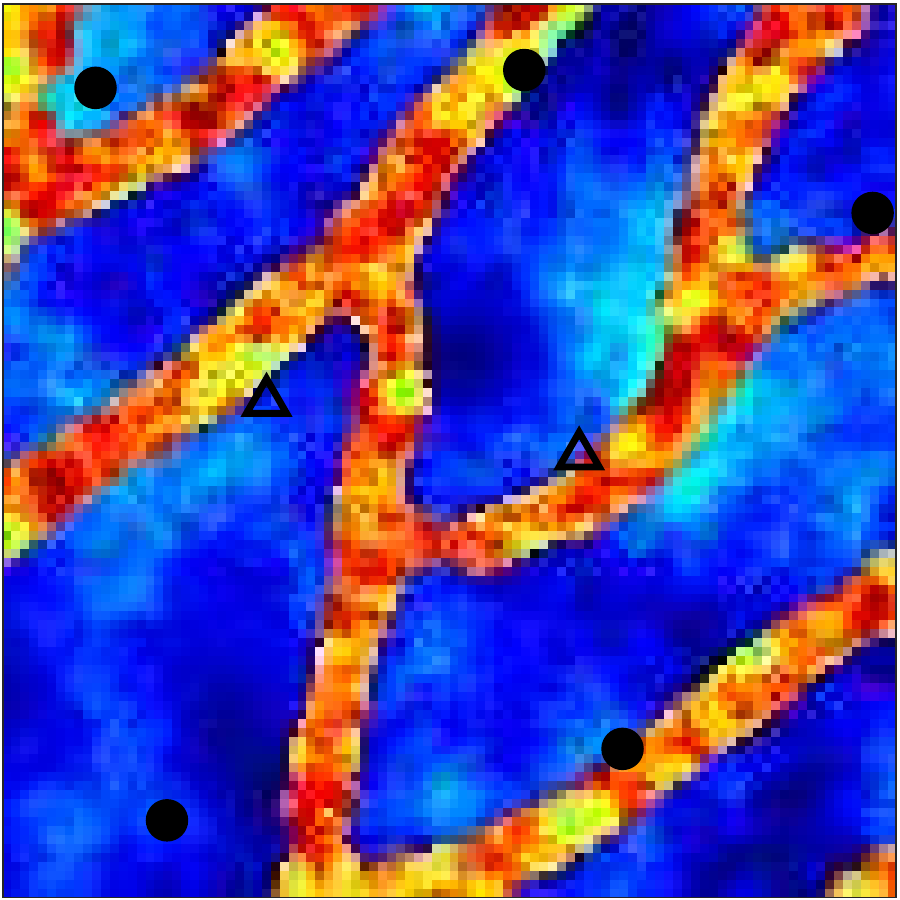}
        \includegraphics[width=0.18\linewidth]{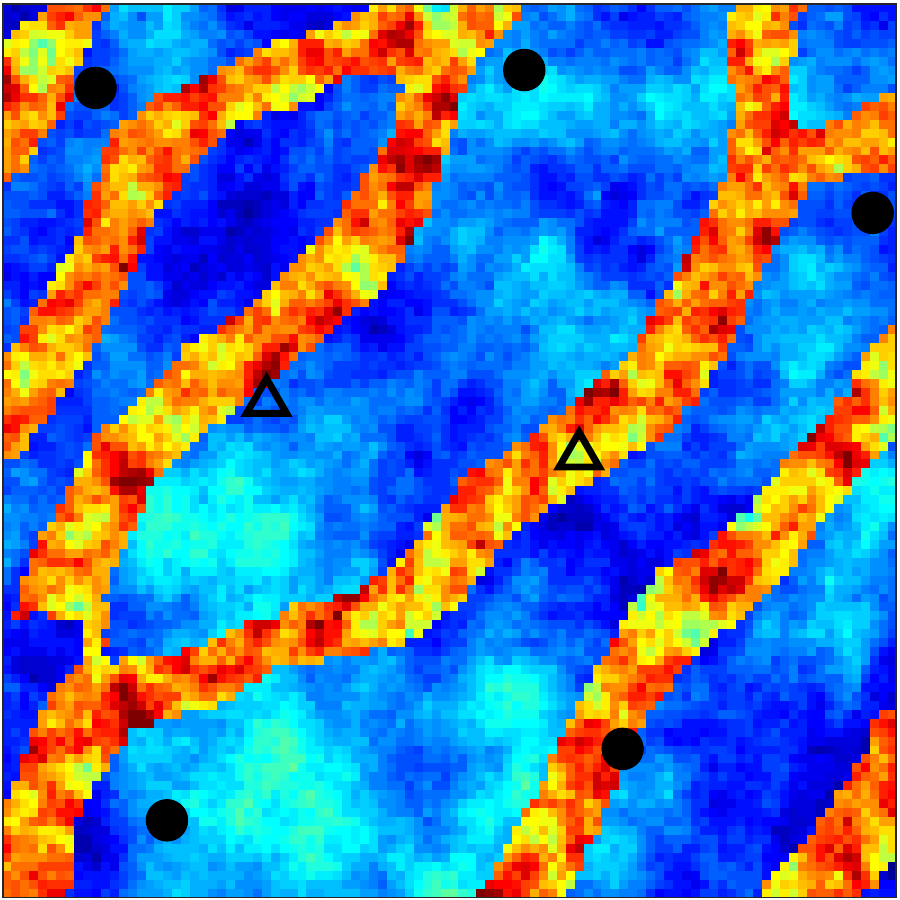}
  \subfloat[Prior]{
        \includegraphics[width=0.18\linewidth]{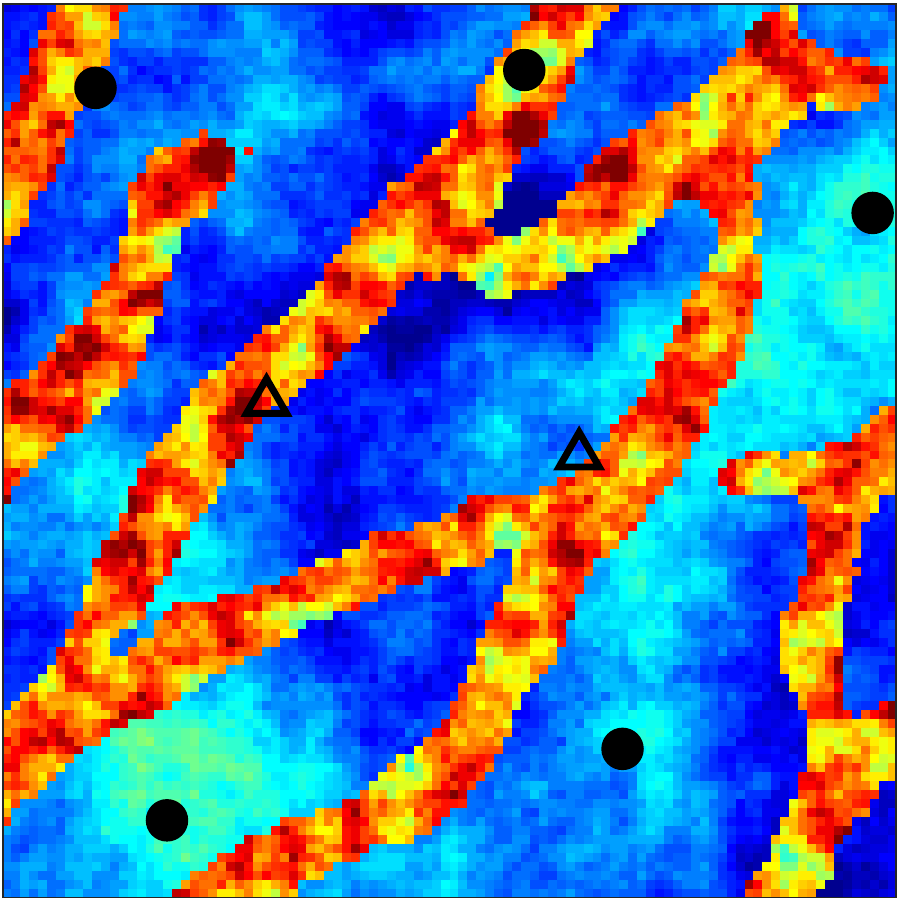}
  }
        \includegraphics[width=0.18\linewidth]{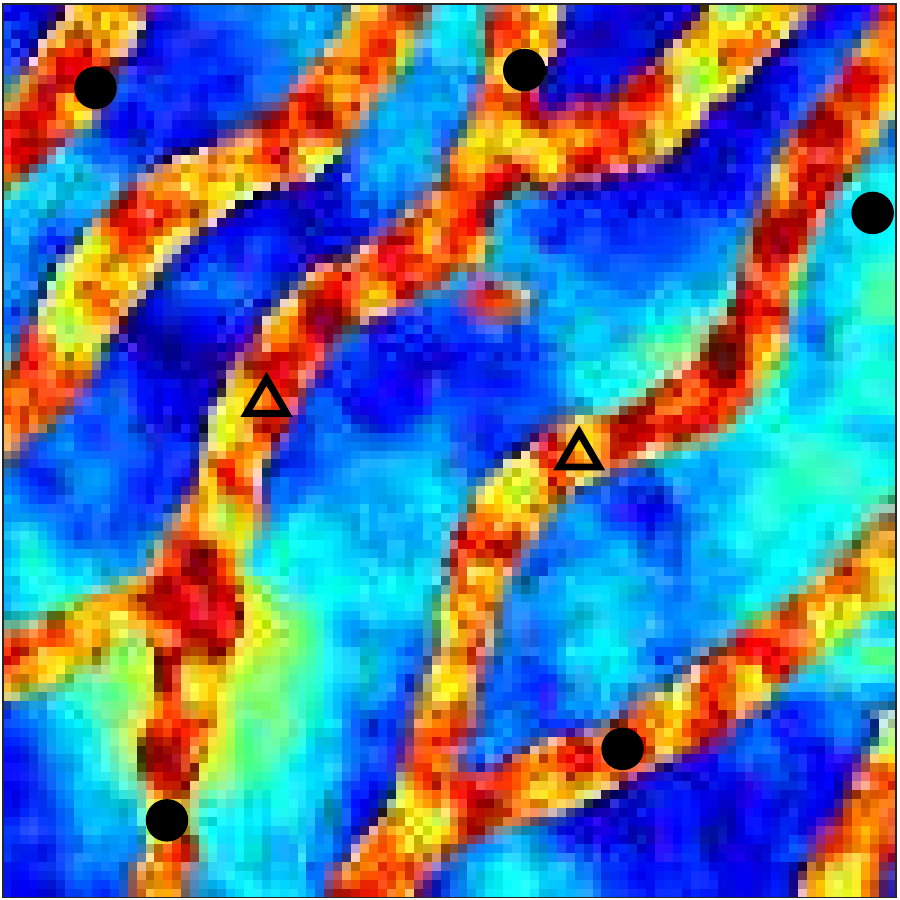}
        \includegraphics[width=0.18\linewidth]{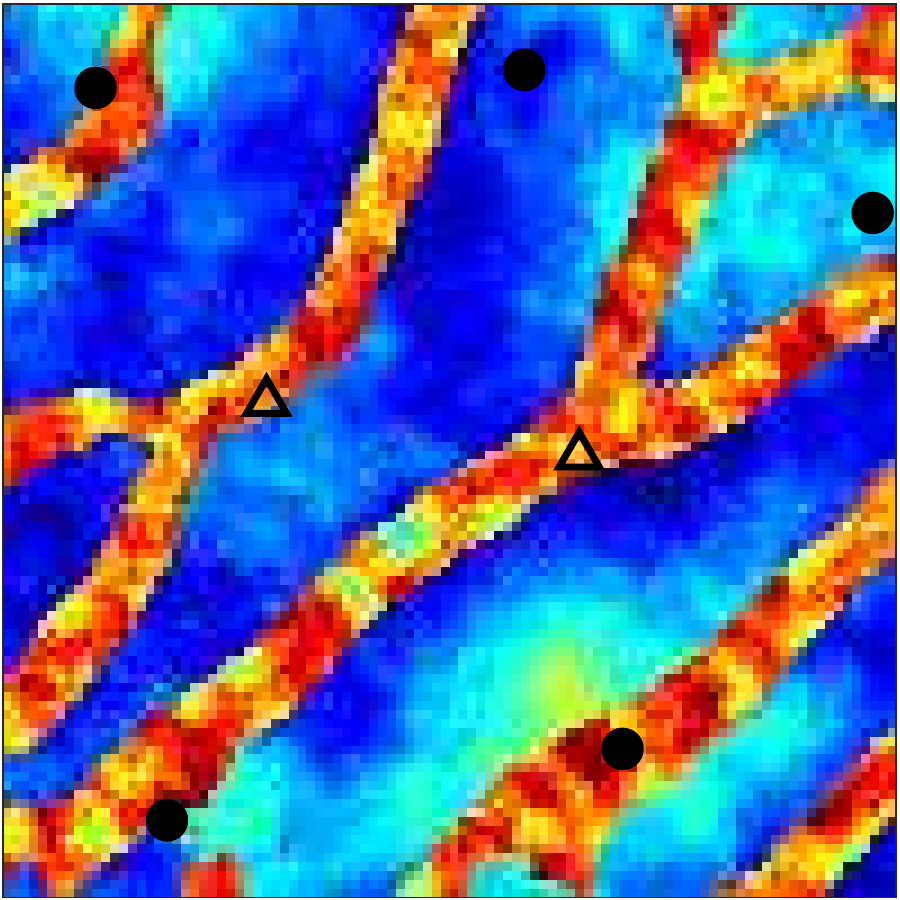}
        \includegraphics[width=0.025\linewidth]{case2_colorbar.pdf}
\linebreak
        \includegraphics[width=0.18\linewidth]{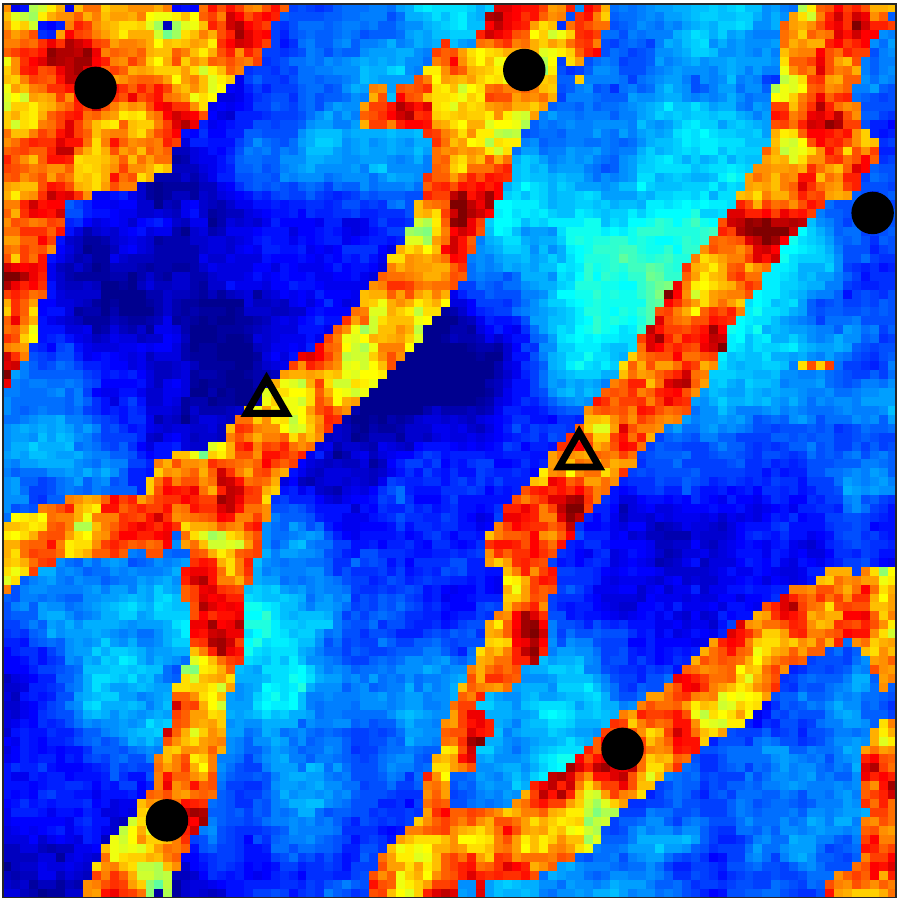}
        \includegraphics[width=0.18\linewidth]{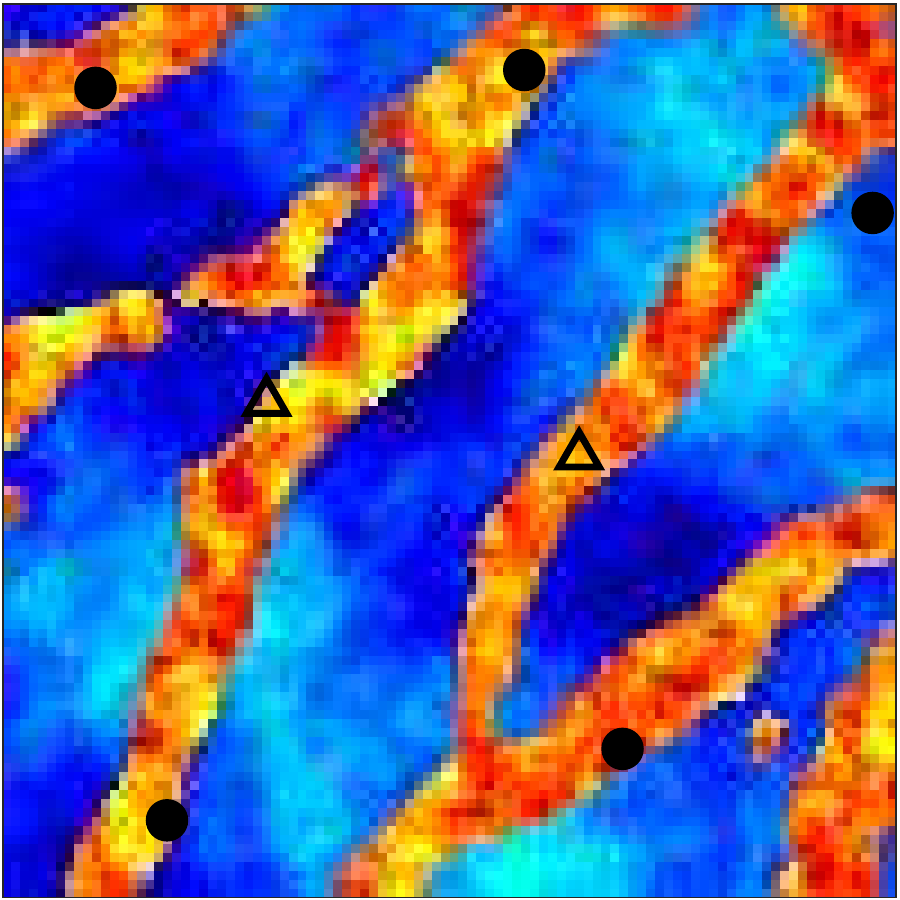}
  \subfloat[Post]{
        \includegraphics[width=0.18\linewidth]{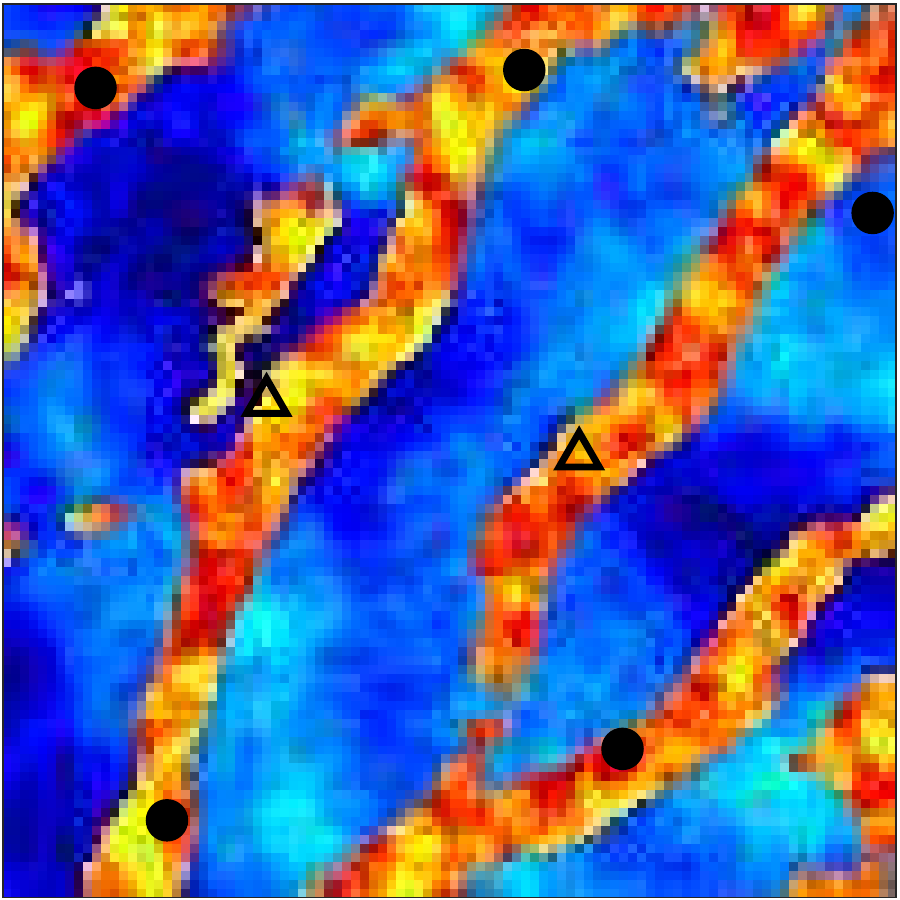}
  }
        \includegraphics[width=0.18\linewidth]{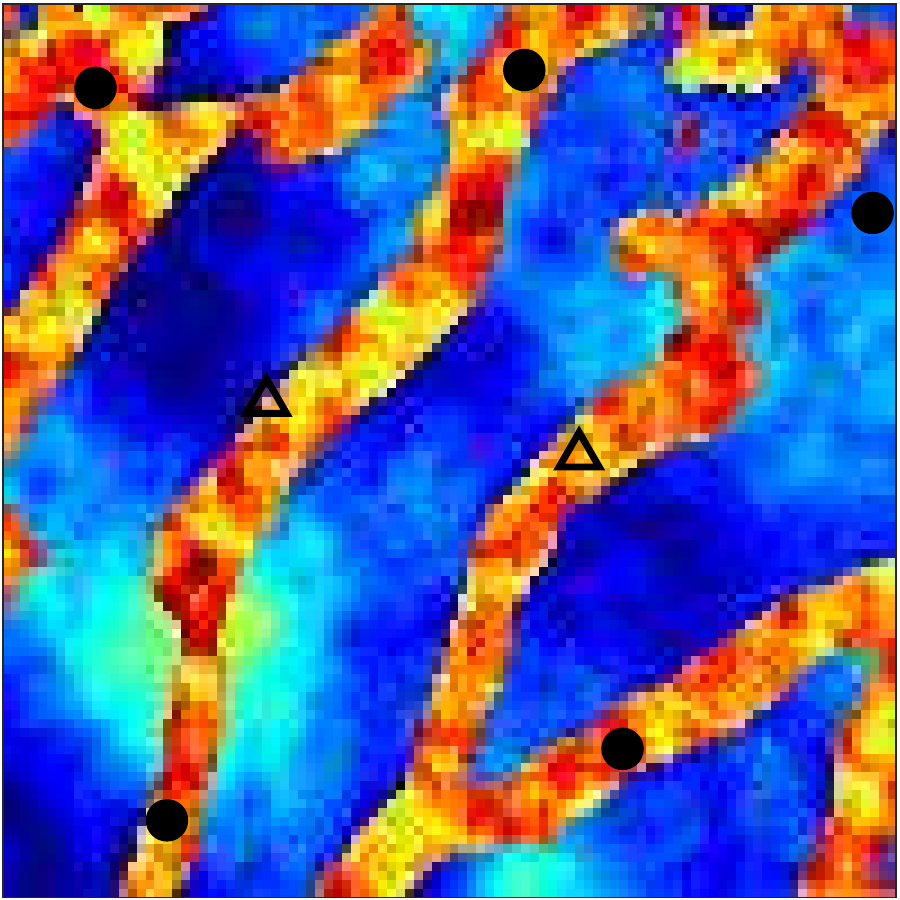}
        \includegraphics[width=0.18\linewidth]{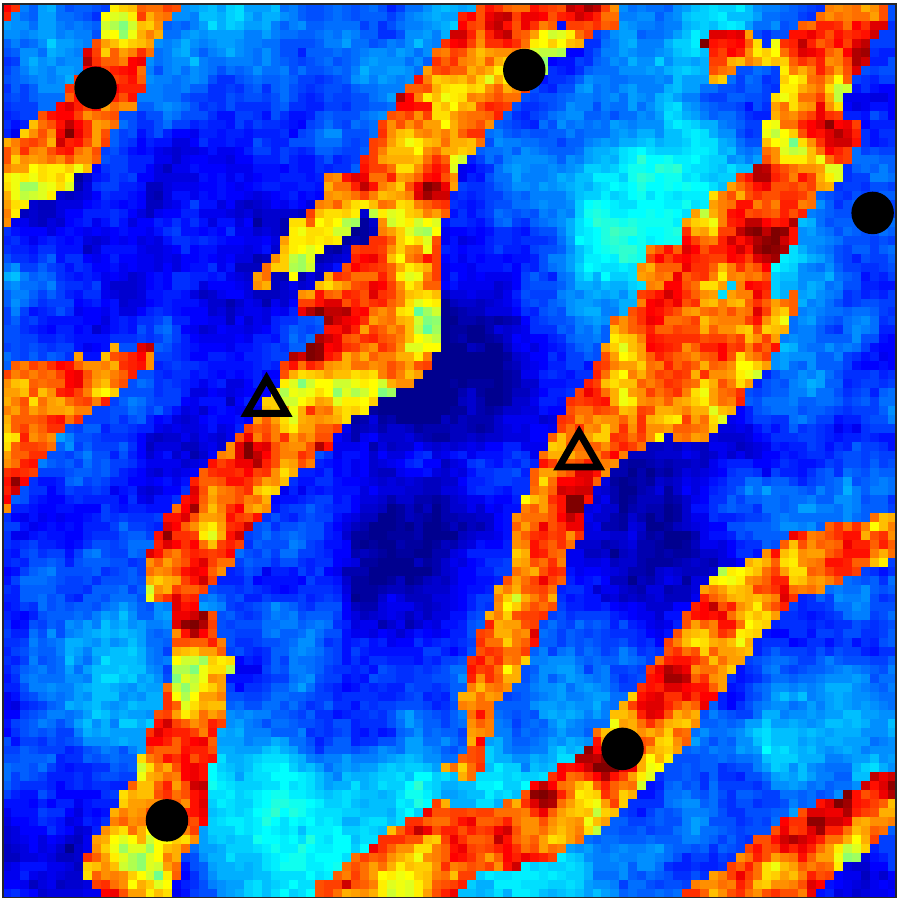}
        \includegraphics[width=0.025\linewidth]{case2_colorbar.pdf}
\caption{First five prior and posterior realizations of log-permeability (ln-mD) after assimilation of production data. Test case 2.}
\label{Fig:Case2-Realization}
\end{figure}

\begin{figure}
  \centering
  \subfloat[Well P1]{
        \includegraphics[width=0.4\linewidth]{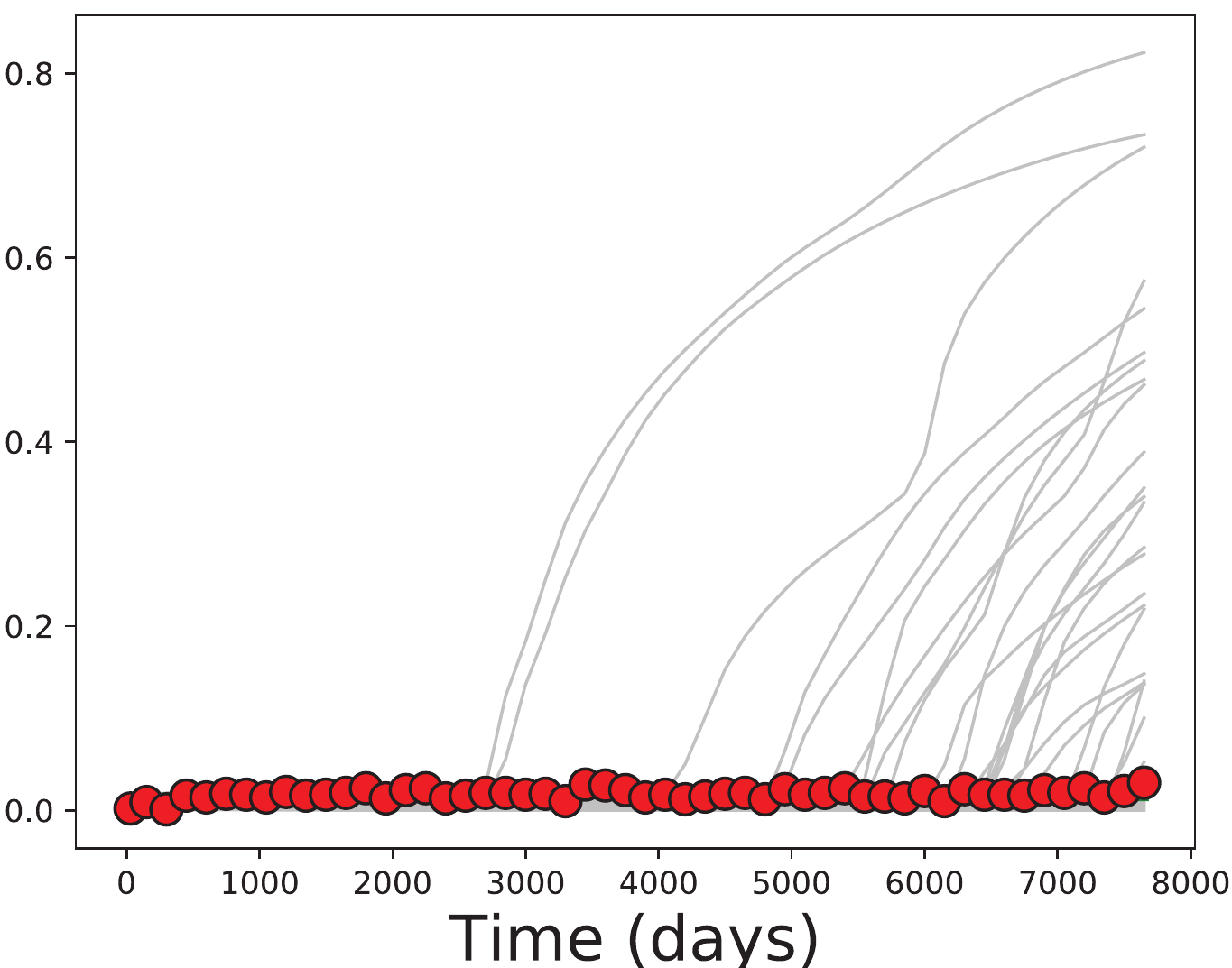}
  }
  \subfloat[Well P2]{
        \includegraphics[width=0.4\linewidth]{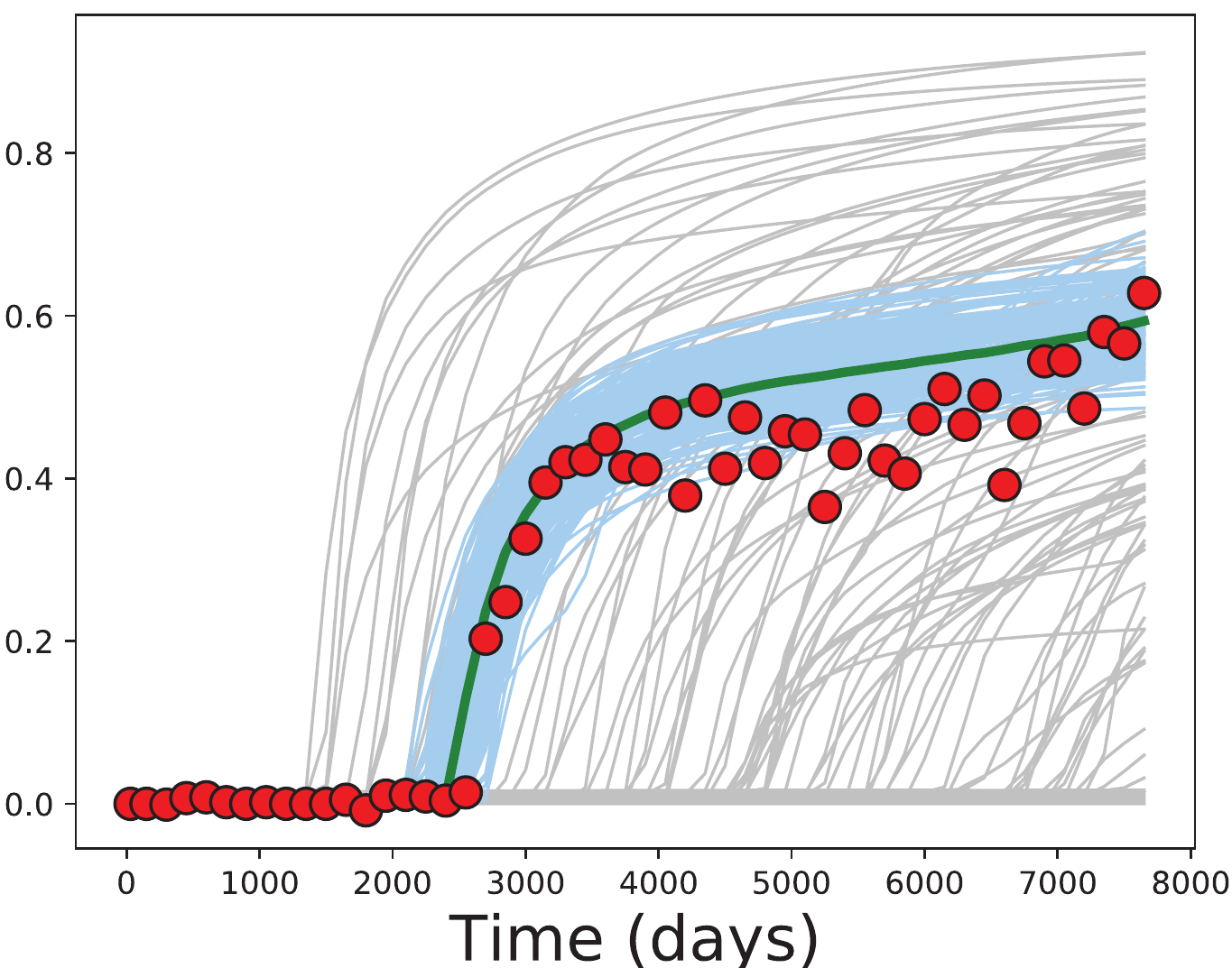}
  }
  \linebreak
  \subfloat[Well P3]{
        \includegraphics[width=0.4\linewidth]{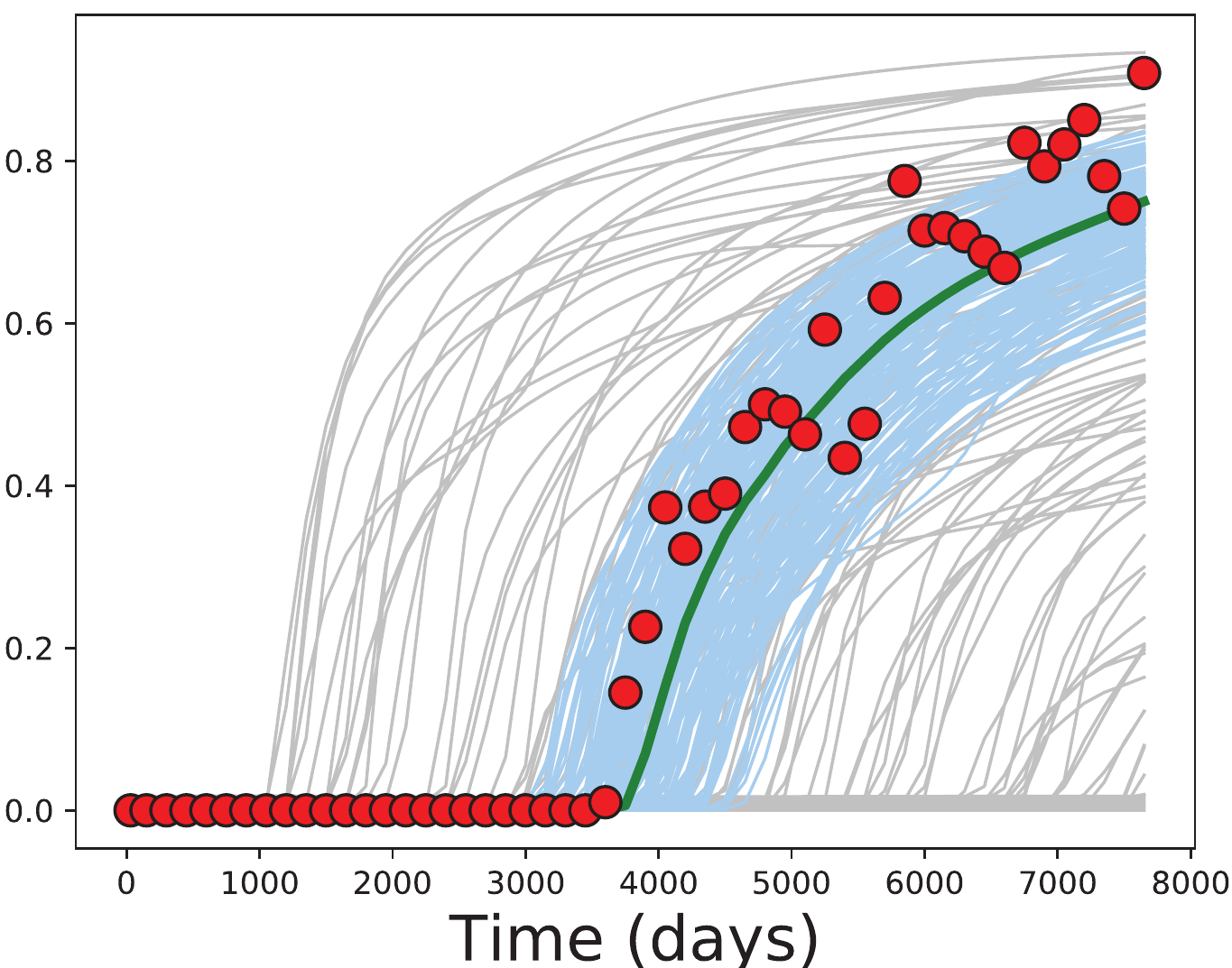}
  }
  \subfloat[Well P4]{
        \includegraphics[width=0.4\linewidth]{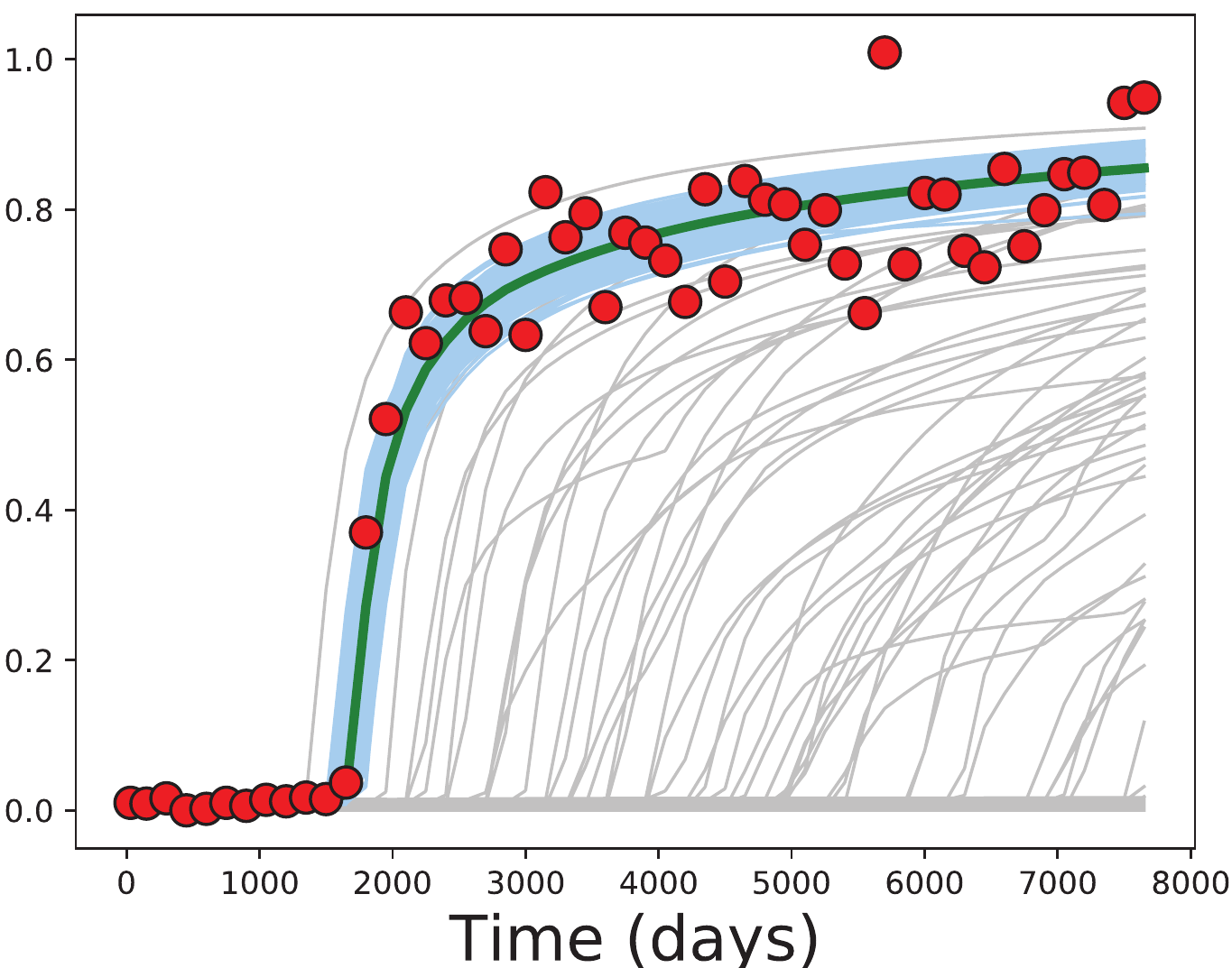}
  }
\caption{Water cut. Test case 2. Red dots are the observed data points, gray and blue curves are the predicted data from the prior and posterior ensembles, respectively. The green curve is the mean of the posterior ensemble.}
\label{Fig:Case2-WCT}
\end{figure}

\begin{figure}
  \centering
  \subfloat[Reference]{
        \includegraphics[width=0.18\linewidth]{case2_reference.pdf}
  }
  \subfloat[ES-MDA]{
        \includegraphics[width=0.18\linewidth]{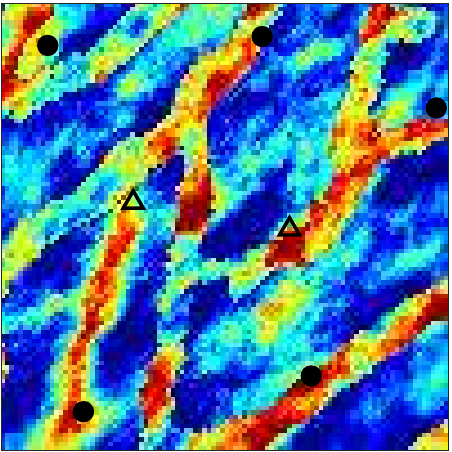}
  }
  \subfloat[ES-MDA-OPCA]{
        \includegraphics[width=0.18\linewidth]{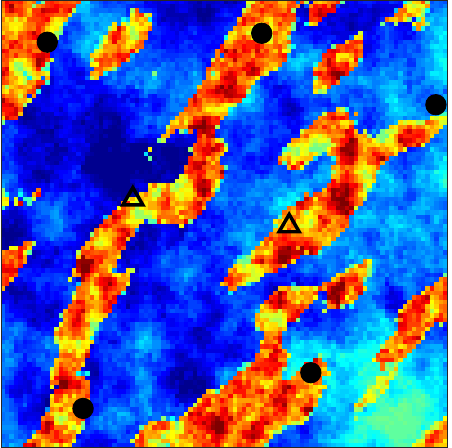}
  }
  \subfloat[ES-MDA-DBN]{
        \includegraphics[width=0.18\linewidth]{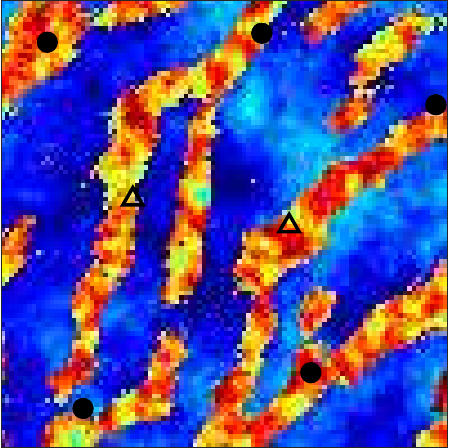}
  }
  \subfloat[ES-MDA-CVAE]{
        \includegraphics[width=0.18\linewidth]{case2_post_001.pdf}
  }
  \subfloat{
        \includegraphics[width=0.025\linewidth]{case2_colorbar.pdf}
  }
\caption{Comparison of the first realization of permeability obtained with standard ES-MDA, ES-MDA-OPCA, ES-MDA-DBN e ES-MDA-CVAE. Test case 2.}
\label{Fig:Case2-Comparison}
\end{figure}

For this test problem, we assimilated water cut data at five oil producing wells and water rate at two water injection wells. The position of the wells is indicated in Fig.~\ref{Fig:Case2-True}. The synthetic measurements were corrupted with Gaussian noise with standard deviation of 5\% to the data predicted by the reference model. We use $N_e = 200$ and $N_a = 20$. Our tests showed that for this problem we needed more MDA iterations than usual, possibly because the parameterization makes the problem more nonlinear. All prior realizations do not include facies data at well locations. During the data assimilation, we update the latent vectors and the permeability values within each facies. Figure~\ref{Fig:Case2-Realization} shows the first five prior and posterior realizations indicating that ES-MDA-CVAE was able to generate plausible facies distributions, i.e., facies with similar features of the prior ones. Figure~\ref{Fig:Case2-WCT} shows the water cut data for four wells indicating reasonable data matches. Figure~\ref{Fig:Case2-Comparison} shows the first realization obtained with ES-MDA, ES-MDA-OPCA, ES-MDA-DBN and ES-MDA-CVAE. The first three results were extracted from \citep{canchumuni:18a}. This figure shows that the standard ES-MDA was no able to preserve well-defined boundaries for the channels. ES-MDA-OPCA and ES-MDA-DBN resulted in better models, however with some discontinuous branches of channels which are not present in the prior models. Again, ES-MDA-CVAE obtained a realization with better representation of the channels.

\clearpage

\subsection{Test Case 3}

\begin{figure}
  \centering
  \subfloat[Permeability]{
    \includegraphics[width=0.5\linewidth]{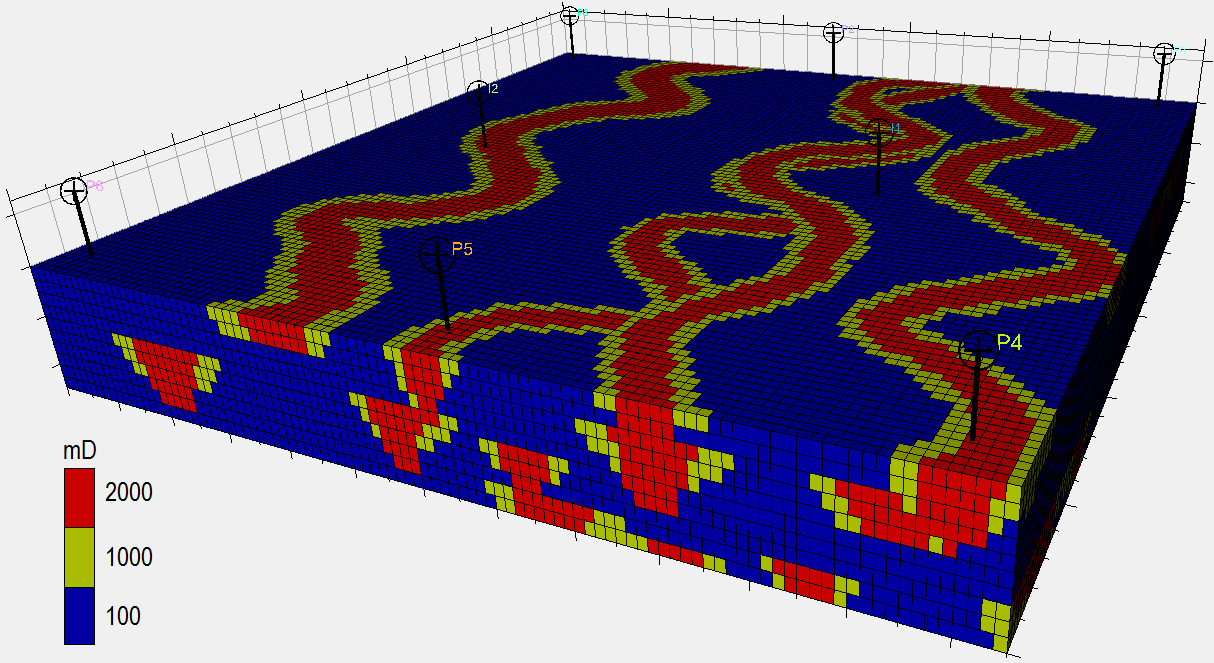}
    }
  \subfloat[Facies]{
    \includegraphics[width=0.5\linewidth]{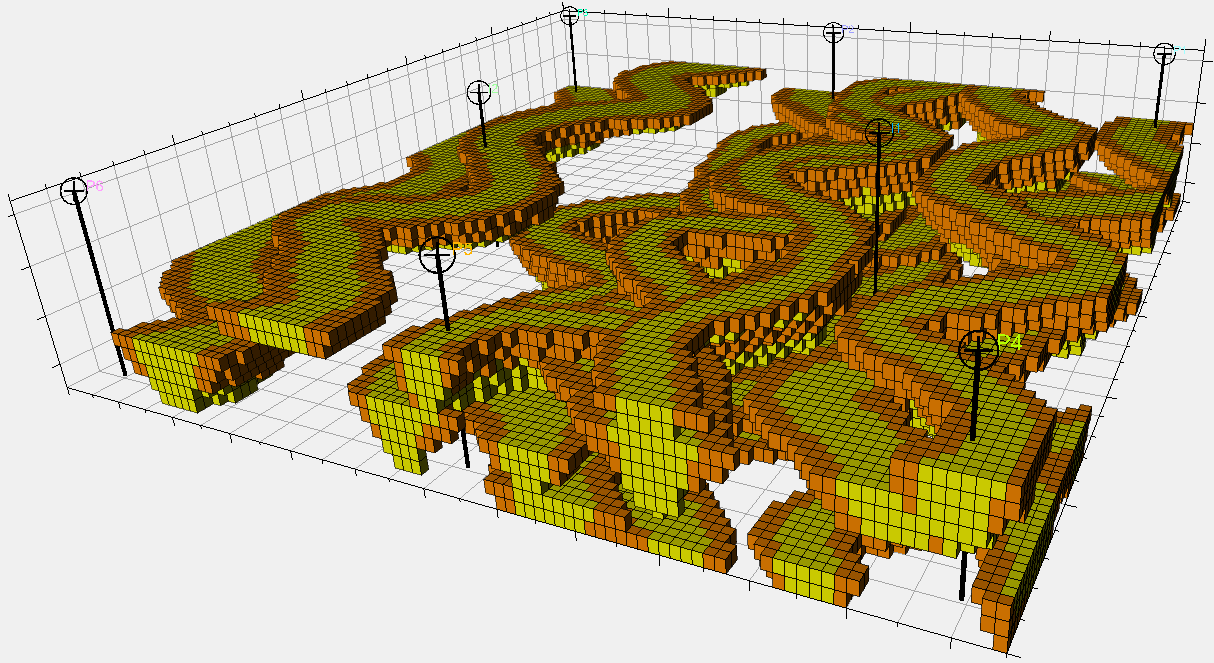}
    }
\caption{Reference model. Test case 3.}
\label{Fig:Case3-True}
\end{figure}

The last test case is a 3D model with fluvial channels generated with object-based simulation. The model has three facies: channel, levee and background sand. Figure~\ref{Fig:Case3-True} shows the permeability and the facies distribution of the reference case. We applied a transparency to the background sand in Fig.~\ref{Fig:Case3-True}b to allow the visualization of the geometry of the channels. We assumed a constant permeability for each facies: 2,000~mD in the channels, 1,000~mD in the levees and 100~mD in the background. This model has 100 $\times$ 100 $\times$ $10$ gridblocks, all gridblocks with 50~m $\times$ 50~m $\times$ 2~m. This reservoir produces with six wells placed near the borders of the model and operated by a constant bottom-hole pressure of 10,000~kPa. There are also two water injection wells placed at the center of the model operating with a fixed bottom-hole pressure of 50,000~kPa.

The 3D geometry of the channel makes this problem particulary challenging because standard convolutional layers are designed for 2D images. One possible approach is to consider each layer of the reservoir model separately. This is the approach used in \citep{laloy:17b}. However, this procedure do not account for the geometry of the facies in the vertical direction as the convolutional operations are performed in 2D. Instead, we used the 3D convolutional layers available in \verb"TensorFlow". Even though the extension of convolutional operations to three dimension is conceptually simple, its training becomes computationally challenging. In fact, Geoffrey Hinton described the used of 3D convolutional networks as a ``nightmare'' \citep{hinton:12a}. The architecture of the network is described Table~\ref{Tab:CVAE-Case3} in the Appendix. In this network, we introduced batch normalization layers \citep{ioffe:15a} to improve stability and reduce training times. This procedure removed the need of the dropout layers used in the previous networks. We considered a training set of 40,000 realization and 10,000 for validation. The training took 49 hours in a cluster with four GPUs (NVIDIA TESLA P100) and the reconstruction accuracy was 89.1\%.

\begin{figure}
  \centering
    \includegraphics[width=0.49\linewidth]{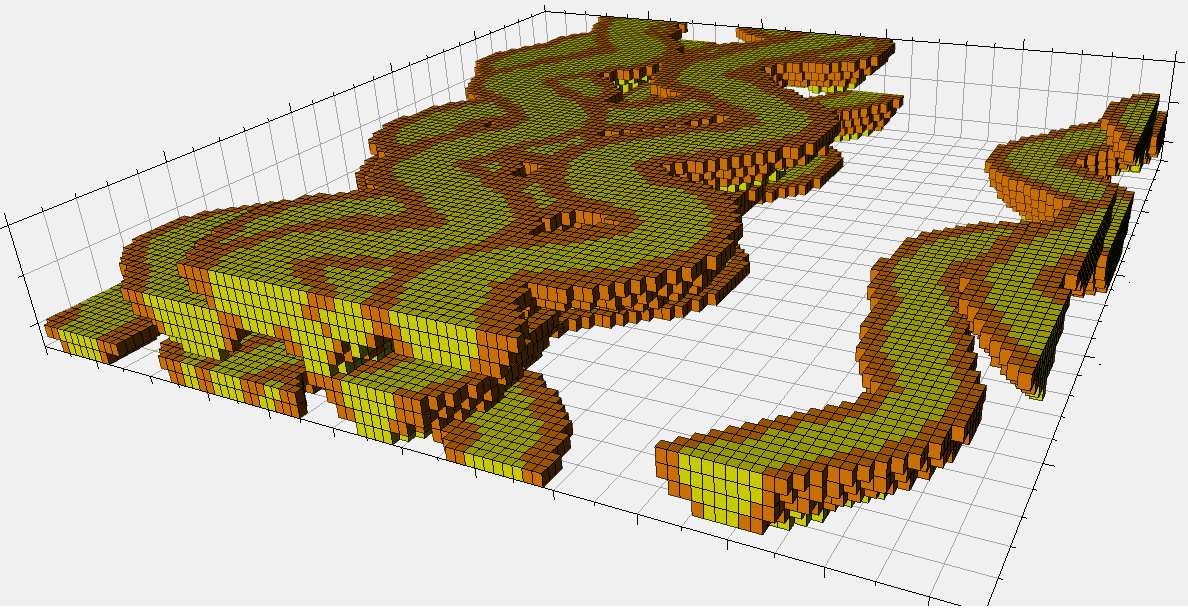}
    \includegraphics[width=0.49\linewidth]{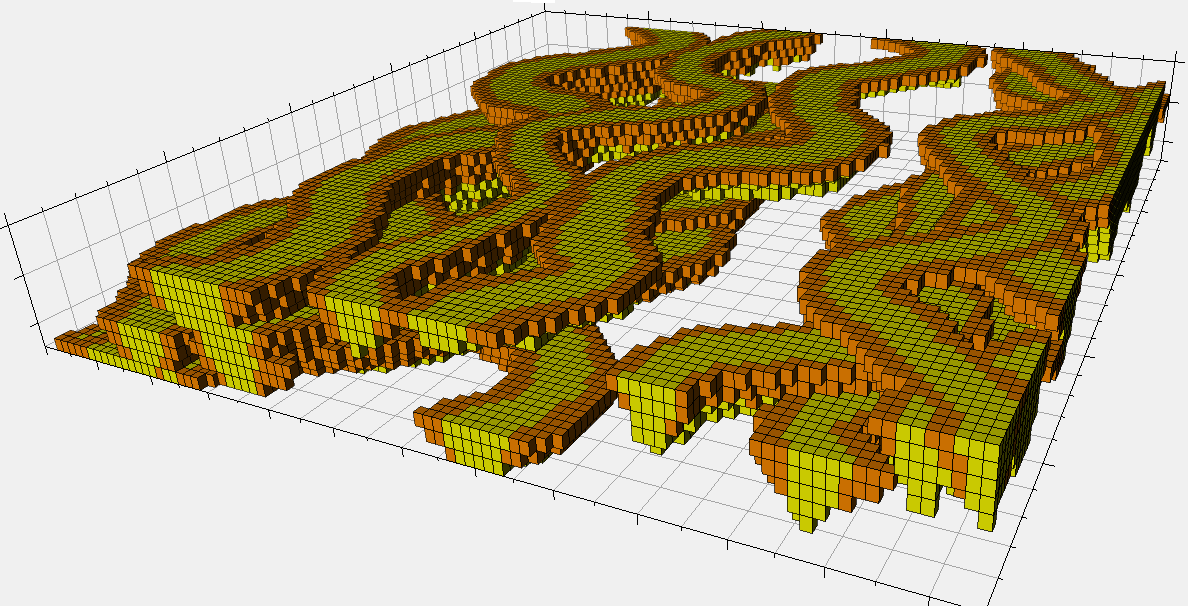}
 \linebreak
    \includegraphics[width=0.49\linewidth]{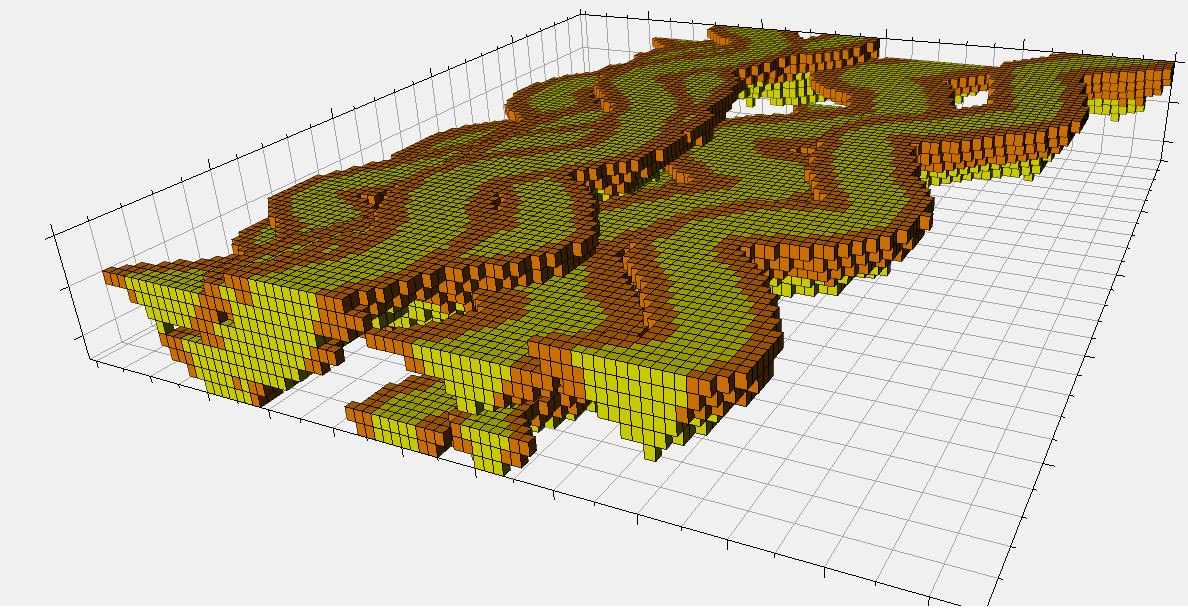}
    \includegraphics[width=0.49\linewidth]{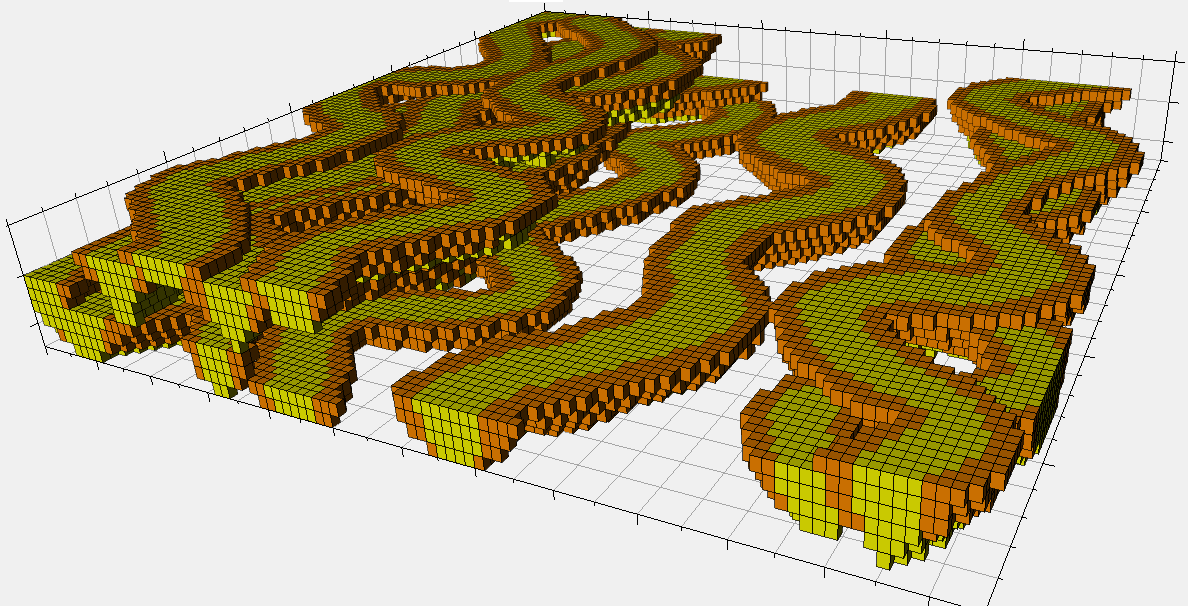}
\caption{First four prior realizations of facies. Test case 3.}
\label{Fig:Case3-Prior}
\end{figure}

\begin{figure}
  \centering
    \includegraphics[width=0.49\linewidth]{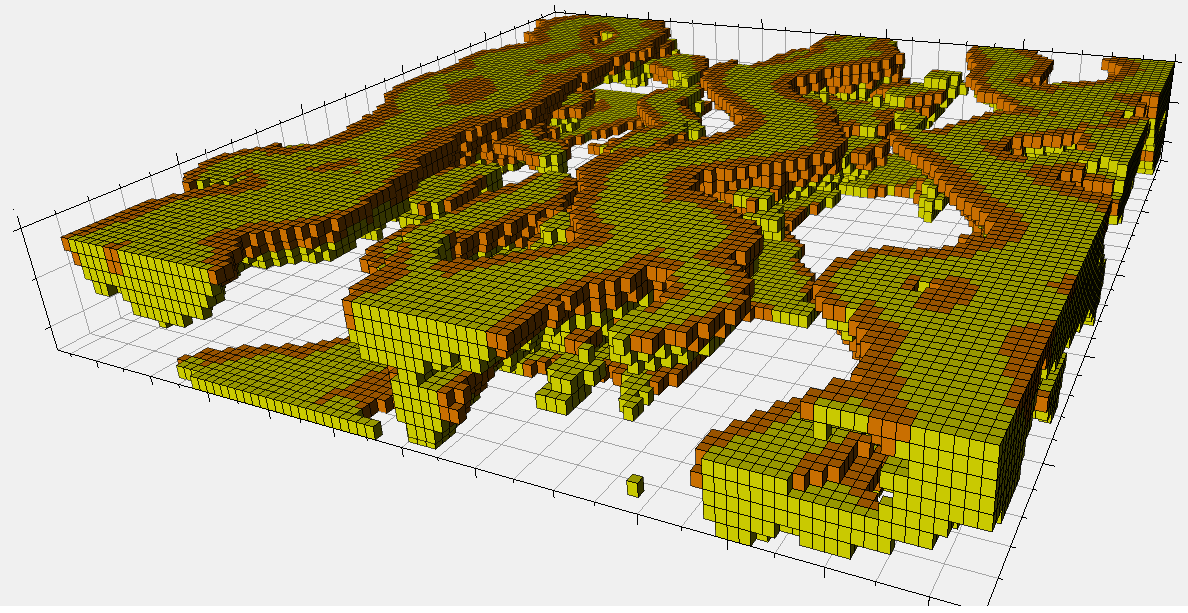}
    \includegraphics[width=0.49\linewidth]{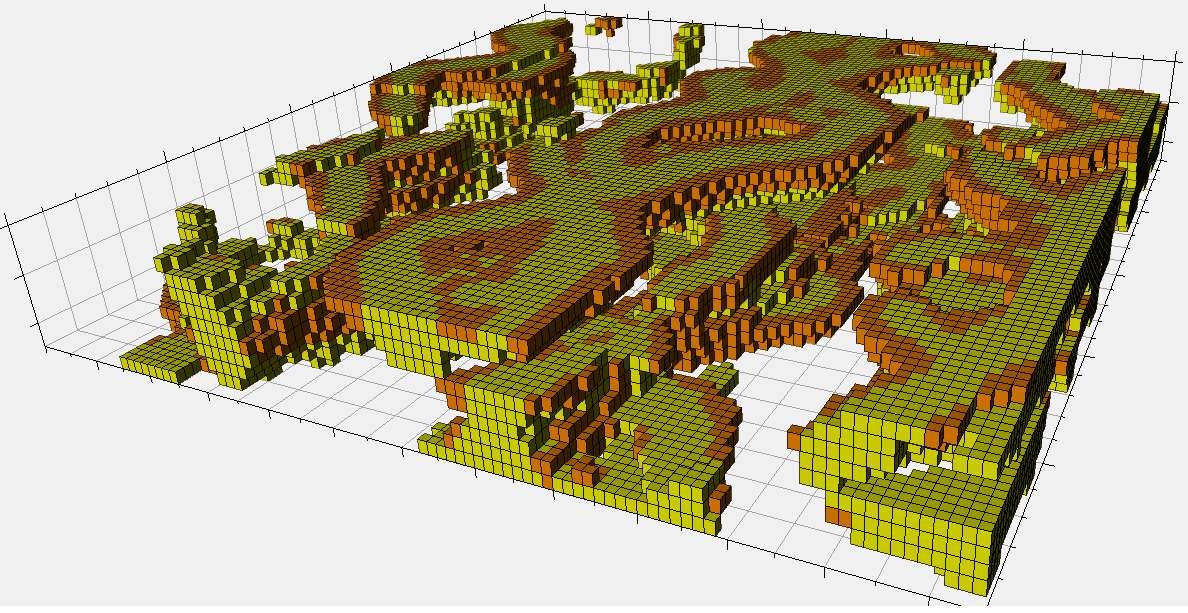}
 \linebreak
    \includegraphics[width=0.49\linewidth]{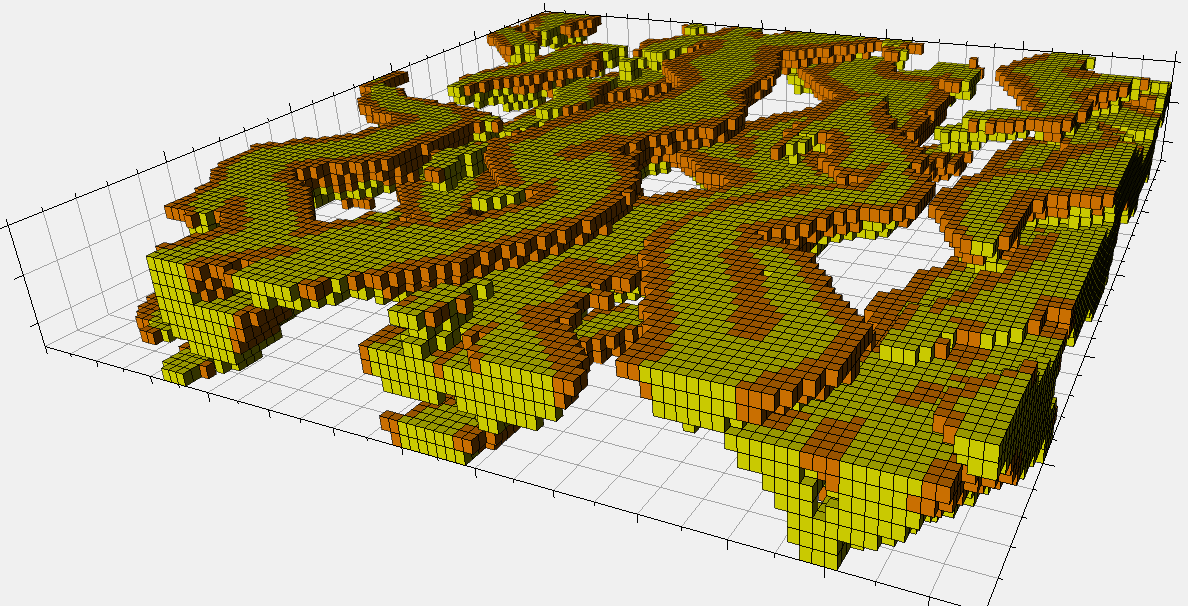}
    \includegraphics[width=0.49\linewidth]{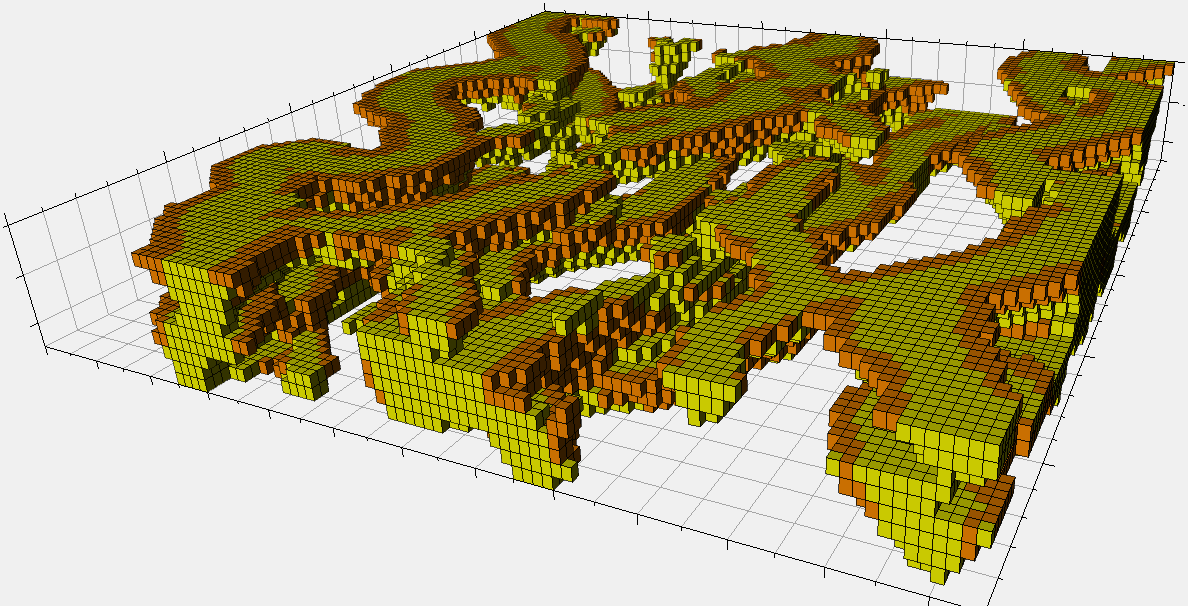}
\caption{First four posterior realizations of facies. Test case 3.}
\label{Fig:Case3-Post}
\end{figure}

\begin{figure}
  \centering
        \includegraphics[width=0.17\linewidth]{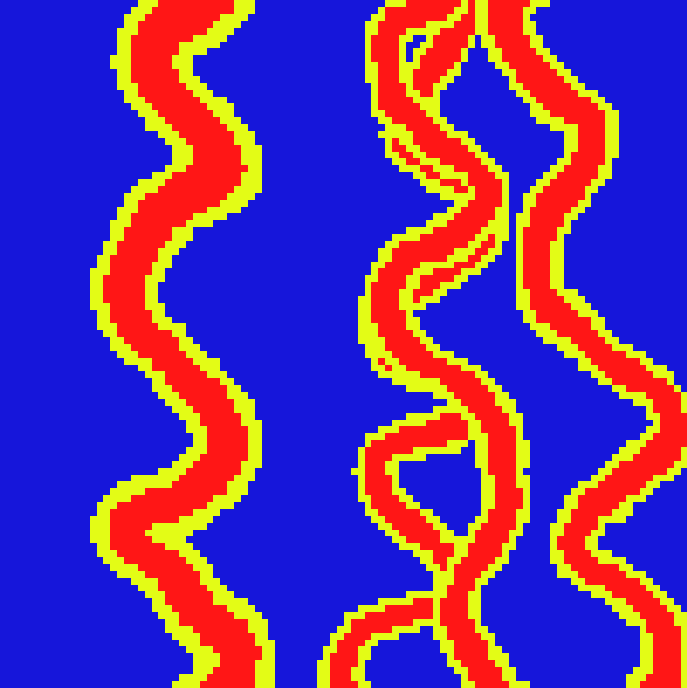}
        \includegraphics[width=0.17\linewidth]{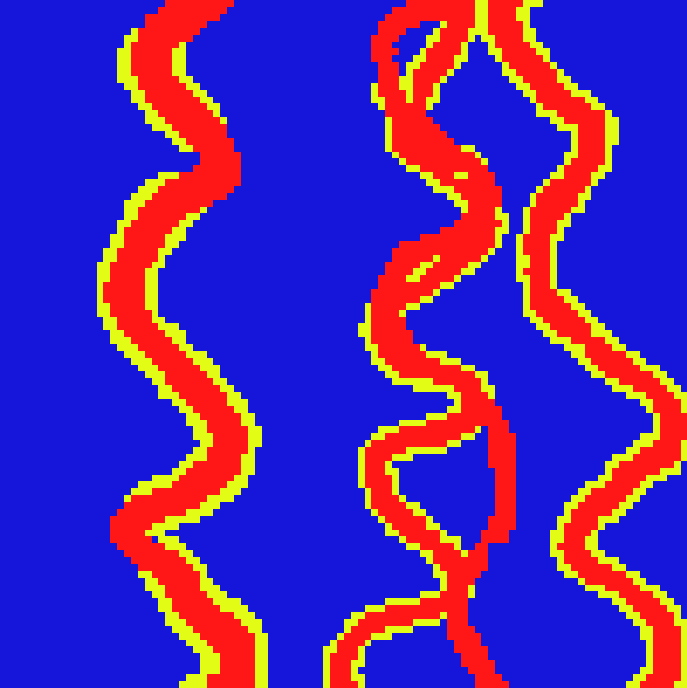}
        \includegraphics[width=0.17\linewidth]{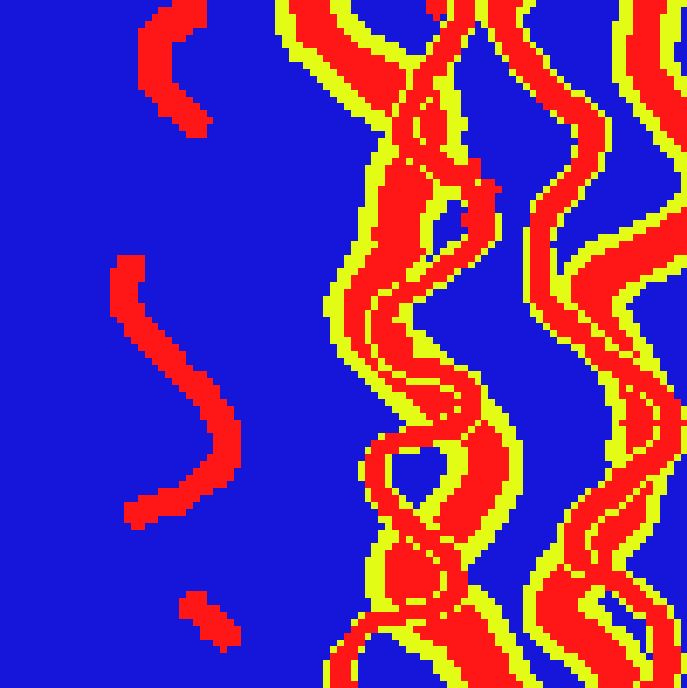}
        \includegraphics[width=0.17\linewidth]{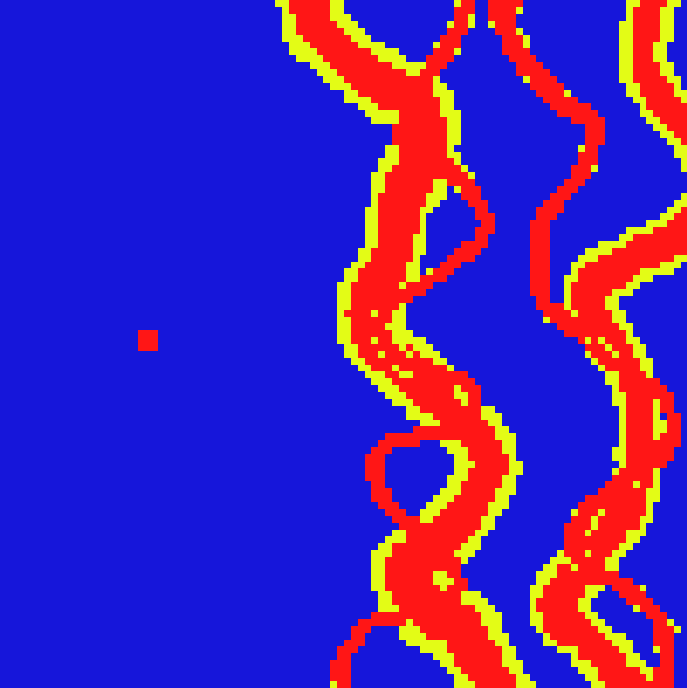}
        \includegraphics[width=0.17\linewidth]{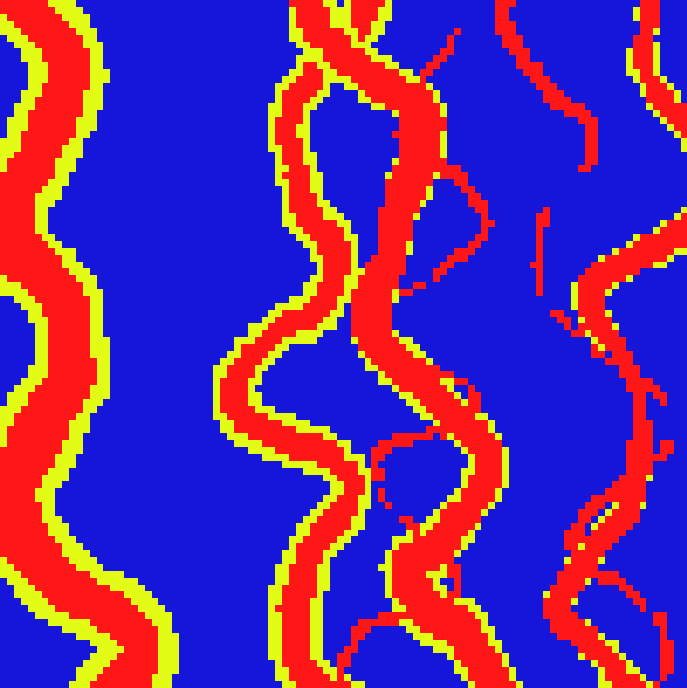}
\linebreak
        \includegraphics[width=0.17\linewidth]{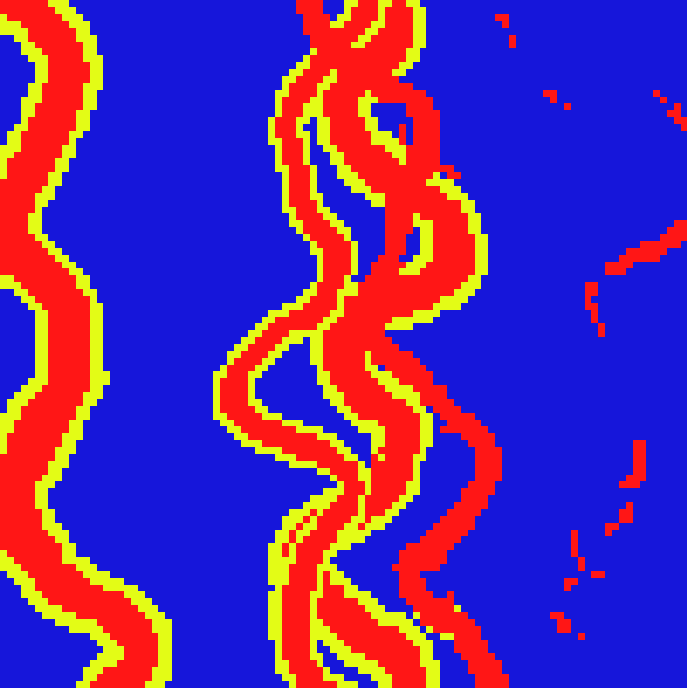}
        \includegraphics[width=0.17\linewidth]{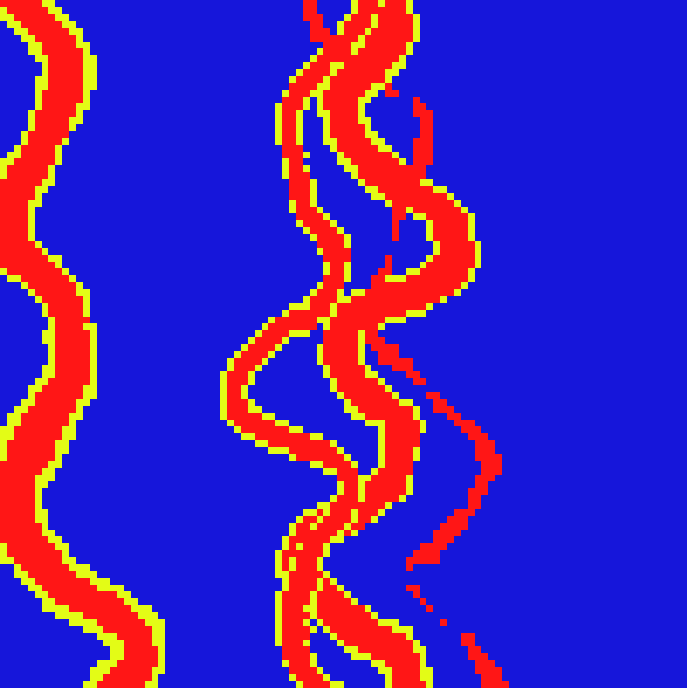}
  \subfloat[Reference]{
        \includegraphics[width=0.17\linewidth]{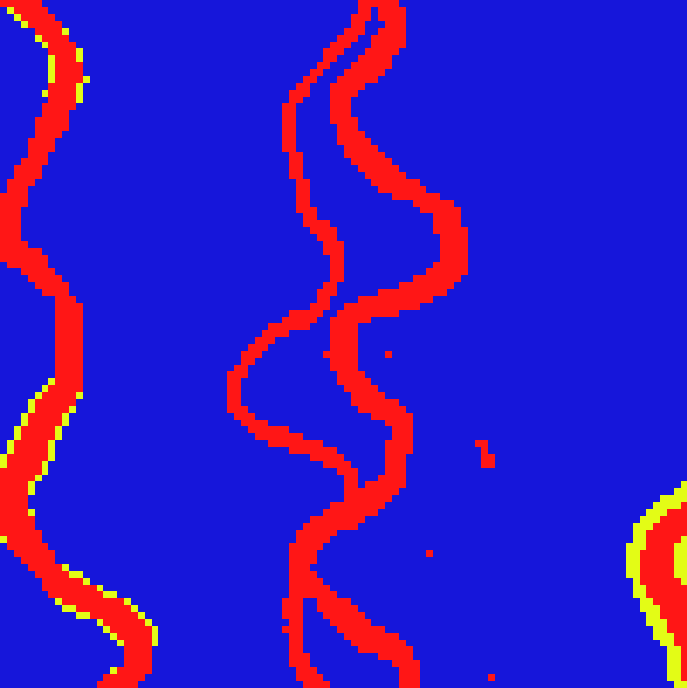}
  }
        \includegraphics[width=0.17\linewidth]{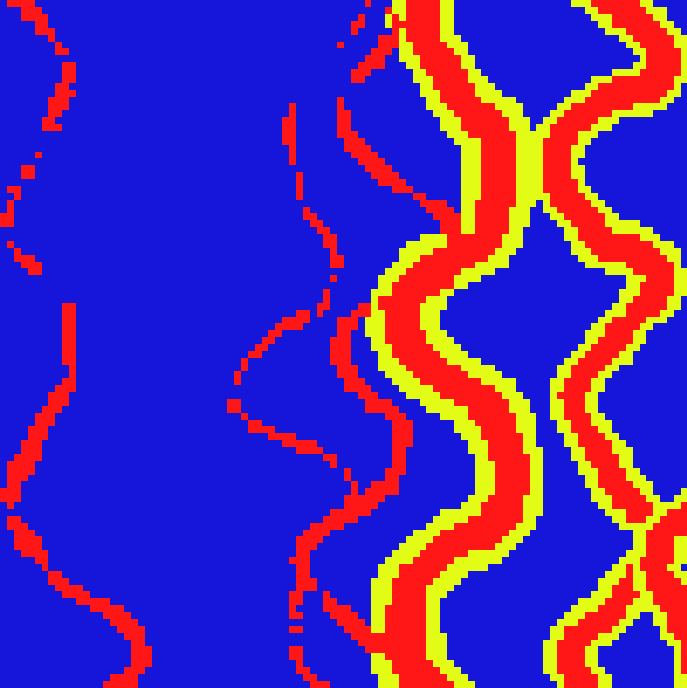}
        \includegraphics[width=0.17\linewidth]{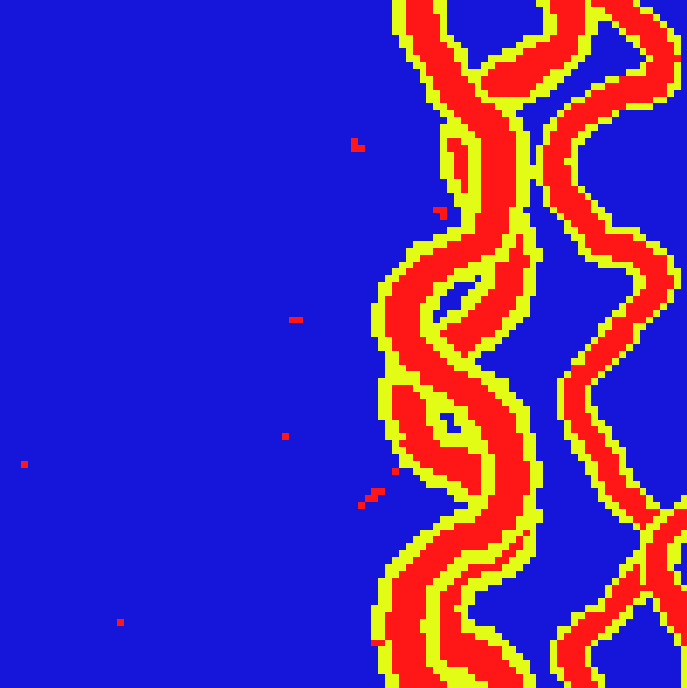}
\linebreak
        \includegraphics[width=0.17\linewidth]{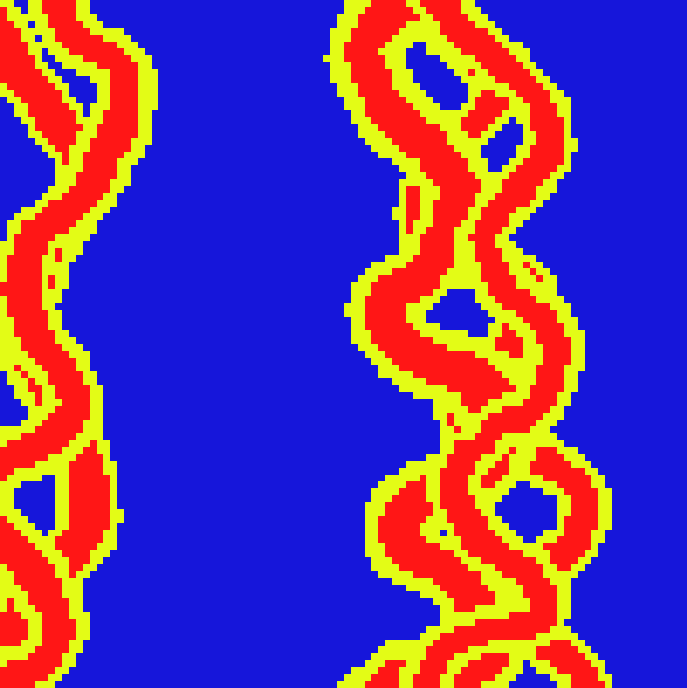}
        \includegraphics[width=0.17\linewidth]{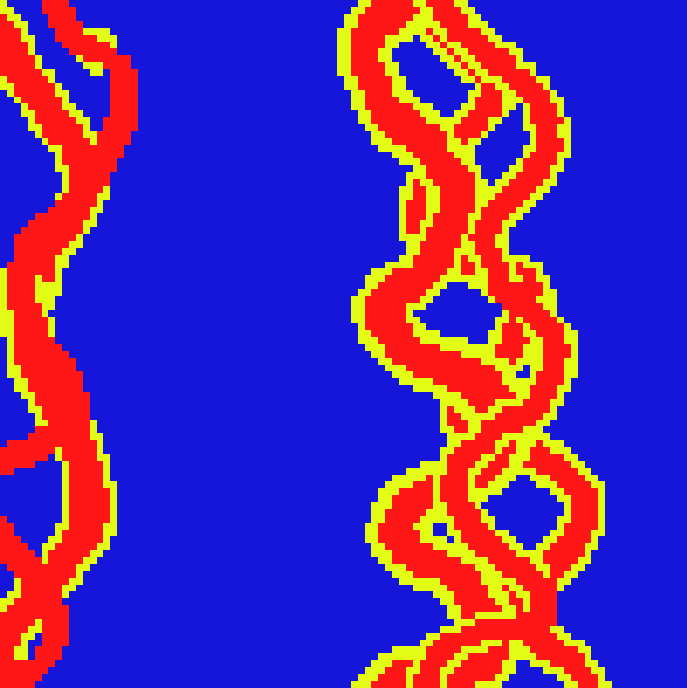}
        \includegraphics[width=0.17\linewidth]{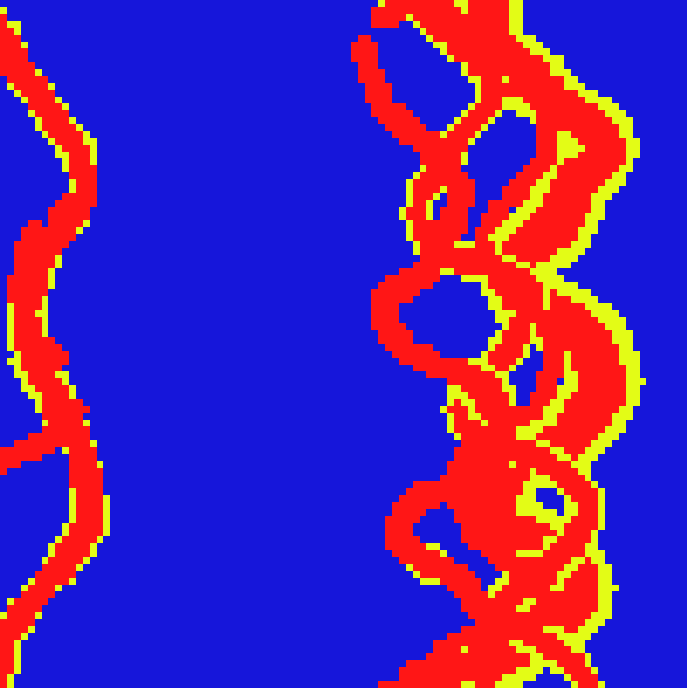}
        \includegraphics[width=0.17\linewidth]{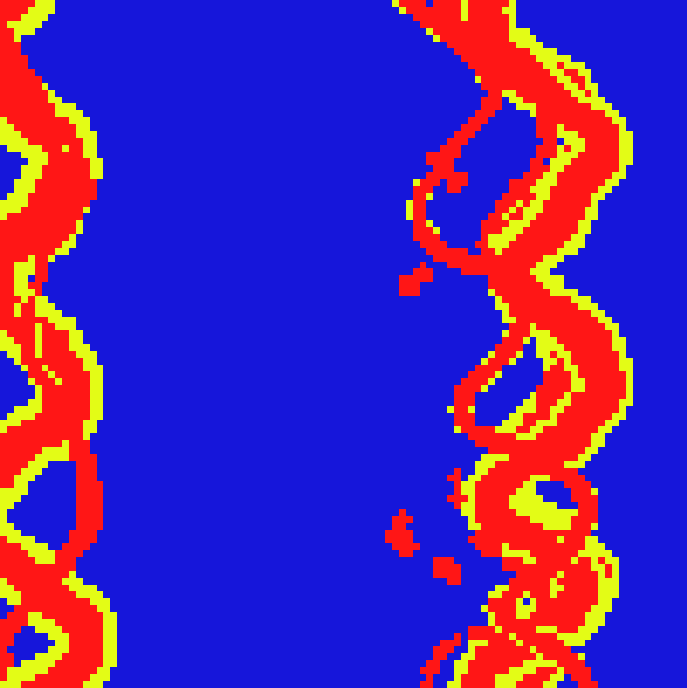}
        \includegraphics[width=0.17\linewidth]{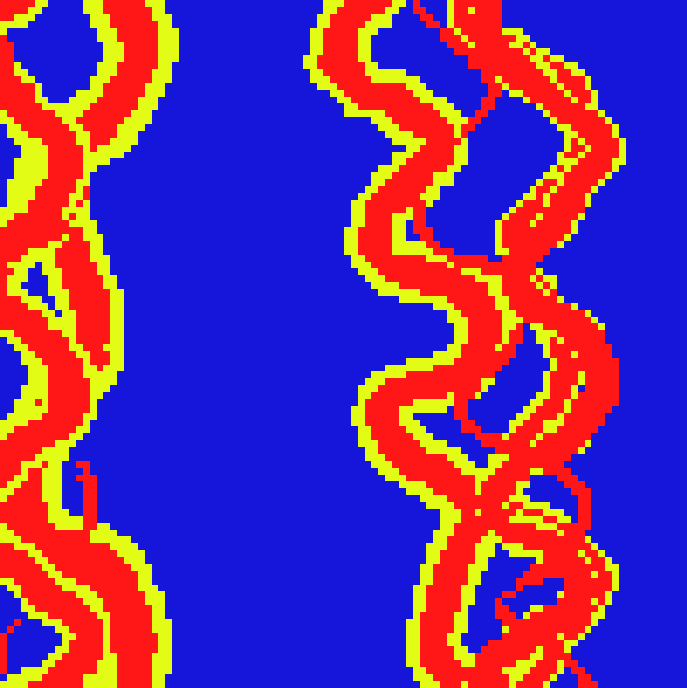}
\linebreak
        \includegraphics[width=0.17\linewidth]{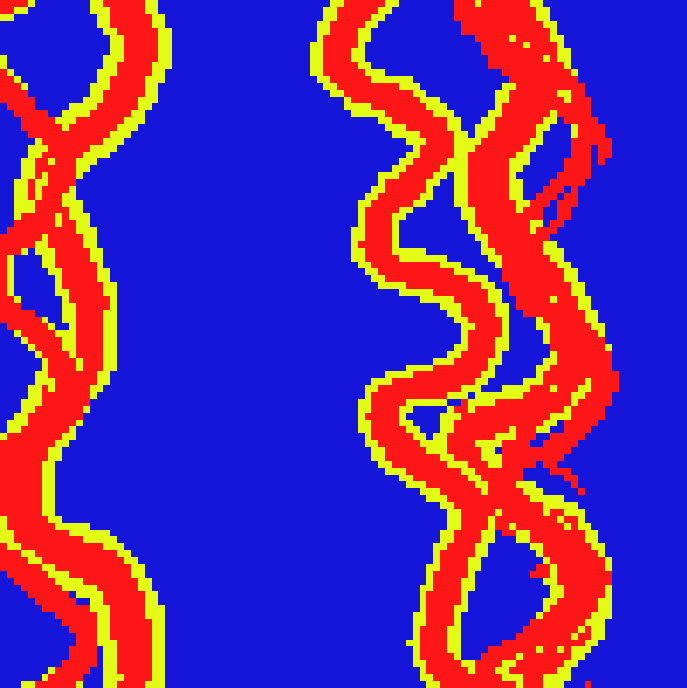}
        \includegraphics[width=0.17\linewidth]{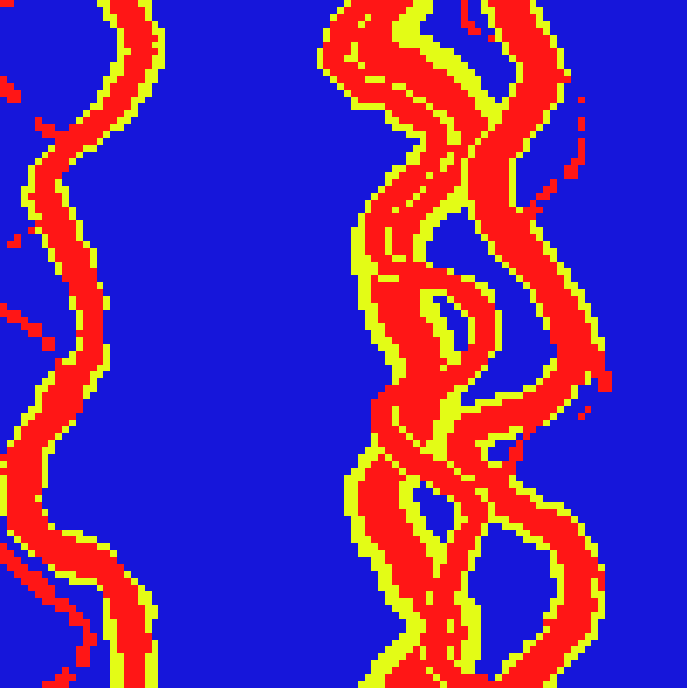}
  \subfloat[Prior]{
        \includegraphics[width=0.17\linewidth]{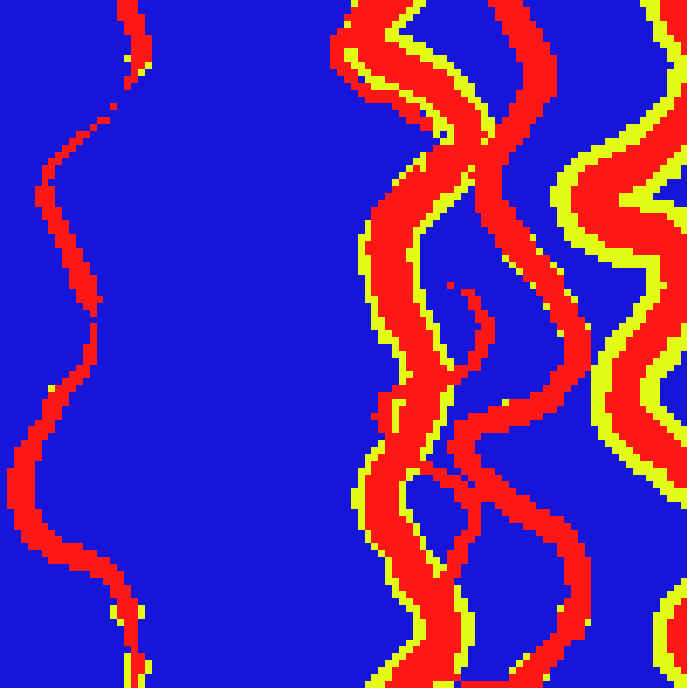}
  }
        \includegraphics[width=0.17\linewidth]{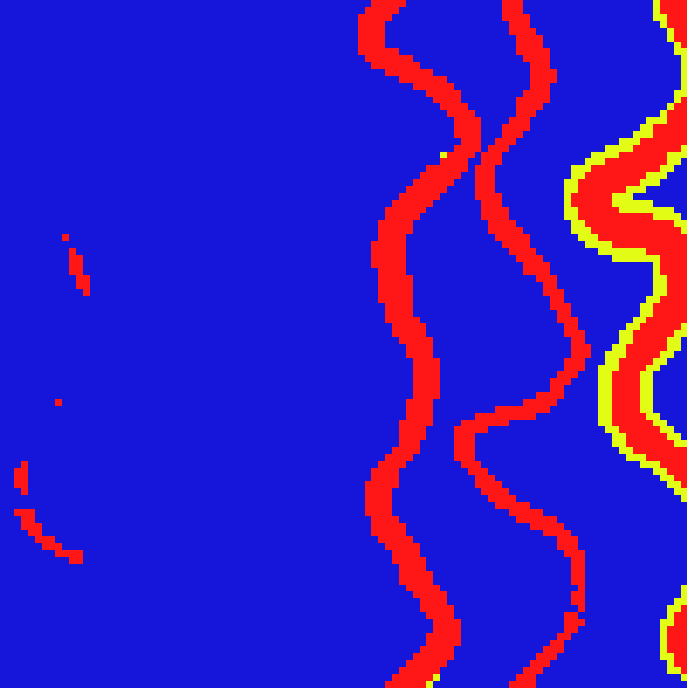}
        \includegraphics[width=0.17\linewidth]{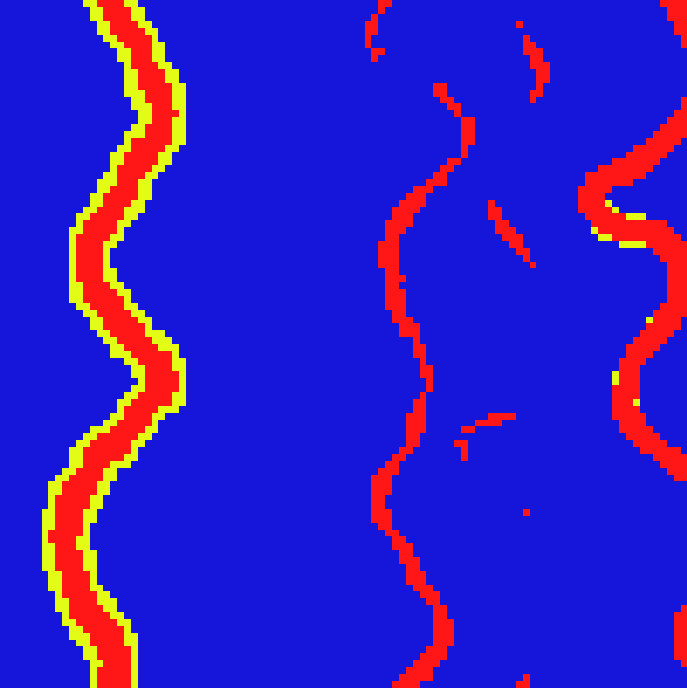}
\linebreak
        \includegraphics[width=0.17\linewidth]{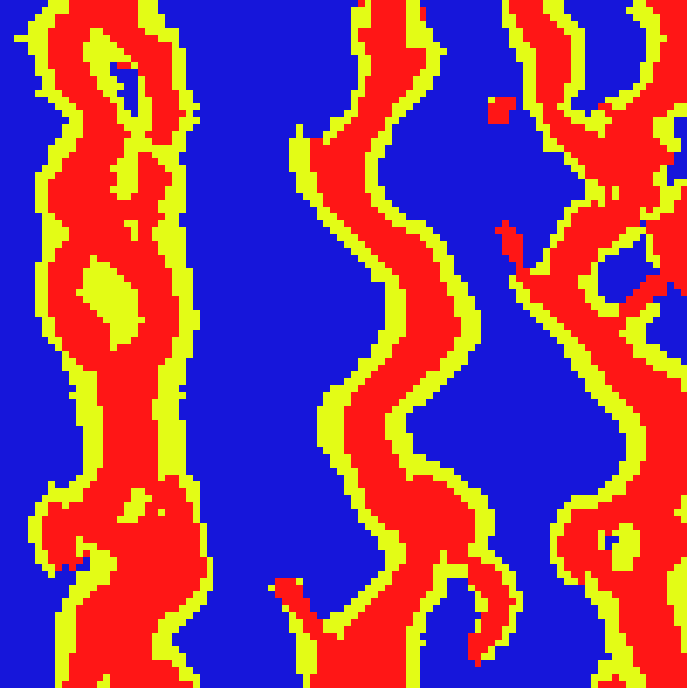}
        \includegraphics[width=0.17\linewidth]{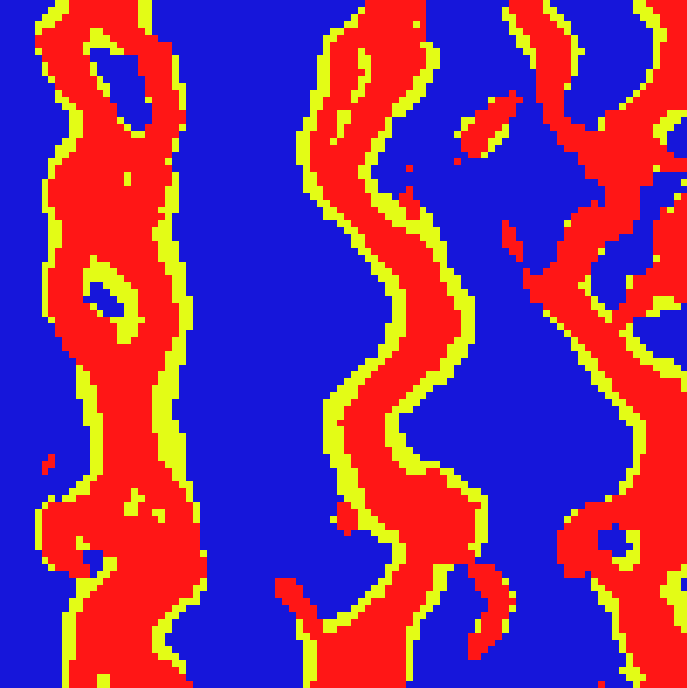}
        \includegraphics[width=0.17\linewidth]{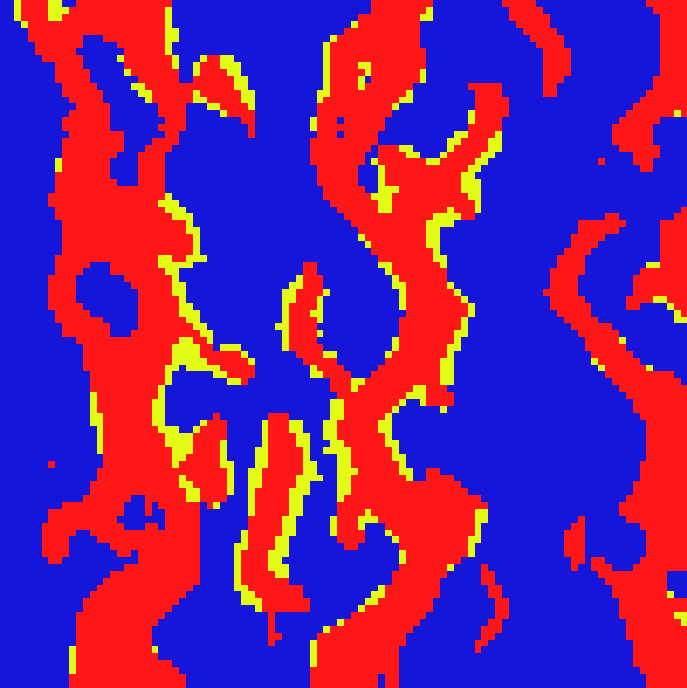}
        \includegraphics[width=0.17\linewidth]{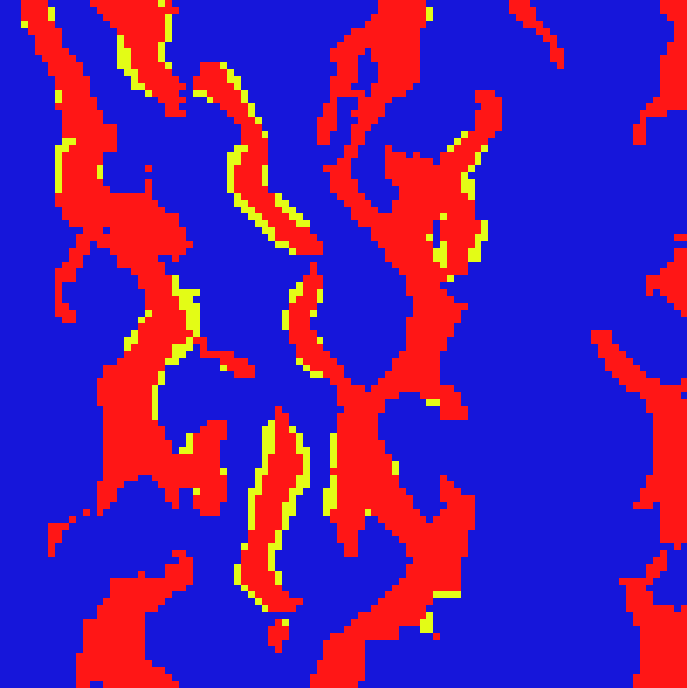}
        \includegraphics[width=0.17\linewidth]{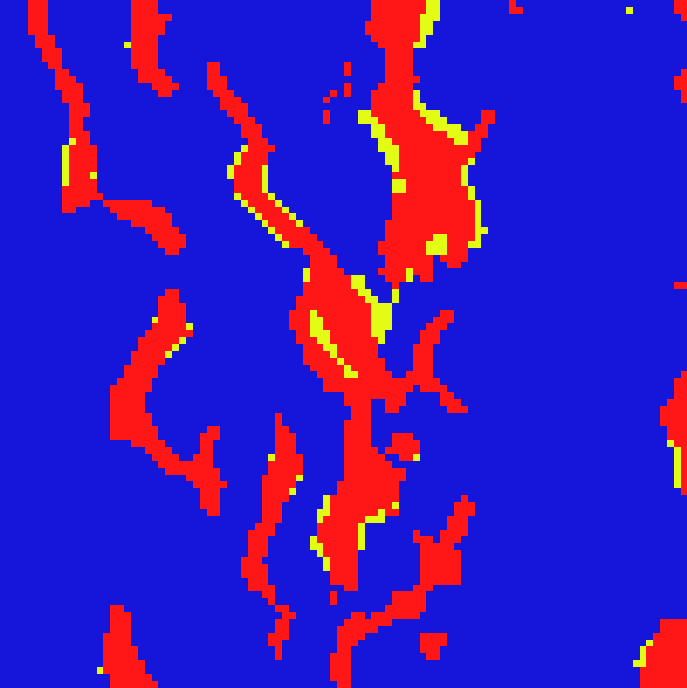}
\linebreak
        \includegraphics[width=0.17\linewidth]{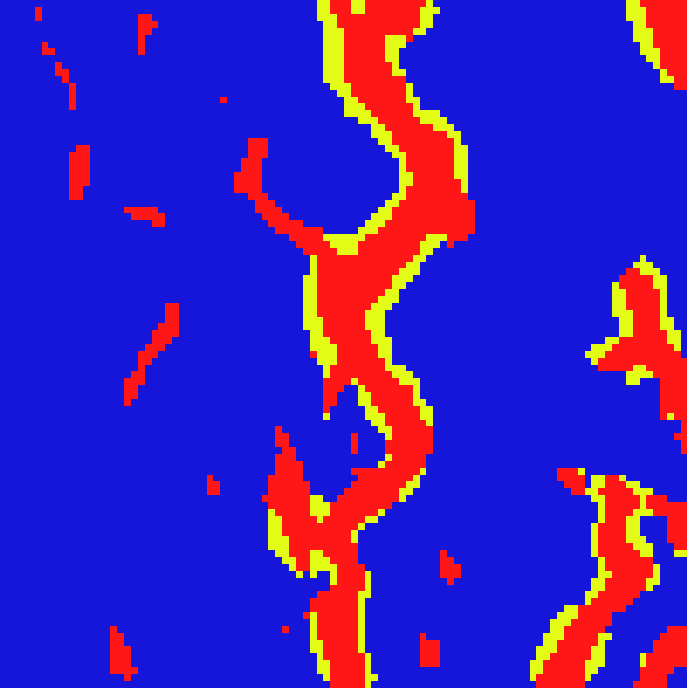}
        \includegraphics[width=0.17\linewidth]{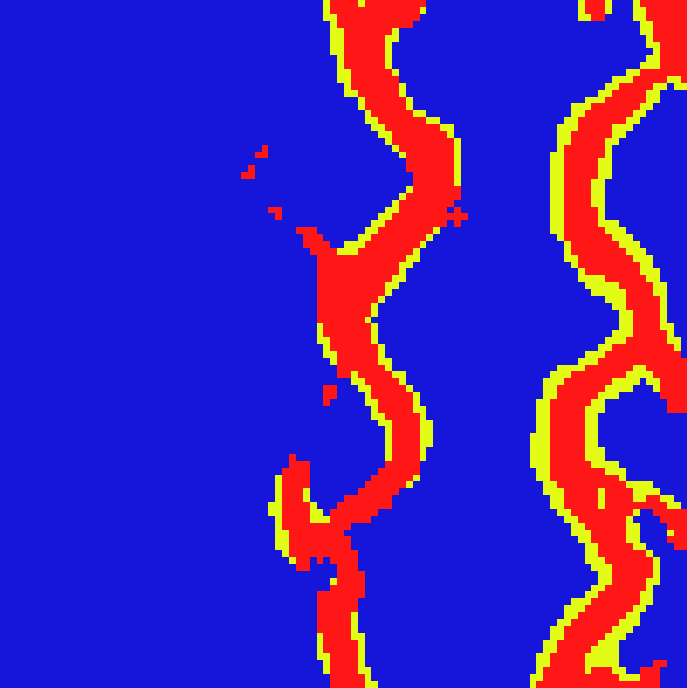}
  \subfloat[Posterior]{
        \includegraphics[width=0.17\linewidth]{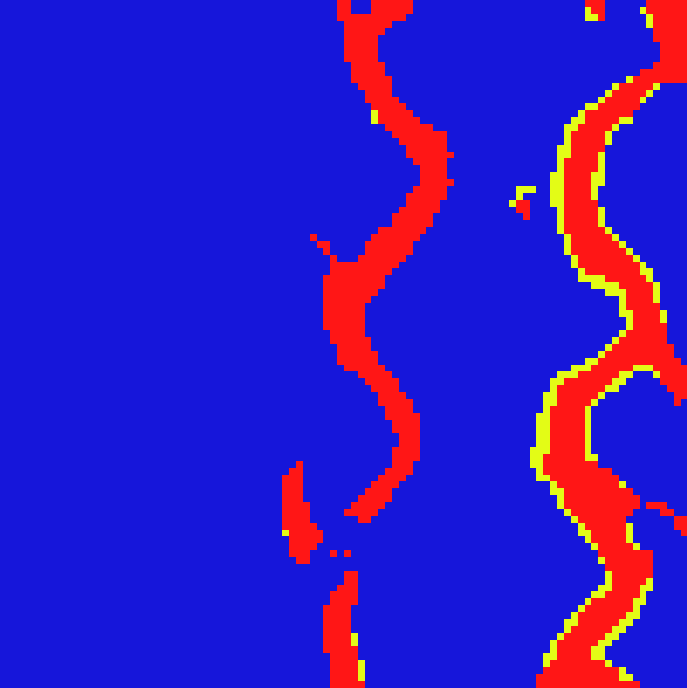}
  }
        \includegraphics[width=0.17\linewidth]{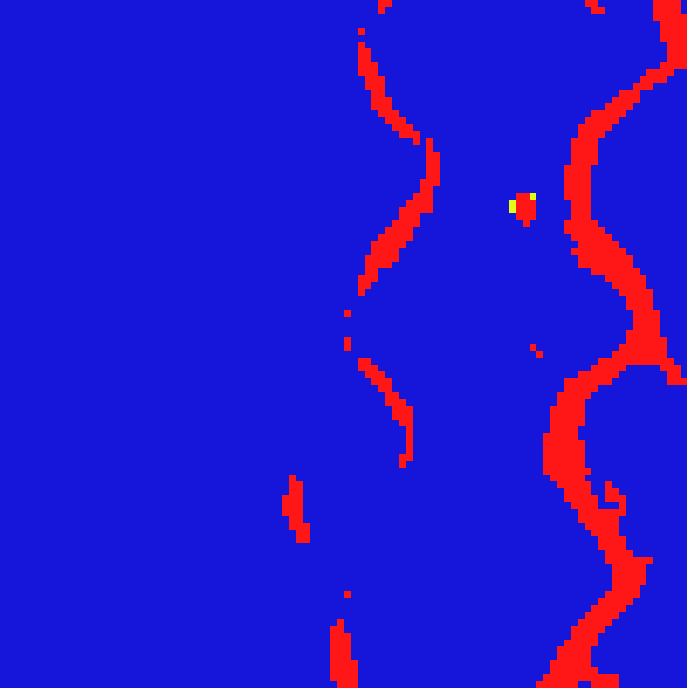}
        \includegraphics[width=0.17\linewidth]{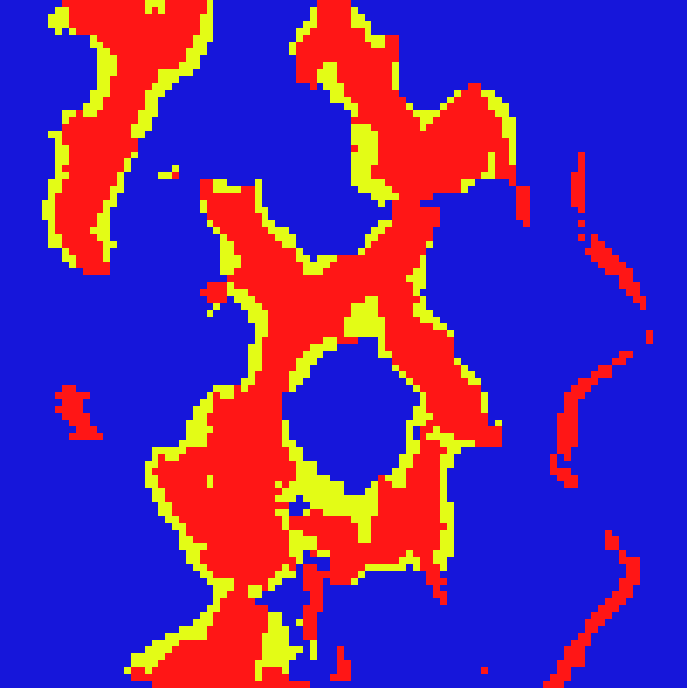}
\caption{Layer-by-layer permeability of the reference and the first realization before and after data assimilation. Test case 3.}
\label{Fig:Case3-Model1}
\end{figure}

\begin{figure}
  \centering
  \subfloat[Well P1]{
        \includegraphics[width=0.4\linewidth]{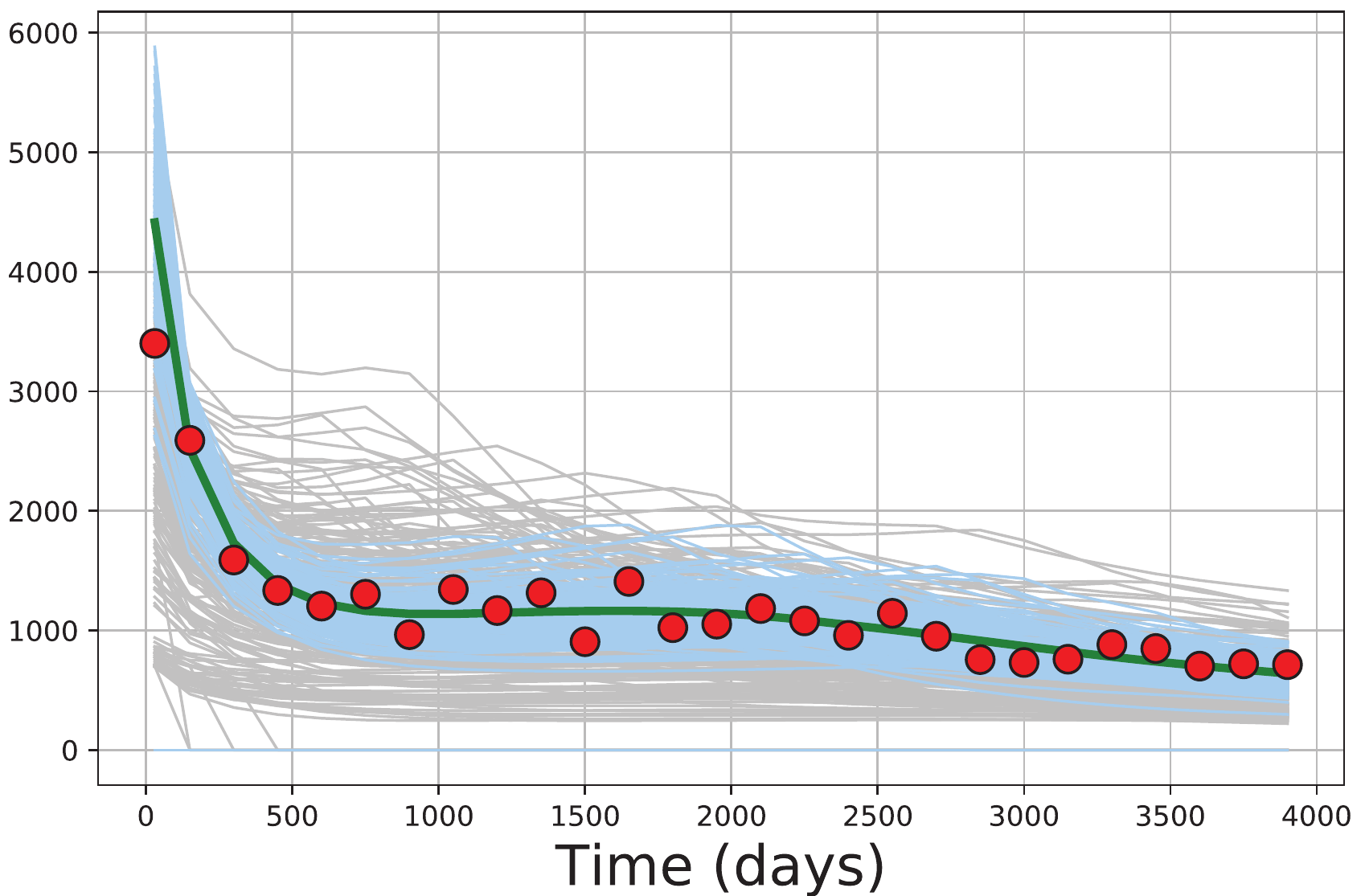}
  }
  \subfloat[Well P2]{
        \includegraphics[width=0.4\linewidth]{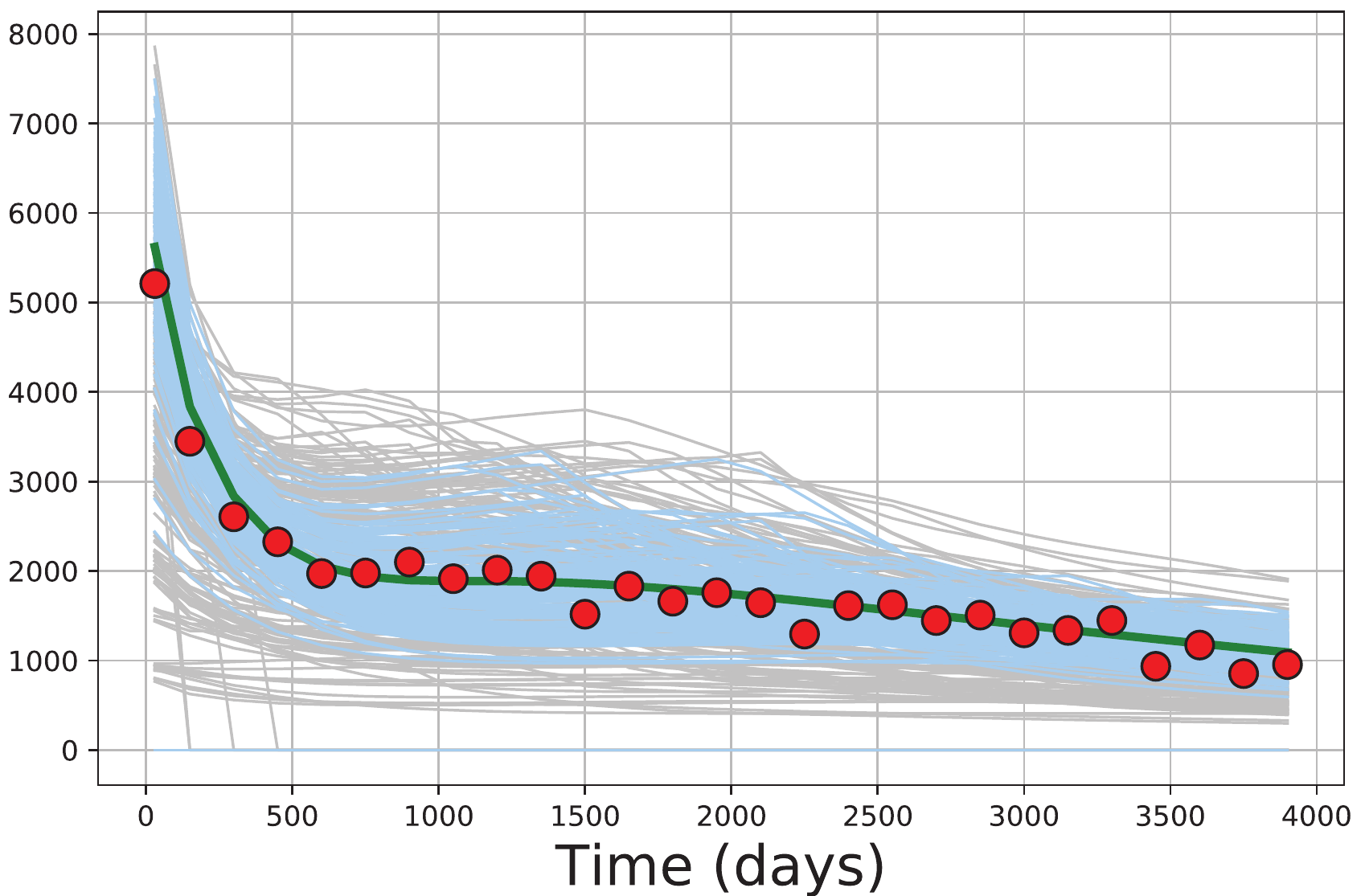}
  }
  \linebreak
  \subfloat[Well P3]{
        \includegraphics[width=0.4\linewidth]{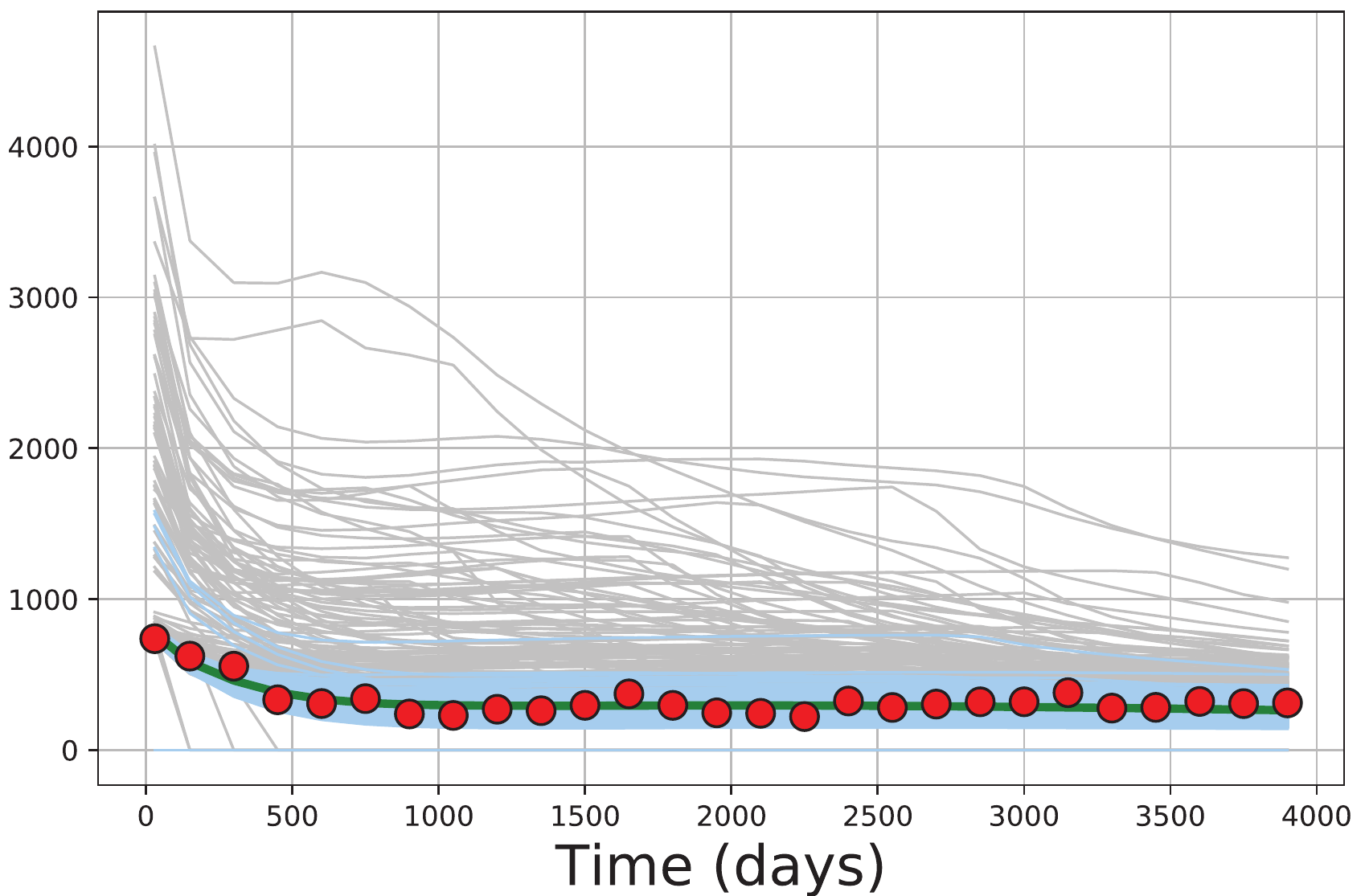}
  }
  \subfloat[Well P4]{
        \includegraphics[width=0.4\linewidth]{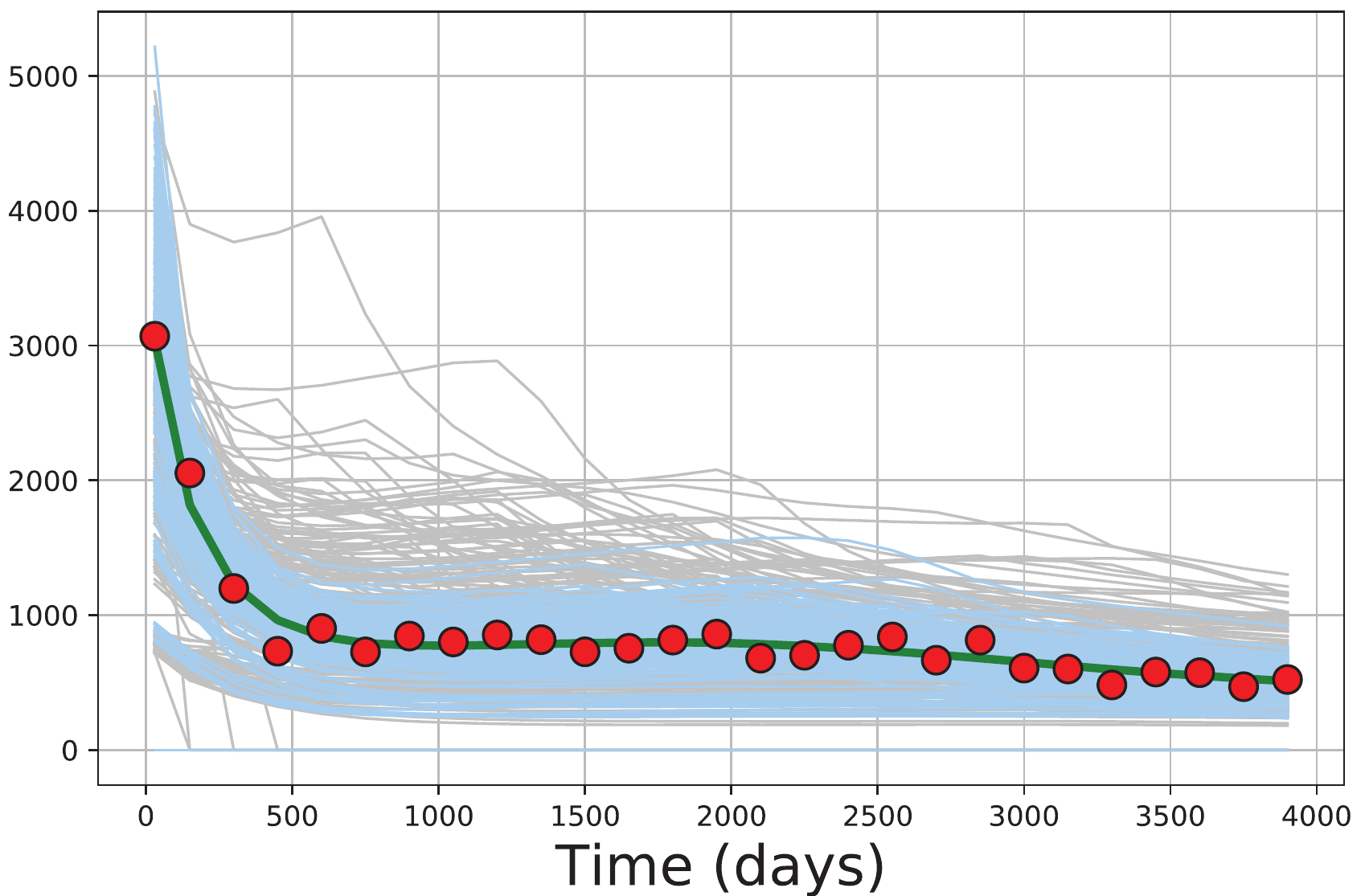}
  }
\caption{Oil rate (m$^3$/days). Test case 3. Red dots are the observed data points, gray and blue curves are the predicted data from the prior and posterior ensembles, respectively. The green curve is the mean of the posterior ensemble.}
\label{Fig:Case3-OPR}
\end{figure}

We applied ES-MDA-CVAE with an ensemble of $N_e = 200$ realizations and $N_a = 20$ MDA iterations with constant inflation factors. The observations corresponded to oil and water rate predicted by the reference case and corrupted with random noise of 5\%. Figures~\ref{Fig:Case3-Prior} and \ref{Fig:Case3-Post} show four realizations of the prior and posterior ensembles, respectively. Overall, the ES-MDA-CVAE was able to preserve several channels with the desired characteristics. Figure~\ref{Fig:Case3-Model1} shows all ten layers of the reference model and the first realization before and after data assimilation. This figure shows that the posterior realization present some facies with well-defined channel-levee sequences. However, the posterior model is clearly distinguishable from the prior and reference models with some discontinuous channels and some oddly-shaped facies; see, e.g., the bottom layer of the posterior realization (Fig.~\ref{Fig:Case3-Model1}c). Ideally, we would like the posterior realizations to be visually indistinguishable from prior realizations generated with the object-based algorithm. Nevertheless, these results are very encouraging and far superior to what would be obtained with standard ES-MDA or even with a OPCA parameterization (this case is no computationally feasible with our previous DBN implementation). In fact, it is important mentioning that this type of model is extremely difficult to history match with the current methods available. Figure~\ref{Fig:Case3-OPR} shows the oil rate at four wells indicating significant improvements in the predictions, although there are still some realizations with poor data matches; for example, there are some models predicting zero oil rates.

\clearpage

\section{Comments}
\label{Sec:Comments}

One important limitation of CVAE parameterization is the fact that we cannot apply distance-based localization \citep{houtekamer:01} to update $\z$ because this vector is in a different space. Hence, it does not make sense to compute the Euclidian distance between a component of $\z$ and the spatial position of a well. Yet, localization is important to mitigate the negative effects of sampling errors and limited degrees of freedom in ensemble data assimilation. In \citep{canchumuni:18a}, we tried to work around this issue by using the number of neurons in the code layer equals to the number of reservoir gridblocks. However, this procedure does not ensure the existence of a direct relation between the entry of $\z$ and the corresponding spatial location of the gridblock in the reservoir model. In fact, because the convolutional layers share parameters, it is conceivable that each component of $\z$ may be associated with the reconstruction of the facies in different regions in the reservoir (the same weights of the convolutional kernels are applied to multiple locations of the input data). Moreover, using the size of the code layer equals to the size of reservoir grid increases significantly the number of training parameters because of the fully-connected layers. In practice, this may make the application unfeasible for larger reservoir models. There are localization procedures which are not formulated in terms of spatial distances that could be applied in this case; see, e.g., \citep{lacerda:18a} and references therein. Unfortunately, in our experience, these procedures are less effective than distance-dependent approaches. We did not use any type of localization in any of the test cases described in this paper. Nevertheless, this is definitely an issue that needs further investigation.

Another practical problem for the application of the method investigated in this paper is the need for a large number of prior realizations to train the CVAE. In the tests cases considered in this paper, we used values between 30,000 and 50,000 realizations. However, in practice, we may need larger numbers for more complex models. Unfortunately, generating several realizations of the geological model with standard geostatistical algorithms may be very challenging. One possible solution we intend to investigate in the future is to use data augmentation \citep{yaeger:97a,taylor:17a} and transfer learning techniques \citep{hoo:16a,cheng:17a}. Data augmentation consist of a series of affine transformations applied to the input data to increase the training set. Typical augmentation strategies include mirroring, cropping and rotating images. Transfer learning is a strategy to use previously trained networks either as initialization or fixed parts of the implemented network. For example, in a preliminary test we applied the parameters of the network trained for the first test case as an initialization for the network in the second test case. This process resulted in a reduction of 50\% in the training time. Finally, it is necessary to investigate procedures to reduce the computational requirements for training the networks. Note that our last test case has 100,000 gridblocks, which is relatively small compared to the size of the models employed operationally. Yet, the training required approximately two days in a cluster with four GPUs.

\section{Conclusions}
\label{Sec:Conclusion}

In this paper, we investigated the use of a CVAE to parameterize facies in geological models and used ES-MDA to condition these models to observed data. We tested the procedure in three synthetic reservoir history-matching problems with channelized features and increasing level of complexity. The first two test problems corresponded 2D cases. The proposed procedure outperformed previous results obtained with standard ES-MDA, ES-MDA with OPCA and DBN parameterizations. The third test problem considered 3D channels and three facies. This case required the use of 3D convolutional layers in the network increasing significantly the training time. There is also a noticeable decrease in the reconstruction accuracy for this case and the conditional realizations exhibit some features not present in the prior geological description of the model. Nevertheless, the overall performance of the method is very encouraging and indicates that the use of deep-learning-based parameterizations is a research direction worth pursuing. In the continuation of this research, we intend to use our trained CVAEs as the generative models in GANs. The objective is to improve the reconstruction accuracy, especially for the third test case.

\section*{Acknowledgement}
\label{Sec:Acknowledgement}

The authors thank Petrobras for the financial support.


\section*{Appendix: Architecture of the networks}
\label{Sec:Appendix}

\begin{landscape}

\begin{table}
\caption{CVAE architecture. Test case 1.}
\label{Tab:CVAE-Case1}
\begin{scriptsize}
\begin{center}
\begin{tabular}{lll}
\toprule
\textbf{Layer} & \textbf{Configuration}  & \textbf{Comment} \\
\midrule
  \multicolumn{3}{c}{\textbf{Encoder}}\\\cmidrule{2-2}
  Input & Shape = (45, 45, 2) & Two facies \\
  2D convolution 1 & Kernels = 32, size = (2, 2), stride = (2, 2), activation = ReLU  & -- \\
  2D convolution 2 & Kernels = 32, size = (3, 3), stride = (2, 2), activation = ReLU & -- \\
  2D convolution 3 & Kernels = 16, size = (3, 3), stride = (1, 1), activation = ReLU & -- \\
  Flatten & -- & Setup for the fully-connected layer \\
  Fully-connected 1 & Neurons = 1024, activation = ReLU & -- \\
  Dropout &  10\%  & Strategy to avoid overfitting \\
  Fully-connected 2 & Neurons = 100, activation = linear & Mean of the VAE \\
  Fully-connected 3 & Neurons = 100, activation = linear & Log-variance of the VAE \\
\midrule
  \multicolumn{3}{c}{\textbf{Code}}\\\cmidrule{2-2}
  Lambda & -- & Sampling $\z$ \\
\midrule
  \multicolumn{3}{c}{\textbf{Decoder}}\\\cmidrule{2-2}
  Fully-connected 4 & Neurons = 1024, activation = ReLU & -- \\
  Dropout &  10\%  & Strategy to avoid overfitting \\
  Fully-connected 5 & Neurons = 2034, activation = ReLU & -- \\
  Reshape & Output size = (12, 12, 16) & Setup for the transpose convolution \\
  2D transposed convolution 1 & Kernels = 16, size = (3, 3), stride = (1, 1), activation = ReLU & -- \\
  2D transposed convolution 2 & Kernels = 32, size = (3, 3), stride = (2, 2), activation = ReLU & -- \\
  2D transposed convolution 3 & Kernels = 32, size = (2, 2), stride = (1, 2), activation = ReLU & -- \\
  Bilinear up-sampling & Output size = (45, 45, 32) & Resize output dimension \\
  2D convolution 4 & Kernels = 2, size = (3, 3), stride = (1, 1), activation = sigmoid & Output image \\
\bottomrule
\end{tabular}
\end{center}
\end{scriptsize}
\end{table}
\end{landscape}

\begin{landscape}
\begin{table}
\caption{CVAE architecture. Test case 3.}
\label{Tab:CVAE-Case3}
\begin{scriptsize}
\begin{center}
\begin{tabular}{lll}
\toprule
\textbf{Layer} & \textbf{Configuration}  & \textbf{Comment} \\
\midrule
  \multicolumn{3}{c}{\textbf{Encoder}}\\\cmidrule{2-2}
  Input & Shape = (100, 100, 10, 3) & Three facies \\
  3D convolution 1 & Kernels = 32, size = (3, 3, 3), stride = (2, 2, 2), activation = ReLU  & -- \\
  Max-pooling 1 & Pool = (2, 2, 1) & Dimension reduction \\
  3D convolution 2 & Kernels = 32, size = (3, 3, 3), stride = (2, 2, 2), activation = ReLU & -- \\
  3D convolution 3 & Kernels = 32, size = (2, 2, 2), stride = (1, 1, 1), activation = ReLU & -- \\
  3D convolution 4 & Kernels = 16, size = (2, 2, 2), stride = (1, 1, 1), activation = ReLU & -- \\
  Max-pooling 2 & Pool = (2, 2, 1) & Dimension reduction \\
  Flatten & -- & Setup for the fully-connected layer \\
  Fully-connected 1 & Neurons = 2000, activation = linear & -- \\
  Batch normalization & -- & Regularization \\
  Activation & ReLU & -- \\
  Fully-connected 2 & Neurons = 100, activation = linear & Mean of the VAE \\
  Fully-connected 3 & Neurons = 100, activation = linear & Log-variance of the VAE \\
\midrule
  \multicolumn{3}{c}{\textbf{Code}}\\\cmidrule{2-2}
  Lambda & -- & Sampling $\z$ \\
\midrule
  \multicolumn{3}{c}{\textbf{Decoder}}\\\cmidrule{2-2}
  Fully-connected 4 & Neurons = 2000, activation = linear & -- \\
  Batch normalization & -- & Regularization \\
  Activation & ReLU & -- \\
  Fully-connected 5 & Neurons = 5000, activation = ReLU & -- \\
  Reshape & Output size = (25, 25, 5, 16) & Setup for the transposed convolution \\
  3D transposed convolution 1 & Kernels = 16, size = (2, 2, 2), stride = (1, 1, 1), activation = ReLU & -- \\
  3D transposed convolution 2 & Kernels = 32, size = (2, 2, 2), stride = (1, 1, 1), activation = ReLU & -- \\
  3D transposed convolution 3 & Kernels = 32, size = (3, 3, 3), stride = (2, 2, 2), activation = ReLU & -- \\
  3D transposed convolution 4 & Kernels = 32, size = (3, 3, 3), stride = (2, 2, 2), activation = ReLU & -- \\
  3D transposed convolution 5 & Kernels = 3, size = (3, 3, 3), stride = (1, 1, 2), activation = sigmoid & Output \\
\bottomrule
\end{tabular}
\end{center}
\end{scriptsize}
\end{table}
\end{landscape}

\end{document}